%% file: ELSP_Latex_Template.tex
\documentclass{ELSP}

\PassOptionsToPackage{unicode}{hyperref}
\PassOptionsToPackage{hyphens}{url}

\usepackage[utf8]{inputenc}

\IfFileExists{upquote.sty}{\usepackage{upquote}}{}
\IfFileExists{microtype.sty}{
    \usepackage{microtype}
    \UseMicrotypeSet[protrusion]{basicmath}
}{}

\usepackage{xcolor}
\IfFileExists{xurl.sty}{\usepackage{xurl}}{}
\IfFileExists{bookmark.sty}
    {\usepackage{bookmark}}
    {\usepackage{hyperref}}

\hypersetup{
    pdfcreator={ELSP},
    colorlinks=false,
    pdfborder={0 0 0},
    hidelinks
}
\usepackage{graphicx}
\usepackage{xparse}
\usepackage{amsmath,amssymb}
\usepackage{times}
\usepackage{mathptmx}

\usepackage{booktabs}
\usepackage{array}
\usepackage{multirow}
\usepackage{makecell}
\usepackage{longtable}
\usepackage{adjustbox}
\usepackage{calc}
\usepackage{changepage}

\usepackage{caption}
\usepackage{subcaption}
\usepackage{float}
\usepackage{stfloats}

\usepackage{geometry}
\usepackage{enumitem}
\usepackage{ragged2e}
\usepackage{titlesec}
\usepackage{setspace}
\usepackage{fancyhdr}

\titleformat{\section}[hang]
    {\fontsize{12pt}{12pt}\selectfont
     \bfseries\color[RGB]{0,131,255}}
    {\thesection}{0.5em}{}

\titleformat{\subsection}[hang]
    {\fontsize{12pt}{12pt}\selectfont\itshape}
    {\thesubsection}{0.5em}{}

\titleformat{\subsubsection}[hang]
    {\fontsize{12pt}{12pt}\selectfont}
    {\thesubsubsection}{0.5em}{}

\usepackage[numbers,sort,compress]{natbib}

\setcitestyle{
    numbers,
    square,
    comma,
    sort&compress
}

\newlength{\mylen}
\setcitestyle{citesep={,\kern-\mylen}}

\fancypagestyle{firstpage}{
    \fancyhf{}
    \setlength{\headsep}{6pt}

    \fancyhead[L]{%
        \sffamily\fontsize{9pt}{8pt}\selectfont
        \textbf{\textcolor[RGB]{0,131,255}{ELSP}}%
    }

    \fancyhead[R]{%
        \sffamily\fontsize{9pt}{8pt}\selectfont
        \textbf{\textcolor[RGB]{0,131,255}{Article}}%
    }

    \fancyfoot[L]{%
        \sffamily\fontsize{9pt}{8pt}\selectfont
        Liu et al. | ELSP%
    }
}

\let\ELSPOriginalIncludeGraphics\includegraphics

\RenewDocumentCommand{\includegraphics}{s O{} m}{%
    \IfFileExists{#3}{%
        \IfBooleanTF{#1}
            {\ELSPOriginalIncludeGraphics*[#2]{#3}}
            {\ELSPOriginalIncludeGraphics[#2]{#3}}%
    }{%
        \PackageWarning{ELSP-conversion}{Missing source image: #3}%
        \begingroup
        \setlength{\fboxsep}{2pt}%
        \fbox{%
            \parbox[c][1.65cm][c]
            {\dimexpr\linewidth-2\fboxsep-2\fboxrule\relax}{%
                \centering
                \scriptsize
                Missing source image\\[-1pt]
                \nolinkurl{#3}%
            }%
        }%
        \endgroup
    }%
}

\begin{document}

\thispagestyle{firstpage}

\let\thefootnote\relax
\footnotetext{%
    \newline
    \vspace{12pt}
    \raisebox{-\height}[0pt][0pt]{%
        \begin{minipage}[h]{\linewidth}

            \begin{minipage}[h]{0.15\linewidth}
                \includegraphics[
                    width=\linewidth,
                    height=1cm
                ]{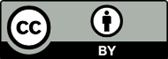}
            \end{minipage}
            \hfill
            \begin{minipage}[h]{0.82\linewidth}
                \justifying
                \footnotesize
                Copyright by the authors. Published by ELSP.
                This work is licensed under the Creative Commons
                Attribution 4.0 International License, which permits
                unrestricted use, distribution, and reproduction in
                any medium provided the original work is properly cited.
            \end{minipage}

        \end{minipage}
    }
}

\setstretch{1.24}

\begin{flushleft}

{\sffamily\small\noindent Article}

\vspace{8pt}

{\raggedright
\papertitle{IT-TextFusion: Iterative Text-Image Interaction with Text-Guided Residual Refinement for Degradation-Aware Image Fusion%
\hfill
\href{https://crossmark.crossref.org/dialog/?doi=10.55092/XXXX}{%
\raisebox{-0.2em}{\includegraphics[width=0.93cm,height=0.93cm]{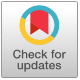}}%
}}
}

\vspace{12pt}

\authorname{Siyang}{Liu}{2}\textbf{,}
\authorname{Peiyi}{Zhou}{2}\textbf{,}
\authorname{Tianle}{Jin}{2}\textbf{,}
\authorname{Rongrong}{Bian}{2}\textbf{,}
\authorname{Zheke}{Jin}{2}\textbf{, and}
\authorname{Mengze}{Gao}{1*}\textbf{}

\vspace{12pt}

\formatintroduction{1}
{School of Automation, Southeast University, Nanjing, China}

\formatintroduction{2}
{Chair of Robotics, Artificial Intelligence and Real-time Systems,
Technical University of Munich, Munich, Germany}

\vspace{6pt}

\noindent * Corresponding author: Mengze Gao.

\end{flushleft}

\vspace{12pt}

\noindent
\textbf{\textcolor[RGB]{0,131,255}{Highlights:}}

\vspace{12pt}

\begin{itemize}[
    left=0pt,
    labelwidth=0pt,
    labelsep=17pt,
    itemsep=0pt,
    parsep=0pt,
    topsep=0pt
]
    \item Hierarchical text conditioning guides fusion across multiple
    decoder stages.

    \item Degradation-aware prompts provide scenario-specific global
    semantic guidance.

    \item Cross-Gate Fusion adaptively aggregates visible and infrared
    features at multiple hierarchical scales.

    \item Text-guided residual refinement improves high-resolution
    feature reconstruction under degradations.
\end{itemize}

\vspace{10pt}

\noindent
\textbf{\textcolor[RGB]{0,131,255}{Abstract:}}
Text-guided image fusion has recently emerged as an effective paradigm
for integrating multi-modal information while enabling flexible and
task-oriented fusion control. However, existing text-guided fusion
methods often rely on shallow semantic-visual interaction and limited
attention mechanisms, which restrict their ability to robustly handle
complex degradations and fully exploit textual guidance. In this paper,
we propose an iterative text-guided image fusion framework that
incorporates text-conditioned feature interaction across multiple
fusion and refinement stages. The proposed method integrates
deepest-level Cross-Attention, multi-scale Cross-Gate Fusion, and
stage-specific text-conditioned modulation, allowing the global text
embedding to condition hierarchical feature fusion and residual
refinement. By repeatedly injecting the pooled text embedding across
hierarchical decoder and refinement stages, the proposed framework
provides degradation-aware global semantic conditioning while
preserving complementary information from the visible and infrared
modalities. Experiments on several benchmark datasets show that the
proposed method improves several information-preservation and
perceptual-quality metrics, while exhibiting metric-dependent
trade-offs on some datasets.

\vspace{12pt}

\noindent
\textbf{\textcolor[RGB]{0,131,255}{Keywords:}}
infrared-visible image fusion; text-guided image fusion;
iterative text-image interaction; degradation-aware fusion;
text-guided residual refinement

\input{sections/1_Introduction}
\input{sections/2_RelatedWork}
\input{sections/3_Method}
\input{sections/4_Experiments}

\input{sections/5_Conclusion}

\input{sections/Acknowledgements}

\bibliography{ref}

\end{document}

%% file: sections/1_Introduction.tex
\section{Introduction}

In recent years, autonomous driving has attracted increasing attention due to rapid progress in sensing, learning, and system integration technologies. Autonomous driving systems aim to reduce driver workload, improve road safety, and mitigate traffic congestion. Among the three core components of autonomous driving systems--perception, prediction, and motion planning--perception plays a fundamental role by providing accurate environmental understanding to support reliable decision-making in advanced driver-assistance systems (ADAS). Core perception tasks such as semantic segmentation~\cite{cao2024sdpt} and object detection~\cite{cao2021fusion} rely heavily on visual sensing, with cameras remaining the most widely deployed sensors in autonomous vehicles due to their rich spatial and appearance information ~\cite{huang2022multi,zhang2023cmx}. Cao et al.~\cite{11246605} further investigate fisheye object detection for autonomous driving, introducing a feature-aligned pyramid module and a location-aligned detection head.

Frame-based RGB images generally provide high visual fidelity, and recent advances in deep learning and end-to-end training have enabled strong perception performance under favorable conditions. However, perception systems based on a single sensing modality inherently suffer from limitations in robustness and reliability when operating in challenging environments. Conventional cameras are particularly vulnerable to adverse conditions such as low illumination, overexposure, motion blur, and sensor noise. Furthermore, complex traffic scenes frequently contain occlusions and
large illumination variations, which further reduce the reliability
of camera-based perception. These challenges significantly degrade the performance of downstream perception tasks, including object detection and semantic segmentation, highlighting the necessity for more robust sensing strategies. As a result, multi-modal sensor fusion has emerged as a promising approach to alleviate these limitations by leveraging complementary information from different sensing modalities ~\cite{cao2021fusion,huang2022multi,zhang2024e2e,cui2024survey}.

\begin{figure}[t]
    \centering
    \includegraphics[width=\linewidth]{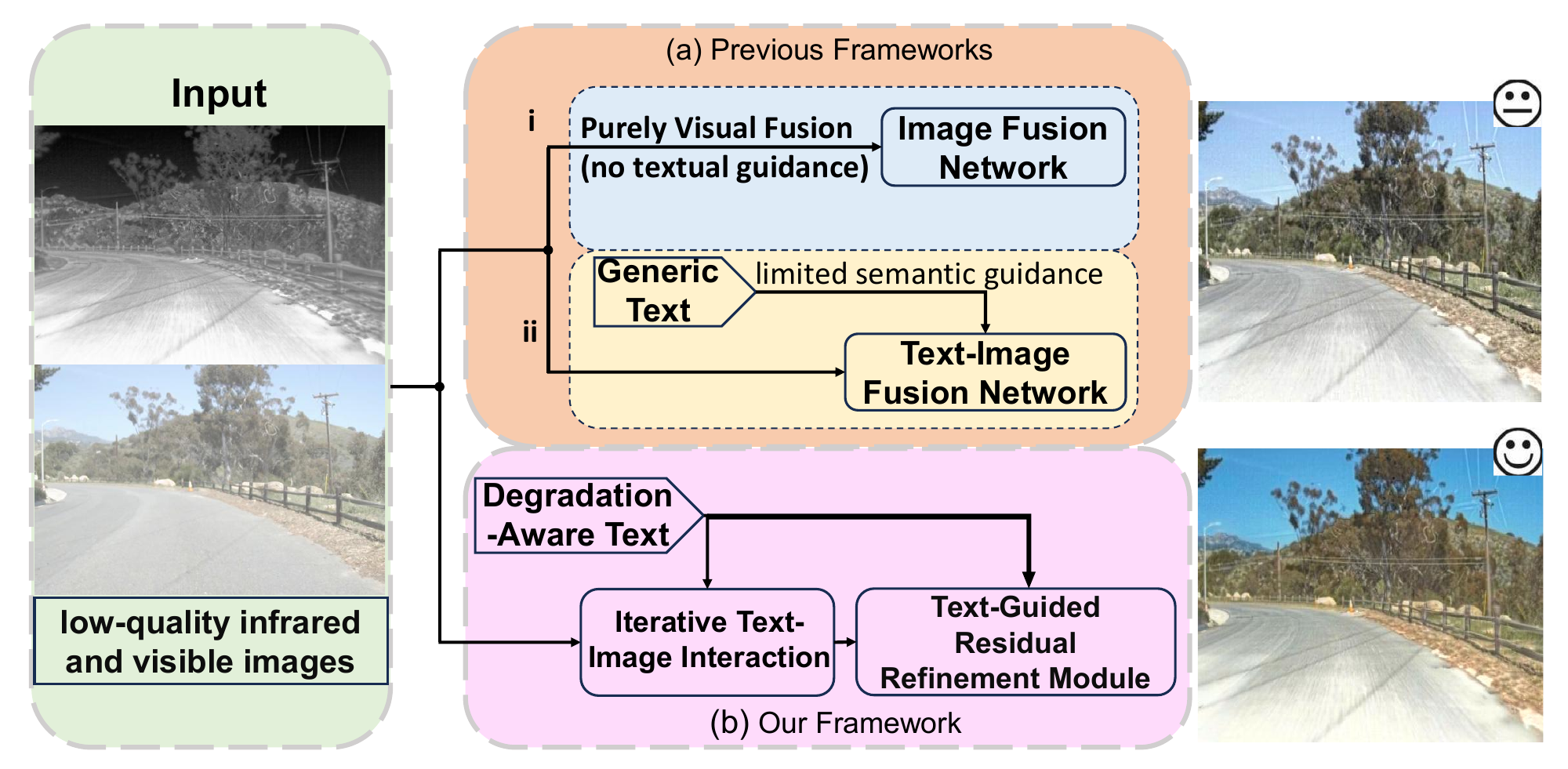}
   \caption{Comparison of image fusion frameworks. (a) Previous frameworks:
(i) purely visual fusion without textual guidance; and (ii) text-image
fusion using generic text that provides only limited semantic guidance.
(b) The proposed framework, in which degradation-aware text guides both
iterative text-image interaction and the subsequent text-guided residual
refinement.}\label{fig:text_if_overview}
\end{figure}

While conventional RGB cameras remain strong standalone sensors, combining multiple sensing modalities enables a more comprehensive and resilient representation of the environment. Depth sensors provide geometric structure, event cameras excel in high-speed and high-dynamic-range scenarios ~\cite{cao2021fusion,zhang2023cmx,cao2024embracing}, and LiDAR systems offer accurate spatial and angular measurements ~\cite{zhang2023cmx}. Infrared cameras, in particular, capture thermal radiation and are highly effective for nighttime perception and for detecting heat-emitting objects such as pedestrians and vehicles ~\cite{yi2024text}. Although some approaches explore fusion across multiple modalities simultaneously ~\cite{zhang2023delivering}, this work focuses on the fusion of visible and infrared imagery, which offers strong complementarity for safety-critical perception tasks.

A central challenge in multi-modal perception lies in determining how to effectively integrate heterogeneous sensor information, a problem for which no trivial solution exists ~\cite{chen2022multimodal}. Existing fusion strategies are typically categorized into early, middle, and late fusion. Early fusion combines raw sensor data at the input level but is highly sensitive to noise. Late fusion integrates modality-specific outputs at the decision level, often through algebraic or probabilistic methods, but fails to capture fine-grained cross-modal interactions at lower feature levels ~\cite{chen2022multimodal}. Middle fusion, which merges modality-specific features within intermediate network layers, has therefore become a widely adopted strategy. For example, NeuroGrasp~\cite{9787342} combines RGB and event information in a multi-modal neural network for robotic grasp pose estimation. Nonetheless, effectively exploiting informative signals from multiple modalities while suppressing noise and inconsistencies remains a fundamental challenge in multi-modal fusion. In this work, we adopt a middle-fusion strategy and aim to address this challenge by leveraging semantic guidance from textual prompts under complex degradation conditions.

Recent progress in vision-language models has highlighted the effectiveness of leveraging textual semantics to guide visual representation learning. Large-scale vision-language models trained on diverse image-text pairs exhibit strong cross-modal alignment and semantic generalization capabilities, providing transferable semantic representations for downstream visual
tasks~\cite{yi2024text}. In particular, CLIP~\cite{radford2021learning} demonstrates remarkable robustness and transferability, providing rich semantic embeddings that have been successfully adopted in various vision applications, including image fusion~\cite{yang2024mgfusion}.
Building upon these advances, recent text-guided fusion frameworks such as Text-IF~\cite{yi2024text}, TextFusion~\cite{cheng2025textfusion}, and TITFormer~\cite{li2025titformer} incorporate linguistic descriptions into the fusion pipeline, allowing semantic prompts to regulate feature aggregation and refinement. These methods demonstrate that textual representations can provide
additional semantic conditions for feature aggregation and refinement beyond conventional purely visual fusion strategies.

Nevertheless, purely visual fusion remains dominant in many earlier approaches. As illustrated in Fig.~\ref{fig:text_if_overview}, representative methods such as~\cite{xu2020u2fusion,zhao2023metafusion,zhao2023ddfm} rely exclusively on image-based fusion mechanisms without leveraging auxiliary semantic information. While effective under relatively clean imaging conditions, their robustness is often limited when facing complex real-world degradations, including low illumination, reduced contrast, and sensor noise.
Despite the encouraging developments in text-guided fusion, several challenges remain insufficiently addressed.
Textual information is commonly introduced as a high-level conditional signal, without explicitly modeling fine-grained semantic structures or degradation characteristics. Consequently, the interaction between linguistic cues and visual features often remains relatively shallow, restricting the model's ability to exploit textual guidance across different feature resolutions.
Moreover, many existing approaches rely on limited cross-modal interaction, where text features influence image representations through a single modulation stage or simple affine transformations. Such designs may restrict the effect of the global text condition on
hierarchical feature fusion, particularly when different degradations
affect visual information at different feature resolutions.
In addition, although refinement modules are frequently employed, semantic guidance is not always propagated consistently throughout the correction stages. This may lead to incomplete suppression of residual artifacts under severe or spatially heterogeneous degradations.

To overcome these limitations, we propose a unified text-guided fusion framework
that strengthens hierarchical text-conditioned feature fusion and residual
refinement. The proposed
approach improves semantic specificity through degradation-aware text engineering, which generates more informative and task-relevant linguistic guidance. In addition, the Cross-Gate Fusion module is introduced to regulate inter-modal feature aggregation by adaptively weighting the reliability of visual responses under degradation.
To maintain textual conditioning across hierarchical scales, we design
an Iterative Text-Image Interaction Module \textbf{(ITIM)}, which repeatedly
injects the pooled global text embedding into decoder features
through independently parameterized stage-specific modulation
functions.
Furthermore, a Text-Guided Residual Refinement Module \textbf{(TG-RRM)} is
developed to apply global text-conditioned modulation within the
multi-scale refinement process, supporting high-resolution residual
feature correction before final reconstruction.
Overall, our contributions can be summarized as follows:
\begin{itemize}
    \item We improve two complementary aspects of the fusion pipeline: degradation-aware text engineering increases the semantic specificity of textual guidance, while the Cross-Gate Fusion module improves multi-scale feature aggregation by adaptively suppressing unreliable visual responses.
    \item We propose \textbf{ITIM}, which incorporates textual guidance into visual features across multiple fusion stages, providing hierarchical global text conditioning throughout the fusion process.
    \item We propose \textbf{TG-RRM}, which integrates text-conditioned modulation into the refinement stage to enhance structural consistency and detail preservation.
    \item In addition, we evaluate the proposed method on an extended EMS dataset covering nine representative degradation types. Experimental results show improvements on selected fusion and no-reference perceptual-quality metrics, together with trade-offs in MI, $\mathrm{SSIM}_{\mathrm{sum}}$, and CLIP-IQA.

\end{itemize}

%% file: sections/2_RelatedWork.tex
\section{Related Work}

\subsection{General Image Fusion Methods}

With the success of deep convolutional networks, encoder-decoder architectures
and residual learning have become the dominant paradigms for image fusion
tasks~\cite{long2015fully, ronneberger2015u}.
Owing to skip connections and parallel modality-specific backbones, deep fusion
models are able to effectively preserve complementary information from different
sources~\cite{chen2022multimodal, wang2024terf,cao2024embracing}.
More recently, transformer-based architectures have been introduced into image
fusion to model long-range dependencies via self-attention~\cite{vaswani2017attention}.
Vision Transformer (ViT)~\cite{dosovitskiy2020image} represents images as sequences
of patch tokens, enabling global context modeling that has proven beneficial for
fusion tasks~\cite{cao2023ghostvit,zhang2023delivering, yi2024text, li2025titformer}.
CMX~\cite{zhang2023cmx} proposes a cross-modal feature rectification module to refine
RGB and arbitrary modality features before fusion, followed by cross-attention-based
feature aggregation. Building upon CMX, CMNeXt~\cite{zhang2023delivering} extends this framework
to support an arbitrary number of modalities via dynamic feature selection and
parallel pooling. 

Alternative fusion strategies have also been explored. Adaptive Instance Normalization (AdaIN)~\cite{huang2017arbitrary} provides an efficient feature-statistics alignment operation and has inspired lightweight modulation strategies in multi-modal representation learning. In contrast, Bayesian fusion
approaches~\cite{chen2022multimodal} perform late fusion at the detection level by combining
independently trained detectors through probabilistic inference, improving robustness
under modality misalignment. Closely related in
employing feedback mechanisms, the E2E-MFD model ~\cite{zhang2024e2e} utilizes an Object-Region-Pixel Phylogenetic Tree (ORPPT) for hierarchical, coarse-to-fine fusion. Finally, Text-IF~\cite{yi2024text} incorporates semantic text
guidance into image fusion, enabling interactive and degradation-aware fusion
through transformer-based cross-attention and convolutional decoding.

\subsection{Text-Image Models}

Transformer-based representation learning has significantly advanced multi-modal
understanding by enabling joint modeling of visual and textual data. A major
milestone in this direction is CLIP~\cite{radford2021learning}, which aligns images
and text in a shared embedding space through large-scale contrastive pretraining,
supporting zero-shot recognition and semantic retrieval.
Building upon CLIP, DenseCLIP~\cite{rao2022denseclip} extends vision-language
pretraining to dense prediction tasks by introducing pixel-text similarity
maps and context-aware prompting, improving performance in semantic segmentation
and detection. Diffusion-based models such as Stable Diffusion~\cite{rombach2022high}
further demonstrate the effectiveness of textual conditioning for fine-grained
visual control by integrating text encoders with generative architectures.

Motivated by these advances, incorporating textual semantics into image fusion
has emerged as a natural extension. TextFusion~\cite{cheng2025textfusion} introduces
affine text-guided modulation within a transformer-based fusion framework, enabling
controllable fusion outcomes. TeRF~\cite{wang2024terf} further explores region-aware
fusion by integrating language commands into segmentation-guided pipelines. TITFormer~\cite{li2025titformer} combines textual inputs with simulated infrared imagery
for image enhancement, leveraging cross-modal attention to extract contextual
features from text. Similarly, MGFusion~\cite{yang2024mgfusion} injects semantic
information through CLIP-guided feature modulation, enhancing multi-modal
representations without complex architectural designs. Text-IF~\cite{yi2024text}
represents a unified framework that dynamically adapts fusion strategies based
on semantic text inputs.

Existing text-guided fusion methods mainly differ in the location,
representation, and frequency of textual conditioning. TextFusion
primarily employs text-conditioned feature modulation, TeRF emphasizes
region-aware language control, MGFusion uses CLIP-derived semantic
features to modulate multi-modal representations, and Text-IF adapts
fusion to degradation descriptions. In contrast, IT-TextFusion
maintains global text conditioning across hierarchical
decoding and residual feature refinement. The proposed method contributes a stage-specific textual conditioning framework that integrates multi-scale visual fusion with high-resolution residual refinement.

%% file: sections/3_Method.tex
\section{Method}
\begin{figure}[!t]
    \centering
    \includegraphics[width=\textwidth]{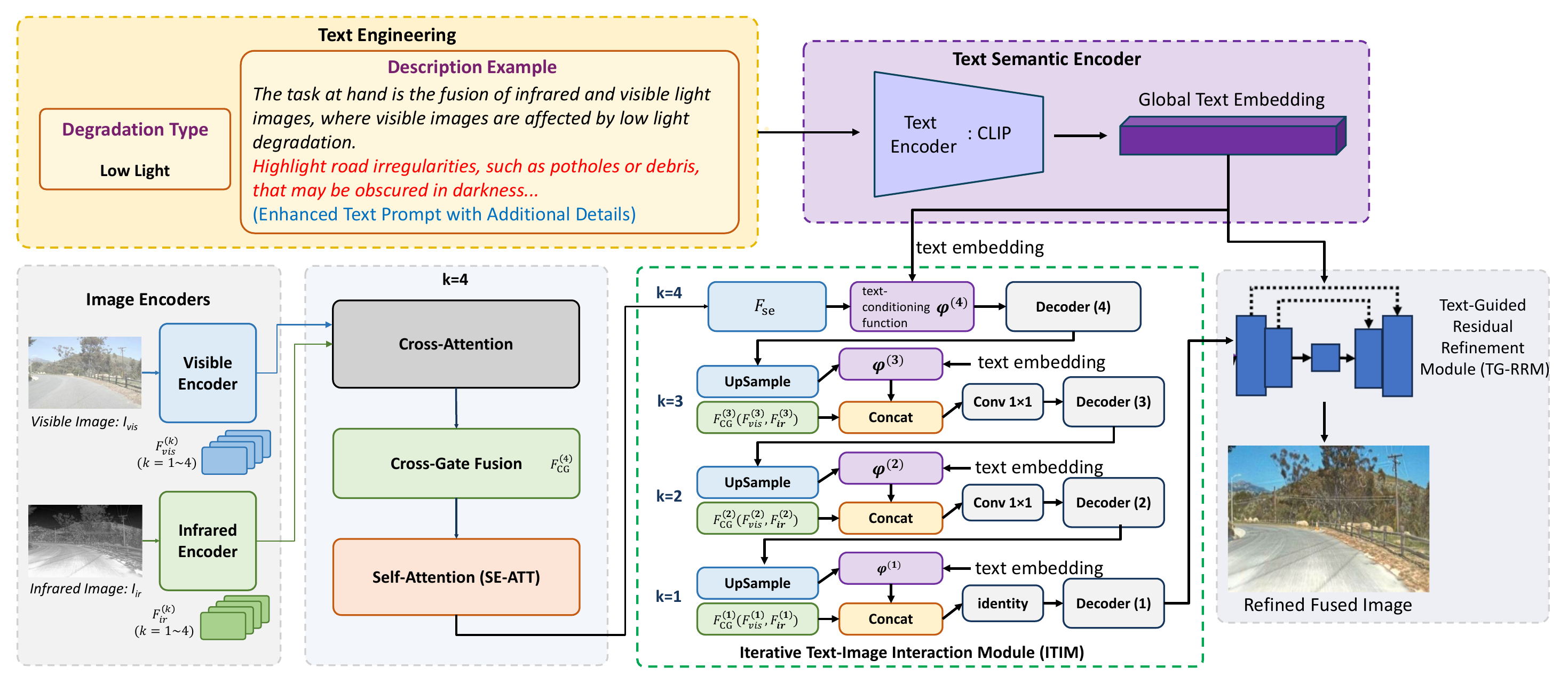}
    \caption{Overall workflow of the proposed method. It consists of two main parts: a text semantic encoder that extracts a global text embedding from the detailed description produced by Text Engineering, and an image fusion pipeline comprising Cross-Attention, Cross-Gate Fusion, Self-Attention (SE-ATT), Iterative Text-Image Interaction Module (ITIM), and Text-Guided Residual Refinement Module (TG-RRM).}
    \label{fig:framework}
\end{figure}

This section presents the overall workflow of our framework, as illustrated in Fig.~\ref{fig:framework}. We first introduce Text Engineering,
which constructs degradation-aware prompts to provide semantic guidance
for image fusion. We then describe the Image Fusion Pipeline,
including the image encoders, Cross-Attention, Cross-Gate Fusion, and the Iterative Text-Image Interaction Module
(ITIM). Next, the Text Semantic Encoding and Feature Modulation subsection
explains how the global text embedding is encoded and used to
condition visual features at each hierarchical stage. We further
present the Text-Guided Residual Refinement Module (TG-RRM),
which performs text-conditioned residual feature correction before
image reconstruction. Finally, the Loss Functions subsection describes the composition of the
overall training objective.

\subsection{Text Engineering}
\label{subsec:prompt_engineering}

Textual guidance has recently emerged as an effective mechanism for regulating
multi-modal fusion. Nevertheless, many existing text-guided fusion frameworks rely on coarse or generic task descriptions that primarily express high-level fusion objectives while failing to explicitly identify the affected modality,
degradation category, or scenario-specific semantic priority. Such generic descriptions may provide insufficient
degradation-specific conditioning because the same or similar prompt
can be used under substantially different input conditions.
Consequently, prompts that do not explicitly distinguish the affected
modality and degradation type may produce similar global text
embeddings, limiting their ability to differentiate different
fusion scenarios.

To address this issue, we manually construct degradation-aware prompt
sets offline to enrich the semantic content of the textual conditions. Instead of
adopting minimal task descriptions, the prompts are designed to jointly encode
the fusion objective, dominant degradation characteristics, and
application-relevant semantic priorities. Since the prompts are organized into predefined category-specific
sets, different textual conditions can be selected for different
degradation scenarios without changing the fusion-network
architecture.
Concretely, beyond describing the fusion task itself, the textual guidance is
augmented with manually specified semantic emphasis that varies according to the
target scenario. In safety-critical applications such as autonomous driving,
objects including vehicles, pedestrians, and road structures must remain
clearly distinguishable despite adverse imaging conditions. By explicitly
encoding such priorities, the textual input conveys not only what degradation is present but also which scene elements should receive greater attention during fusion.
As an illustrative example, the basic task description \emph{``The task at hand is the fusion of infrared and visible light images, where visible images are affected by low light degradation''}
can be extended with additional semantic guidance such as
\emph{``... Highlight road irregularities, such as potholes or debris, that may be obscured in darkness.''}
The resulting prompt therefore provides a predefined global text-conditioning signal containing more specific degradation information than a
generic fusion-task description.

By exposing the model to degradation-aware and semantically enriched
prompts, the fusion network receives category-specific global text-conditioning signals during training. The selected prompt is encoded by the
frozen CLIP text encoder and used to condition hierarchical feature
fusion and residual refinement, while the fusion-network architecture
remains unchanged.

\subsection{Image Fusion Pipeline}
\label{subsec:image fusion pipeline}

\textbf{Image Encoders.} The visible and infrared image encoders independently process the source visible and infrared images as inputs.
To effectively capture both spatial structures and high-level semantic information, we employ Transformer-based blocks~\cite{zamir2022restormer} as the backbone feature extractor.
This design enables the encoder to learn comprehensive and discriminative representations by jointly modeling local details and long-range dependencies across different feature scales. The encoding process is formulated as follows:
\begin{equation}
F_{\text{vis}} = \mathcal{F}^\mathrm{e}_\mathrm{v}(I_{\text{vis}}), \quad
F_{\text{ir}} = \mathcal{F}^\mathrm{e}_\mathrm{i}(I_{\text{ir}}),
\end{equation}
where $I_{\text{vis}} \in \mathbb{R}^{H \times W \times 3}$ and
$I_{\text{ir}} \in \mathbb{R}^{H \times W \times 3}$ denote the visible and infrared network inputs, respectively. In the implementation, the infrared image is loaded in three-channel RGB format before being passed to the encoder.
$H$ and $W$ represent the height and width of the input image.
$\mathcal{F}^\mathrm{e}_\mathrm{v}$ and $\mathcal{F}^\mathrm{e}_\mathrm{i}$ denote the visible encoder and infrared encoder. $F_{\mathrm{vis}}$ and $F_{\mathrm{ir}}$ collectively denote the
four-level visible and infrared feature pyramids, respectively:
$F_{\mathrm{vis}}=\{F_{\mathrm{vis}}^{(k)}\}_{k=1}^{4}$ and
$F_{\mathrm{ir}}=\{F_{\mathrm{ir}}^{(k)}\}_{k=1}^{4}$.
Here, $F_{m}^{(k)}$ denotes the feature of modality
$m\in\{\mathrm{vis},\mathrm{ir}\}$ at the $k$-th encoder level.
In our framework, multi-level features are extracted at four different stages of the encoder, corresponding to hierarchical representations at progressively reduced spatial resolutions.
These multi-scale features are later utilized as inputs to different stages of the fusion and decoding modules.

\textbf{Cross-Attention.}
The Cross-Attention layer performs bidirectional cross-modal interaction at the deepest encoder level by using cross-modal affinities to aggregate modality-specific value features. Before generating the query, key, and value
representations, the two features are independently normalized using
16-group GroupNorm. The normalized features are then projected by
modality-specific $1\times1$ convolutions:
\begin{equation}
\begin{aligned}
[Q_m,K_m,V_m]
=
\operatorname{Reshape}_{h}
\left(
\operatorname{Split}_{3}
\left[
\operatorname{Conv}_{\mathrm{qkv},m}^{1\times1}
\left(
\operatorname{GN}_{m}
\left(
F_m^{(4)}
\right)
\right)
\right]
\right),\quad
m\in\{\mathrm{vis},\mathrm{ir}\},
\end{aligned}
\label{eq:image_cross_qkv}
\end{equation}
where
$Q_m,K_m,V_m
\in
\mathbb{R}^{B\times h\times d\times N_4}$,
$N_4=H_4W_4$, and $d=C_4/h$. In the specific implementation,
$h=1$ and $d=C_4$.

Subsequently, bidirectional cross-modal interaction is performed. In the visible branch, the infrared query is matched against the visible keys and values, and the attended response is projected and added back to the visible feature. Symmetrically, the visible query is used to retrieve information from the infrared keys and values for the infrared branch. The implemented Cross-Attention can therefore be written as
\begin{equation}
\begin{aligned}
F_{\mathrm{v}}^{\mathrm{ca}}
&=
F_{\mathrm{vis}}^{(4)}
+
\mathcal{P}_{\mathrm{v}}
\left[
\operatorname{Softmax}
\left(
\frac{
Q_{\mathrm{i}}K_{\mathrm{v}}^{\top}
}{
\sqrt{d}
}
\right)
V_{\mathrm{v}}
\right],\quad
F_{\mathrm{i}}^{\mathrm{ca}}
=
F_{\mathrm{ir}}^{(4)}
+
\mathcal{P}_{\mathrm{i}}
\left[
\operatorname{Softmax}
\left(
\frac{
Q_{\mathrm{v}}K_{\mathrm{i}}^{\top}
}{
\sqrt{d}
}
\right)
V_{\mathrm{i}}
\right],
\end{aligned}
\label{eq:image_cross_attention}
\end{equation}
where $\sqrt{d}$ is the scaling factor and $\mathcal{P}_\mathrm{v}(\cdot)$ and $\mathcal{P}_\mathrm{i}(\cdot)$ denote the modality-specific output projections. The single-head attention responses are reshaped into spatial feature
maps, passed through modality-specific $1\times1$ output projections,
and added to the original unnormalized level-4 encoder features through
residual connections.

\begin{figure}[t]
    \centering
    \includegraphics[width=\columnwidth]{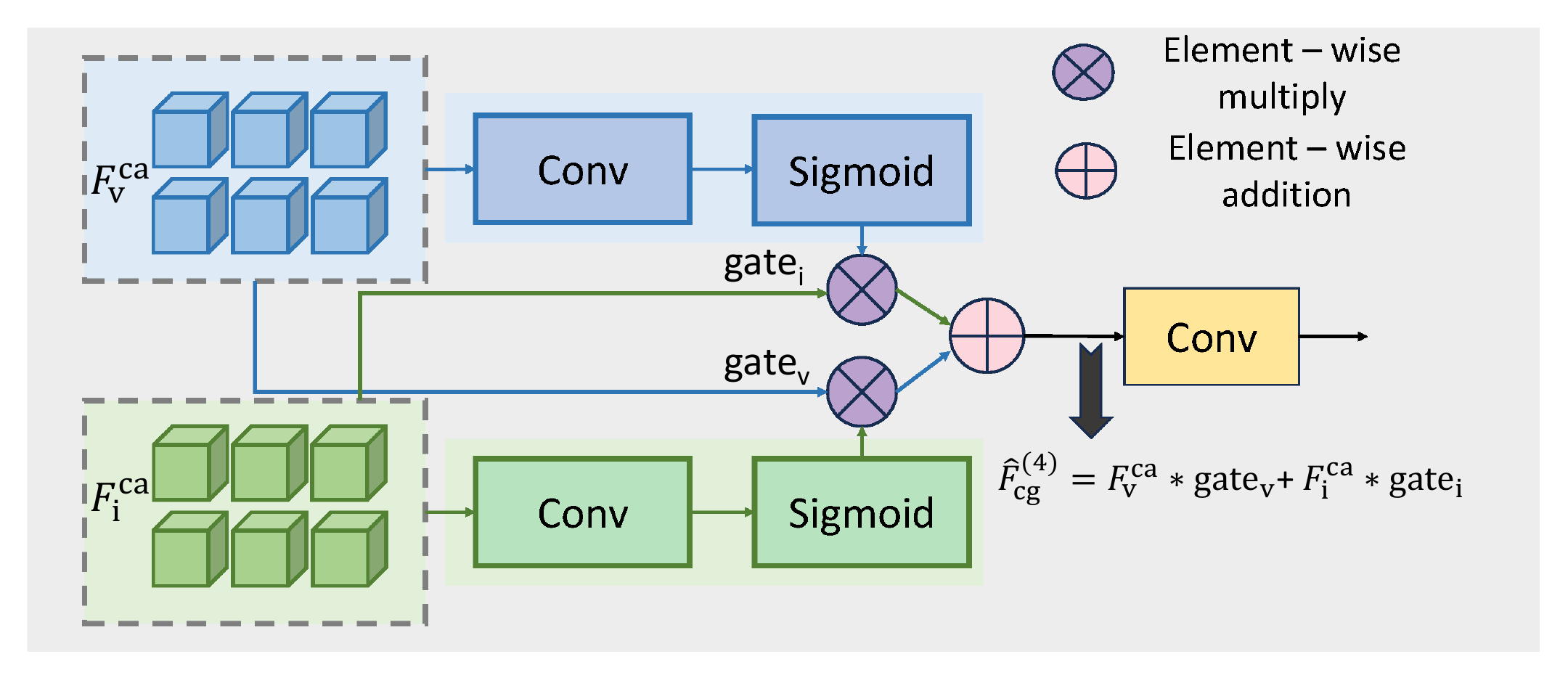}
    \caption{Architecture of the Cross-Gate Fusion module, illustrated using its application at the deepest level ($k=4$). Cross-attended visible and infrared features generate cross-modal gates to regulate the opposite modality, after which the cross-gated features are added and projected through a convolution.}
    \label{fig:cross_gate}
\end{figure}

\textbf{Cross-Gate Fusion.} While Cross-Attention enables effective information exchange between visible and infrared features, it does not explicitly regulate the reliability of the transferred information.
Therefore, a Cross-Gate Fusion module is introduced to adaptively modulate the cross-attended features, allowing the network to selectively preserve informative cues and suppress degraded or irrelevant responses. At the deepest level, the Cross-Gate Fusion module takes the
cross-attended features as inputs and can be expressed as
$F_\mathrm{cg}^{(4)}=\mathcal{F}_\mathrm{{CG}}^{(4)}
(F_\mathrm{v}^\mathrm{{ca}},F_\mathrm{i}^\mathrm{{ca}})$,
where $\mathcal{F}_\mathrm{{CG}}^{(4)}(\cdot,\cdot)$ denotes the
Cross-Gate Fusion operation at the deepest level.
The application of Cross-Gate Fusion at the finer hierarchical
levels is described in the subsequent ITIM formulation. The detailed architecture is illustrated in Fig.~\ref{fig:cross_gate}. First, we calculate the gating weights for the two image feature maps from the Cross-Attention layer:
\begin{equation}
\begin{aligned}
\mathrm{gate}_{\mathrm{i}}
&=
\sigma
\left(
\operatorname{Conv}_{\mathrm{i}}^{1\times1}
\left(
F_{\mathrm{v}}^{\mathrm{ca}}
\right)
\right),\quad
\mathrm{gate}_{\mathrm{v}}
=
\sigma
\left(
\operatorname{Conv}_{\mathrm{v}}^{1\times1}
\left(
F_{\mathrm{i}}^{\mathrm{ca}}
\right)
\right),
\end{aligned}
\label{eq:cross_gate_weights}
\end{equation}where $\sigma$ denotes the sigmoid function. The learned gating weights are then used to adaptively regulate the contribution of each modality at every spatial location, enabling selective preservation of informative features while suppressing less reliable responses, as formulated below:
\begin{equation}
\widehat{F}_{\mathrm{cg}}^{(4)}
=
F_{\mathrm{v}}^{\mathrm{ca}}
\odot
\mathrm{gate}_{\mathrm{v}}
+
F_{\mathrm{i}}^{\mathrm{ca}}
\odot
\mathrm{gate}_{\mathrm{i}},
\label{eq:cross_gate_fusion}
\end{equation}where $\odot$ denotes the Hadamard product. For the final output, an additional convolutional layer is introduced to further refine the fused features and enhance their representation capability: $F_\mathrm{cg}^{(4)}
=\operatorname{Conv}_{1\times1}^{(4)}
(\widehat{F}_\mathrm{cg}^{(4)})$.

\textbf{Iterative Text-Image Interaction Module (ITIM).} The deepest-level Cross-Gate Fusion output is first processed by a
single-head Self-Attention (SE-ATT) module. Before attention, the
feature is normalized using 16-group GroupNorm, and a
$1\times1$ convolution generates the query, key, and value
representations. The attention response is output-projected and combined with a residual connection:

\begin{equation}
[Q_\mathrm{cg},K_\mathrm{cg},V_\mathrm{cg}]
=
\operatorname{Tokenize}
\left(
\operatorname{Split}_{3}
\left[
\operatorname{Conv}_\mathrm{qkv}^{1\times1}
\left(
\operatorname{GN}
\left(
F_\mathrm{cg}^{(4)}
\right)
\right)
\right]
\right), \\
F_\mathrm{se}=F_\mathrm{cg}^{(4)}+\mathcal{P}_\mathrm{se}\!\left[
\operatorname{Softmax}
\left(
\frac{Q_\mathrm{cg}K_\mathrm{cg}^{\top}}{\sqrt{d}}
\right)V_\mathrm{cg}\right],
\end{equation}
where $Q_\mathrm{cg}$, $K_\mathrm{cg}$, and $V_\mathrm{cg}$ denote the query, key, and value projections of $F_\mathrm{cg}^{(4)}$, respectively, and $\mathcal{P}_\mathrm{se}(\cdot)$ denotes the output projection. 
After this early fusion stage, the framework enters the phase where textual information actively participates in the fusion process.

Before applying text-conditioned feature modulation, the
global text embedding is projected into a stage-specific
semantic conditioning vector. At each hierarchical stage, the
visual feature is represented as spatial tokens, whereas the
single global text embedding is projected to provide the
semantic conditioning signal. The resulting text-conditioned
representation is subsequently mapped to affine modulation
parameters that regulate the corresponding visual features.

Since a single global sentence-level text embedding is used
rather than a sequence of token-level text features, the
operation provides global semantic conditioning at each
hierarchical stage instead of token-wise language-vision
attention. The main role of this design is therefore to
repeatedly inject the prompt semantics into visual
representations throughout hierarchical decoding. The
stage-specific conditioning parameters allow the same textual
guidance to affect features at different semantic resolutions,
thereby enabling progressive text-conditioned feature
refinement.

At the deepest encoder level ($k=4$), the visible and infrared
features first undergo the Cross-Attention
described above, followed by Cross-Gate Fusion and Self-Attention (SE-ATT). The resulting fused feature $F_{\mathrm{se}}$ is then
conditioned on the global text embedding through the stage-specific
text-conditioning function $\varphi^{(4)}(\cdot,\cdot)$ and processed
by a Transformer-based decoder block ~\cite{zamir2022restormer}:
\begin{equation}
F_{\mathrm{d}}^{(4)}
=
\mathcal{F}_{\mathrm{D}}^{(4)}
\left(
\varphi^{(4)}
\left(
F_{\mathrm{se}},
F_{\mathrm{text}}
\right)
\right).
\label{eq:itim_deepest}
\end{equation}

For the subsequent levels $k\in\{3,2,1\}$, the decoded feature from
the immediately coarser level is first upsampled to the current
spatial resolution and then modulated by the same global text
embedding. In parallel, the visible and infrared encoder features at
the corresponding level are reintroduced and fused using a
stage-specific Cross-Gate Fusion module. The text-conditioned decoder
feature and the cross-gated encoder feature are subsequently
concatenated along the channel dimension:
\begin{equation}
\begin{aligned}
\widetilde{F}_{\mathrm{u}}^{(k)}
&=
\varphi^{(k)}
\left(
\mathcal{U}^{(k)}
\left(
F_{\mathrm{d}}^{(k+1)}
\right),
F_{\mathrm{text}}
\right),\quad
F_{\mathrm{e}}^{(k)}
=
\mathcal{F}_{\mathrm{CG}}^{(k)}
\left(
F_{\mathrm{vis}}^{(k)},
F_{\mathrm{ir}}^{(k)}
\right),\\
F_{\mathrm{d}}^{(k)}
&=
\mathcal{F}_{\mathrm{D}}^{(k)}
\left(
\mathcal{R}^{(k)}
\left(
[\widetilde{F}_{\mathrm{u}}^{(k)};
F_{\mathrm{e}}^{(k)}]
\right)
\right),
\quad k=3,2,1.
\end{aligned}
\label{eq:itim_hierarchical}
\end{equation}
Here, $\mathcal{U}^{(k)}(\cdot)$ denotes the implemented upsampling
operator, $[\cdot;\cdot]$ denotes channel-wise concatenation, and
$\mathcal{R}^{(k)}(\cdot)$ denotes the channel-adjustment operation.
Specifically, $\mathcal{R}^{(3)}(\cdot)$ and
$\mathcal{R}^{(2)}(\cdot)$ are implemented using $1\times1$
convolutions, whereas $\mathcal{R}^{(1)}(\cdot)$ is the identity
mapping. Therefore, the concatenated level-1 feature is fed directly
to the final decoder block without an additional channel-reduction
convolution.

After the initial level-4 fusion, the encoder features reintroduced
during the remaining three interactions are the level-3, level-2,
and level-1 visible and infrared features, respectively. At the deepest level, Cross-Attention enables bidirectional high-level semantic interaction between the visible and infrared features. At the finer
levels, the framework combines an upsampled text-conditioned decoder
feature with a same-resolution cross-gated encoder feature. In this
way, the global text embedding is repeatedly injected into the
hierarchical decoding process while modality-specific visual
information is progressively recovered from the encoder branches.

\subsection{Text Semantic Encoding and Feature Modulation}
\label{subsec:text_interaction_archi}

This subsection describes the stage-specific text-conditioning
function $\varphi^{(k)}(\cdot,\cdot)$ used in
Eqs.~\eqref{eq:itim_deepest} and~\eqref{eq:itim_hierarchical}.
Specifically, the visual feature provides the query representation,
whereas the pooled CLIP text embedding provides the key and value
representations. The resulting text-conditioned context is then
projected to the visual channel dimension and used to generate
feature-wise affine modulation parameters.

\textbf{Text Semantic Encoder.}
The text semantic encoder transforms the input prompt into a compact
sentence-level text embedding. Benefiting from large-scale
vision-language pretraining, CLIP~\cite{radford2021learning}
provides text representations with strong semantic discrimination.
To maintain linguistic consistency and avoid overfitting the text
encoder to the fusion datasets, the CLIP text encoder is kept frozen
throughout training. The text feature extraction process is
formulated as
\begin{equation}
F_{\mathrm{text}}
=
\mathcal{F}_{\mathrm{clip}}
\left(
T_{\mathrm{text}}
\right),
\label{eq:clip_text_feature}
\end{equation}
where
$F_{\mathrm{text}}\in\mathbb{R}^{B\times512}$
denotes the pooled global text embedding returned by
the frozen CLIP ViT-B/32 text encoder. Only this single
512-dimensional embedding is retained for each prompt.

\textbf{Stage-Specific Visual-Text Conditioning.}
Let
$F_\mathrm{f}^{(k)}
\in
\mathbb{R}^{B\times C_k\times H_k\times W_k}$
denote the visual feature entering the text-conditioning module at
the $k$-th interaction stage, where $C_k$, $H_k$, and $W_k$ denote
its channel number, height, and width, respectively. The visual
feature is first flattened into $N_k=H_kW_k$ spatial tokens:
\begin{equation}
X_{\mathrm{f}}^{(k)}
=
\operatorname{reshape}
\left(
F_{\mathrm{f}}^{(k)}
\right)
\in
\mathbb{R}^{B\times N_k\times C_k}.
\label{eq:visual_tokenization}
\end{equation}

The visual tokens and the global text embedding are independently
normalized and projected to a common transformer width $D=512$:
\begin{equation}
\begin{aligned}
Z_{\mathrm{f}}^{(k)}
&=
W_{\mathrm{F}}^{(k)}
\operatorname{LN}_{\mathrm{F}}^{(k)}
\left(
X_{\mathrm{f}}^{(k)}
\right),\quad
Z_{\mathrm{text}}^{(k)}
=
W_{\mathrm{T}}^{(k)}
\operatorname{LN}_{\mathrm{T}}^{(k)}
\left(
F_{\mathrm{text}}
\right).
\end{aligned}
\label{eq:stage_projection}
\end{equation}
Here,
$Z_\mathrm{f}^{(k)}
\in
\mathbb{R}^{B\times N_k\times D}$.
The projected text embedding is regarded as a single text token,
such that
$Z_{\mathrm{text}}^{(k)}
\in
\mathbb{R}^{B\times1\times D}$.
All
projection parameters are independently parameterized at different
hierarchical stages.

The query representation is generated from the projected visual
tokens, whereas the key and value representations are generated from
the projected text token:
\begin{equation}
\begin{aligned}
Q^{(k)}
&=
W_\mathrm{Q}^{(k)}Z_\mathrm{f}^{(k)},\quad
K^{(k)}
=
W_\mathrm{K}^{(k)}Z_{\mathrm{text}}^{(k)},\quad
V^{(k)}
=
W_\mathrm{V}^{(k)}Z_{\mathrm{text}}^{(k)}.
\end{aligned}
\label{eq:text_condition_qkv}
\end{equation}

Multi-head attention is then computed using $Q^{(k)}$, $K^{(k)}$, and
$V^{(k)}$ to obtain the text-conditioned context
$C_{\mathrm{ctx}}^{(k)}
\in\mathbb{R}^{B\times N_k\times C_k}$.

\textbf{Semantic Feature Modulation.}
The text-conditioned context $C_{\mathrm{ctx}}^{(k)}$ obtained above
contains the textual information used to regulate the visual feature
at the $k$-th interaction stage. To inject this context into the
visual feature, it is first mapped to a pair of affine modulation
parameters:
\begin{equation}
[
\gamma^{(k)},
\beta^{(k)}
]
=
\operatorname{Split}_{2}
\left[
\operatorname{GELU}
\left(
W_{\gamma\beta}^{(k)}
C_{\mathrm{ctx}}^{(k)}
+
b_{\gamma\beta}^{(k)}
\right)
\right],
\label{eq:context_gamma_beta}
\end{equation}
where
$W_{\gamma\beta}^{(k)}:
\mathbb{R}^{C_k}\rightarrow\mathbb{R}^{2C_k}$.
The $2C_k$-dimensional output is evenly split into the scaling
parameter $\gamma^{(k)}$ and the bias parameter $\beta^{(k)}$.
The two parameters are jointly generated by the same linear
projection followed by GELU, while different hierarchical stages use
independent parameter sets.

After reshaping $\gamma^{(k)}$ and $\beta^{(k)}$ from token form to
spatial feature maps, they are applied to the input visual feature
$F_\mathrm{f}^{(k)}$ as
\begin{equation}
\begin{aligned}
\widehat{F}_\mathrm{f}^{(k)}
&=
\varphi^{(k)}
\left(
F_\mathrm{f}^{(k)},
F_{\mathrm{text}}
\right)
=
\left(
1+
\operatorname{reshape}
\left(
\gamma^{(k)}
\right)
\right)
\odot
F_\mathrm{f}^{(k)}
+
\operatorname{reshape}
\left(
\beta^{(k)}
\right),
\end{aligned}
\label{eq:text_condition_modulation}
\end{equation}
Here, $\gamma^{(k)}$ controls the feature scaling, while
$\beta^{(k)}$ provides an additive feature shift. 
$\widehat{F}_\mathrm{f}^{(k)}$ denotes the text-modulated output
feature, whereas $F_\mathrm{f}^{(k)}$ denotes the visual feature before text
conditioning. The function
$\varphi^{(k)}(\cdot,\cdot)$ represents the complete
text-conditioning operation at the $k$-th stage, and its parameters
are not shared across hierarchical stages.

\begin{figure}[t]
    \centering
\includegraphics[width=0.85\textwidth]{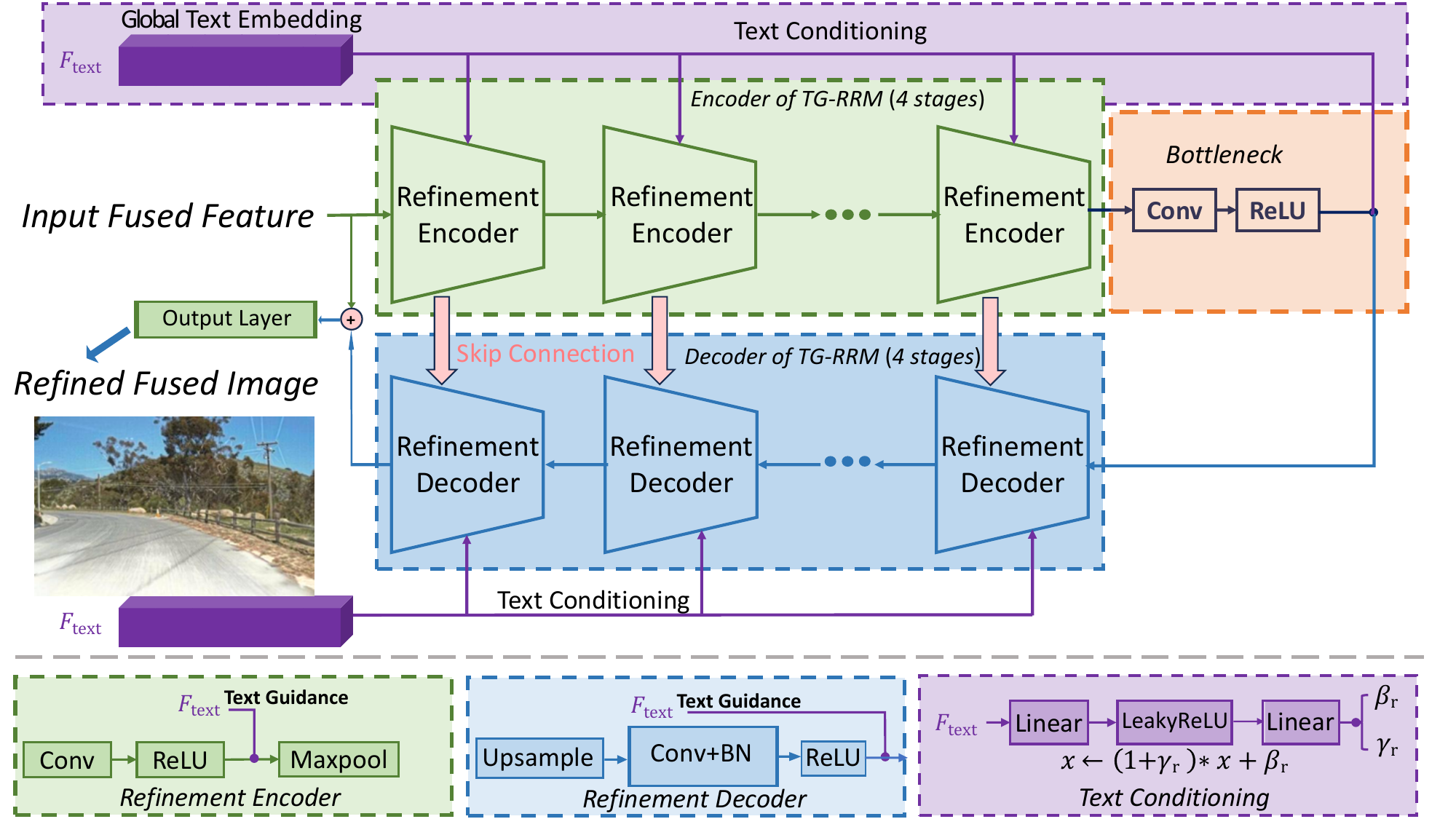}
    \caption{The U-Net structured Text-Guided Residual Refinement Module \textbf{(TG-RRM)}, which consists of four encoder-decoder stages with text-conditioned modulation.}
    \label{fig:TG_RRM}
\end{figure}

\subsection{Text-Guided Residual Refinement Module (TG-RRM)}

After four hierarchical text-image interactions, the network produces a high-resolution fused feature representation that already encodes rich cross-modal information under textual guidance. Although this intermediate feature captures the global fusion structure, it may still contain local ambiguities and degradation-related residual responses before final image reconstruction.

To alleviate these limitations, refinement stages have been widely adopted in
image restoration and fusion frameworks~\cite{zamir2022restormer,sun2024restoring}.
Such strategies enhance visual quality by correcting prediction errors and
improving detail fidelity, typically through convolutional or transformer-based
layers. Related refinement strategies have also been explored in event-based stereo depth estimation, where URNet~\cite{cheng2025urnet} combines local-global refinement with uncertainty modeling to improve disparity prediction. For infrared-visible image fusion, however, refinement under challenging degradation conditions remains difficult. Coarse outputs often contain structural ambiguities and regionally inconsistent responses, which restrict the ability of shallow or single-scale refinement modules to fully recover fine details and coherent structures.

Motivated by these observations, we introduce TG-RRM, designed to enhance fine-grained details and improve structural coherence.
As illustrated in Fig.~\ref{fig:TG_RRM}, TG-RRM adopts a
compact U-Net architecture that takes the high-resolution fused feature as input and predicts a residual feature correction. This correction is added to the input feature, after which the network's output layer reconstructs the three-channel fused image. The encoder-decoder structure enables progressive feature abstraction and reconstruction, while skip connections directly propagate low-level spatial information, facilitating the preservation of fine textures and sharp boundaries.
The refinement process is explicitly conditioned on the global CLIP
text embedding $F_{\text{text}}$, allowing semantic guidance to remain active during residual correction. Specifically, the textual information is injected into both encoder and
decoder stages through a lightweight feature-wise modulation mechanism. The text
embedding is mapped via a small MLP to generate affine parameters $(\gamma_{\mathrm{r}},\beta_{\mathrm{r}})$,
which modulate intermediate feature responses according to
$x \leftarrow (1+\gamma_{\mathrm{r}})\odot x + \beta_{\mathrm{r}}$. The text-conditioned affine transformations used at different encoder stages, the bottleneck, decoder stages, and the residual output stage are independently parameterized and do not share parameters across feature levels. At each modulation site, a separate two-layer MLP with dimensions $512\rightarrow1024\rightarrow2C$ and a LeakyReLU activation between the two linear layers jointly produces the $2C$-dimensional affine vector, which is subsequently split into the scaling term $\gamma_{\mathrm{r}}$ and bias term $\beta_{\mathrm{r}}$. Thus, the two affine parameters share the same MLP within one site, while different scales and refinement sites use independent MLP parameters.

This text-guided refinement strategy enables TG-RRM to adaptively emphasize
task-relevant structures while suppressing degradation-related artifacts. By
combining multi-scale U-Net refinement with semantic conditioning, TG-RRM
produces fusion results with improved visual consistency, sharper structural
details, and better alignment with the intended textual guidance.

\subsection{Loss Functions}
\label{subsec:loss_functions}

The loss function largely determines what source information is preserved
and how different modalities contribute to the final fused result.
Following prior text-guided fusion frameworks, the fusion-related losses
include intensity loss, structural similarity (SSIM) loss~\cite{zhao2016loss},
maximum gradient loss, and color consistency loss. Since no ground-truth fused
images are available, the fusion network is trained using a combination of
complementary loss terms that jointly encourage visual saliency, structural
preservation, and color consistency. The overall training objective is
formulated as a weighted sum of these losses.

\textbf{Intensity loss:}
The intensity loss preserves the source pixel selected by a luminance comparison while retaining the three-channel value of the selected modality. Let
\begin{equation}
\bar I_m=\frac{1}{3}\sum_{c=1}^{3}I_{m,c},\quad
M=\mathbf{1}\!\left[\bar I_{\mathrm{ir}}>\bar I_{\mathrm{vis}}\right],
\end{equation} where $m\in\{\mathrm{vis},\mathrm{ir}\}$ and $M$ is broadcast to the three image channels. The intensity target and loss are
\begin{align}
I_{\max} &= M\odot I_{\mathrm{ir}}+(1-M)\odot I_{\mathrm{vis}},\quad
\mathcal{L}_{\text{int}} = \left\lVert I_{\text{fused}}-I_{\max}\right\rVert_1,
\label{eq:intensity_loss}
\end{align}
where the $\ell_1$ loss is averaged over pixels and channels. For synthetically degraded samples in EMS, the clean source pair stored with the training sample is used to construct the fusion losses. For standard datasets without separate clean counterparts, the available infrared and visible source images are used directly.

\textbf{Structural similarity loss:}
To preserve structural information from both modalities, the implementation computes a visible-image SSIM term in RGB space and an infrared SSIM term after grayscale conversion. The structural similarity loss is defined as
\begin{equation}
\begin{aligned}
\mathcal{L}_{\text{SSIM}}={}&
w_{\text{ssim-vis}}\!\left[1-\mathrm{SSIM}(I_{\text{fused}},I_{\text{vis}})\right]+w_{\text{ssim-ir}}\!\left[1-\mathrm{SSIM}(G(I_{\text{fused}}),G(I_{\text{ir}}))\right],
\end{aligned}
\label{eq:ssim_loss}
\end{equation}
where $G(\cdot)$ denotes RGB-to-grayscale conversion. The SSIM index is computed in its standard form:
\begin{equation}
\mathrm{SSIM}(x, y) =
\frac{(2\mu_x \mu_y + C_1)(2\sigma_{xy} + C_2)}
{(\mu_x^2 + \mu_y^2 + C_1)(\sigma_x^2 + \sigma_y^2 + C_2)},
\label{eq:ssim_standard}
\end{equation}
where $\mu$, $\sigma$, and $\sigma_{xy}$ denote local means, standard deviations, and cross-covariance computed within a sliding window. $C_1$ and $C_2$ are small constants that prevent division by zero.

\textbf{Maximum gradient loss:}
The maximum gradient loss is introduced to preserve sharp edges from both modalities. In the implementation, the visible, infrared, and fused images are first converted to grayscale. Horizontal and vertical absolute Sobel responses are then computed separately, and the fused response is matched to the element-wise maximum source response in each direction:
\begin{equation}
\begin{aligned}
\mathcal{L}_{\mathrm{grad}}
={}&
\left\|
D_x\!\left(G(I_{\mathrm{fused}})\right)
-
\max
\left(
D_x\!\left(G(I_{\mathrm{vis}})\right),
D_x\!\left(G(I_{\mathrm{ir}})\right)
\right)
\right\|_1\\
&+
\left\|
D_y\!\left(G(I_{\mathrm{fused}})\right)
-
\max
\left(
D_y\!\left(G(I_{\mathrm{vis}})\right),
D_y\!\left(G(I_{\mathrm{ir}})\right)
\right)
\right\|_1,
\end{aligned}
\label{eq:gradient_loss}
\end{equation}
where $D_x(\cdot)$ and $D_y(\cdot)$ denote the absolute horizontal and vertical Sobel responses after grayscale conversion. The Sobel filters are applied after one-pixel replicate padding, and
the absolute horizontal and vertical responses are matched separately.

\textbf{Color consistency loss:}
The color consistency loss encourages the fused image to preserve the chrominance information of the visible image. Both images are transformed to YCbCr space, and the Cb and Cr channels are constrained with an $\ell_1$ distance:
\begin{equation}
\mathcal{L}_{\text{color}} =
\left\lVert C_\text{b}(I_{\text{fused}})-C_\text{b}(I_{\text{vis}})\right\rVert_1
+
\left\lVert C_\text{r}(I_{\text{fused}})-C_\text{r}(I_{\text{vis}})\right\rVert_1.
\label{eq:color_loss}
\end{equation}

\textbf{Overall training objective:}
The overall training objective is composed of the intensity, structural
similarity, color consistency, and maximum gradient losses:
\begin{equation}
\begin{aligned}
\mathcal{L}_{\text{total}} ={}&
w_{\text{int}}\,\mathcal{L}_{\text{int}}
+w_{\text{ssim}}\,\mathcal{L}_{\text{SSIM}}
+w_{\text{color}}\,\mathcal{L}_{\text{color}}
+w_{\text{grad}}\,\mathcal{L}_{\text{grad}},
\end{aligned}
\label{eq:total_loss}
\end{equation}
where $w_{\text{int}}$, $w_{\text{ssim}}$, $w_{\text{color}}$, and
$w_{\text{grad}}$ denote the coefficients of the corresponding loss terms.
The coefficients $w_{\text{ssim-vis}}$ and $w_{\text{ssim-ir}}$ are applied
inside $\mathcal{L}_{\text{SSIM}}$, as defined in
Eq.~\eqref{eq:ssim_loss}.

%% file: sections/4_Experiments.tex
\section{Results}

In this section, we first describe the implementation details and experimental configurations.
We then evaluate the effectiveness of the proposed method through both qualitative and quantitative comparisons, with a particular focus on text-guided image fusion.
Finally, ablation experiments are conducted to analyze the contributions of different components in the proposed framework.

\subsection{Implementation Details}
\label{subsec:implementation_dataset}
All models are implemented in PyTorch and trained on NVIDIA RTX A5000 GPU.
We adopt the AdamW optimizer with a learning rate of $1 \times 10^{-4}$, a weight decay of $5 \times 10^{-2}$, and a batch size of 8.
Training is conducted for 120 epochs, and the checkpoint achieving the lowest validation loss is selected for final evaluation.
During training, input images are randomly cropped to a spatial resolution of $96 \times 96$.
Horizontal and vertical flips are applied with a probability of 0.5 to improve generalization.

\subsection{Datasets}
\label{standard datasets}
\textbf{Standard Datasets.} We evaluate the model on several standard infrared-visible fusion datasets, including MSRS~\cite{tang2022piafusion}, LLVIP~\cite{jia2021llvip}, MFNet~\cite{ha2017mfnet}, and RoadScene~\cite{xu2020u2fusion}.
These datasets primarily represent single-degradation conditions.
MSRS and LLVIP mainly contain low-light visible images, whereas MFNet focuses on infrared images with low contrast.
RoadScene contains outdoor traffic scenes that exhibit illumination imbalance and exposure variations, providing an additional real-world evaluation scenario.
Due to the large size of LLVIP, we follow common practice and randomly select 40\% of the dataset, training on this subset for 50 epochs. Since MSRS and LLVIP do not provide ground-truth fused images, the original infrared and visible image pairs are directly used as supervision targets during training and loss computation.

\label{EMS dataset}
\textbf{EMS Degradation Dataset.} In addition to the standard fusion datasets, to evaluate robustness under diverse synthetic degradation conditions that approximate common imaging artifacts, we construct an extended multi-degradation dataset based on three publicly available infrared-visible datasets: MSRS~\cite{tang2022piafusion}, LLVIP~\cite{jia2021llvip}, and MFNet~\cite{ha2017mfnet}.
The extended dataset includes nine common degradation types: \emph{Low Light}, \emph{Rain}, \emph{Visible Blur}, \emph{Overexposure}, \emph{Visible Haze}, \emph{Visible Random Noise}, \emph{Infrared Low Contrast}, \emph{Infrared Random Noise}, and \emph{Infrared Stripe Noise}.
Each degraded sample consists of an RGB–IR image pair and is accompanied by a corresponding textual description, enabling text-conditioned fusion training.  For degradation types without predefined descriptions,
additional prompts are created following the structure and
semantics of existing templates to maintain consistency across
all degradation scenarios.

\begin{table}[t]
\centering
\caption{
Quantitative comparison of IT-TextFusion with existing image fusion methods and the baseline Text-IF~\cite{yi2024text} on the MSRS, LLVIP, and RoadScene datasets using a fixed generic prompt without degradation-aware or scene-specific semantic information (\textbf{Bold}: best performance).
}
\label{tab:quant_msrs_llvip_roadscene}

\resizebox{\textwidth}{!}{
\setlength{\tabcolsep}{4pt}
\renewcommand{\arraystretch}{1.25}

\begin{tabular}{c|ccccc|ccccc|ccccc}
\hline
\multirow{2}{*}{\textbf{Methods}}
& \multicolumn{5}{c|}{\textbf{MSRS Dataset}}
& \multicolumn{5}{c|}{\textbf{LLVIP Dataset}}
& \multicolumn{5}{c}{\textbf{RoadScene Dataset}} \\
\cline{2-16}
& SCD$\uparrow$ & SD$\uparrow$ & EN$\uparrow$ & VIFF$\uparrow$ & $Q^{AB/F}\uparrow$
& SCD$\uparrow$ & SD$\uparrow$ & EN$\uparrow$ & VIFF$\uparrow$ & $Q^{AB/F}\uparrow$
& SCD$\uparrow$ & SD$\uparrow$ & EN$\uparrow$ & VIFF$\uparrow$ & $Q^{AB/F}\uparrow$ \\
\hline
UMF-CMGR ~\cite{wang2022unsupervised}
& 0.981 & 20.819 & 5.600 & 0.430 & 0.266
& 1.029 & 31.501 & 6.569 & 0.509 & 0.352
& 1.613 & 36.251 & 6.973 & 0.554 & 0.429 \\
TarDAL ~\cite{liu2022target}
& 1.484 & 35.460 & 6.347 & 0.673 & 0.426
& 0.817 & 39.070 & 5.349 & 0.330 & 0.252
& 1.415 & 42.609 & 7.054 & 0.525 & 0.391 \\
ReCoNet ~\cite{huang2022reconet}
& 1.191 & 44.374 & 3.895 & 0.438 & 0.367
& 1.345 & 41.234 & 5.514 & 0.513 & 0.364
& 1.589 & 37.580 & 6.822 & 0.504 & 0.354 \\
MURF ~\cite{xu2023murf}
& 0.868 & 16.431 & 5.047 & 0.413 & 0.327
& 0.514 & 21.834 & 6.051 & 0.386 & 0.206
& 1.576 & 36.788 & 6.992 & 0.484 & 0.432 \\
U2Fusion ~\cite{xu2020u2fusion}
& 1.182 & 23.541 & 5.246 & 0.506 & 0.372
& 0.757 & 23.614 & 5.972 & 0.552 & 0.341
& 1.498 & 30.969 & 6.739 & 0.513 & 0.467 \\
MetaFusion ~\cite{zhao2023metafusion}
& 1.486 & 39.432 & 6.368 & 0.726 & 0.478
& 1.317 & 42.446 & 6.823 & 0.833 & 0.493
& 1.581 & \textbf{50.613} & 7.223 & 0.512 & 0.338 \\
DDFM ~\cite{zhao2023ddfm}
& 1.550 & 32.749 & 5.693 & 0.622 & 0.431
& 1.414 & 38.346 & 6.979 & 0.549 & 0.220
& \textbf{1.864} & 44.925 & 7.226 & 0.544 & 0.413 \\
Text-IF ~\cite{yi2024text}
& 1.681 & \textbf{44.564} & 6.789 & 1.046 & 0.676
& \textbf{1.591} & 48.834 & 7.325 & 1.011 & 0.616
& 1.572 & 48.962 & 7.332 & \textbf{0.739} & 0.578 \\
\textbf{IT-TextFusion (Ours)}
&\textbf{1.721} & 42.889 & \textbf{6.882} & \textbf{1.064} & \textbf{0.725} & 1.520 & \textbf{50.451} & \textbf{7.443} & \textbf{1.042} & \textbf{0.760} & 1.102 & 49.102 & \textbf{7.332} & 0.733 & \textbf{0.636} \\
\hline
\end{tabular}
}
\end{table}
\subsection{Evaluation Metrics and Comparison Methods}
\label{subsec:evaluation_metrics}
\textbf{Evaluation Metrics.} Following the evaluation protocol of Text-IF~\cite{yi2024text}, we employ a comprehensive set of classical fusion metrics, including information entropy (EN)~\cite{ma2019infrared}, standard deviation (SD), spatial frequency (SF)~\cite{eskicioglu2002image}, mutual information (MI), sum of correlations of differences (SCD)~\cite{aslantas2015new}, visual information fidelity (VIFF)~\cite{han2013new}, gradient-based fusion quality $Q^{AB/F}$~\cite{ma2019infrared}, and a fusion-specific summed metric constructed based on the Structural Similarity Index Measure (SSIM)~\cite{wang2004image}, denoted as $\mathrm{SSIM}_{\mathrm{sum}}$.

Since no ground-truth fused image is available, the structural
similarity evaluation is computed between the fused image and each
of the two source images and then summed:
\begin{equation}
\mathrm{SSIM}_{\mathrm{sum}}
=
\mathrm{SSIM}(I_{\mathrm{f}},I_{\mathrm{ir}})
+
\mathrm{SSIM}(I_{\mathrm{f}},I_{\mathrm{vis}}),
\label{eq:ssim_sum_metric}
\end{equation}
where $I_{\mathrm{f}}$, $I_{\mathrm{ir}}$, and
$I_{\mathrm{vis}}$ denote the grayscale fused, infrared, and visible
images, respectively. Each pairwise SSIM term is computed using the
standard definition in Eq.~\eqref{eq:ssim_standard}, with the image intensity range set
to 255. Therefore, $\mathrm{SSIM}_{\mathrm{sum}}$ is not bounded by
1 and may take values greater than 1. A higher value indicates that
the fused image jointly preserves more structural information from
the two source modalities.

For datasets without reference images, we additionally report CLIP-IQA~\cite{wang2023exploring}, NIQE~\cite{mittal2012making}, MUSIQ~\cite{ke2021musiq}, and BRISQUE~\cite{mittal2012no}, which assess no-reference perceptual quality and image naturalness
using different statistical or learned representations.
Although CLIP-IQA uses CLIP-derived representations, the adopted
implementation receives only the fused image and does not use the
input degradation-aware prompt. Therefore, it is treated as a
no-reference perceptual-quality metric rather than a direct measure of
image-prompt semantic alignment.
Lower NIQE and BRISQUE values indicate higher perceived naturalness,
whereas higher EN, $Q^{\mathrm{AB/F}}$, VIFF, and
$\mathrm{SSIM}_{\mathrm{sum}}$ values indicate better information,
gradient, visual-fidelity, and joint structural preservation,
respectively. To ensure consistency with prior literature, all metrics are computed using publicly released implementations commonly adopted in recent fusion works.

\textbf{Comparison Methods.} We compare the proposed method with several state-of-the-art (SOTA) methods on multiple datasets, which are described in Section~\ref{standard datasets} In particular, the proposed method is mainly compared with Text-IF~\cite{yi2024text}, a recent representative framework for text-guided image fusion. Other methods for comparison include U2Fusion ~\cite{xu2020u2fusion}, MetaFusion ~\cite{zhao2023metafusion}, DDFM ~\cite{zhao2023ddfm}, MURF ~\cite{xu2023murf}, ReCoNet ~\cite{huang2022reconet}, TarDAL ~\cite{liu2022target}, and UMF-CMGR ~\cite{wang2022unsupervised}, which are conventional image fusion approaches relying purely on visual information without involving any text or language-based guidance. For the synthesized EMS dataset described in Section 4.2, we compare only with Text-IF.


\newlength{\qualimgH}
\setlength{\qualimgH}{1.8cm}   

\newlength{\promptH}
\setlength{\promptH}{0.25cm}    

\newcommand{\promptbox}[1]{%
  \parbox[t]{\textwidth}{%
    \centering
    {\bfseries\small Text:} {\small \emph{#1}}%
  }
  \\[4pt]
}
\newlength{\promptHEMS}
\setlength{\promptHEMS}{0.85cm} 

\newcommand{\promptboxEMS}[1]{%
  \parbox[t][\promptHEMS][t]{\linewidth}{%
    \centering
    {\bfseries\small Text:} {\small\emph{#1}}%
  }\\[2pt]
}

\newcommand{\onegroupEMS}[5]{%
  \promptboxEMS{#1}
  \onerowEMS{#2}{#3}{#4}{#5}
}

\newcommand{\imgcellStd}[2]{%
  \begin{minipage}[t]{0.245\linewidth}
    \centering
    \includegraphics[width=\linewidth,height=\qualimgH,clip,trim=0 0 0 0]{#1}\\[-0.2pt]
    {\footnotesize #2}
  \end{minipage}%
}
\newcommand{\onerowStd}[4]{%
  \imgcellStd{#1}{Infrared}\hspace{0.006\linewidth}%
  \imgcellStd{#2}{Visible}\hspace{0.006\linewidth}%
  \imgcellStd{#3}{Text-IF}\hspace{0.006\linewidth}%
  \imgcellStd{#4}{Ours}%
}

\newcommand{\imgcellEMS}[2]{%
  \begin{minipage}[t]{0.235\linewidth} 
    \centering
    \includegraphics[width=\linewidth,height=\qualimgH,clip,trim=0 0 0 0]{#1}\\[-0.2pt]
    {\footnotesize #2}
  \end{minipage}%
}
\newcommand{\onerowEMS}[4]{%
  \imgcellEMS{#1}{Infrared}\hspace{0.004\linewidth}%
  \imgcellEMS{#2}{Visible}\hspace{0.004\linewidth}%
  \imgcellEMS{#3}{Text-IF}\hspace{0.004\linewidth}%
  \imgcellEMS{#4}{Ours}%
}

\newcommand{\twocolrow}[9]{%
  \promptbox{#1}
  \begin{minipage}[t]{0.495\textwidth}
    \onerowStd{#2}{#3}{#4}{#5}
  \end{minipage}\hspace{0.01\textwidth}%
  \begin{minipage}[t]{0.495\textwidth}
    \onerowStd{#6}{#7}{#8}{#9}
  \end{minipage}
}

\begin{figure}[!t]
  \centering

  \twocolrow
    {In the context of infrared-visible light fusion, visible images may suffer from reduced quality in low-light scenarios.}
    {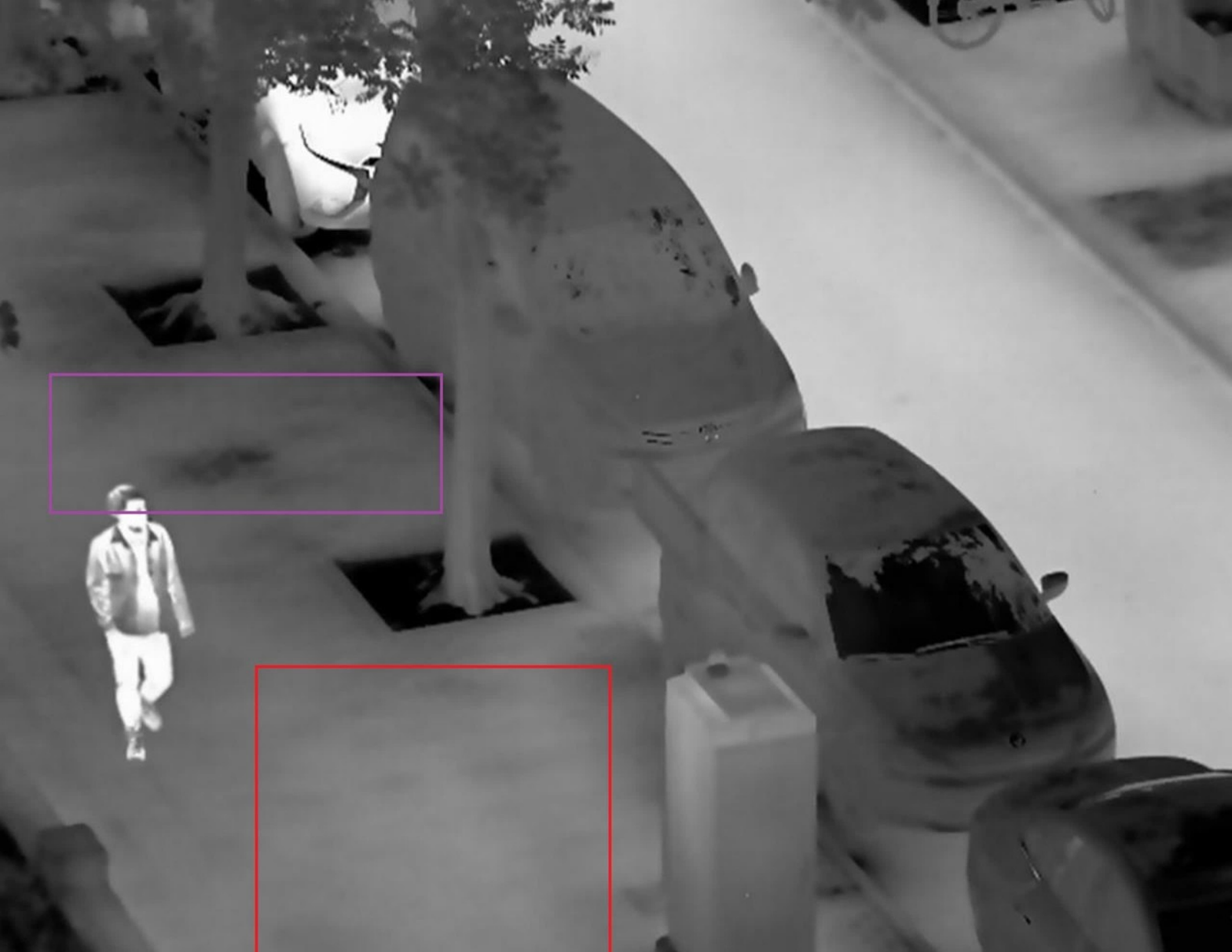}
    {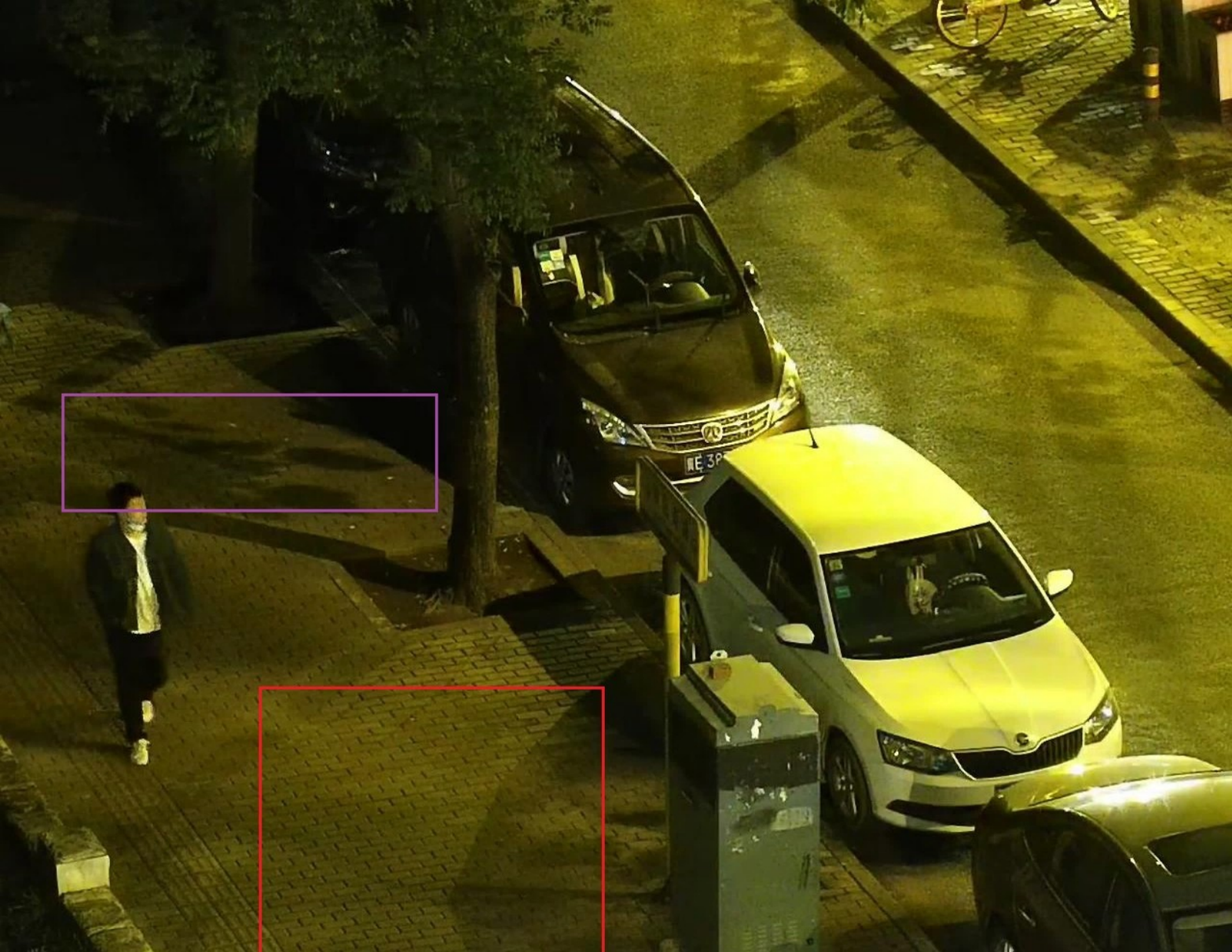}
    {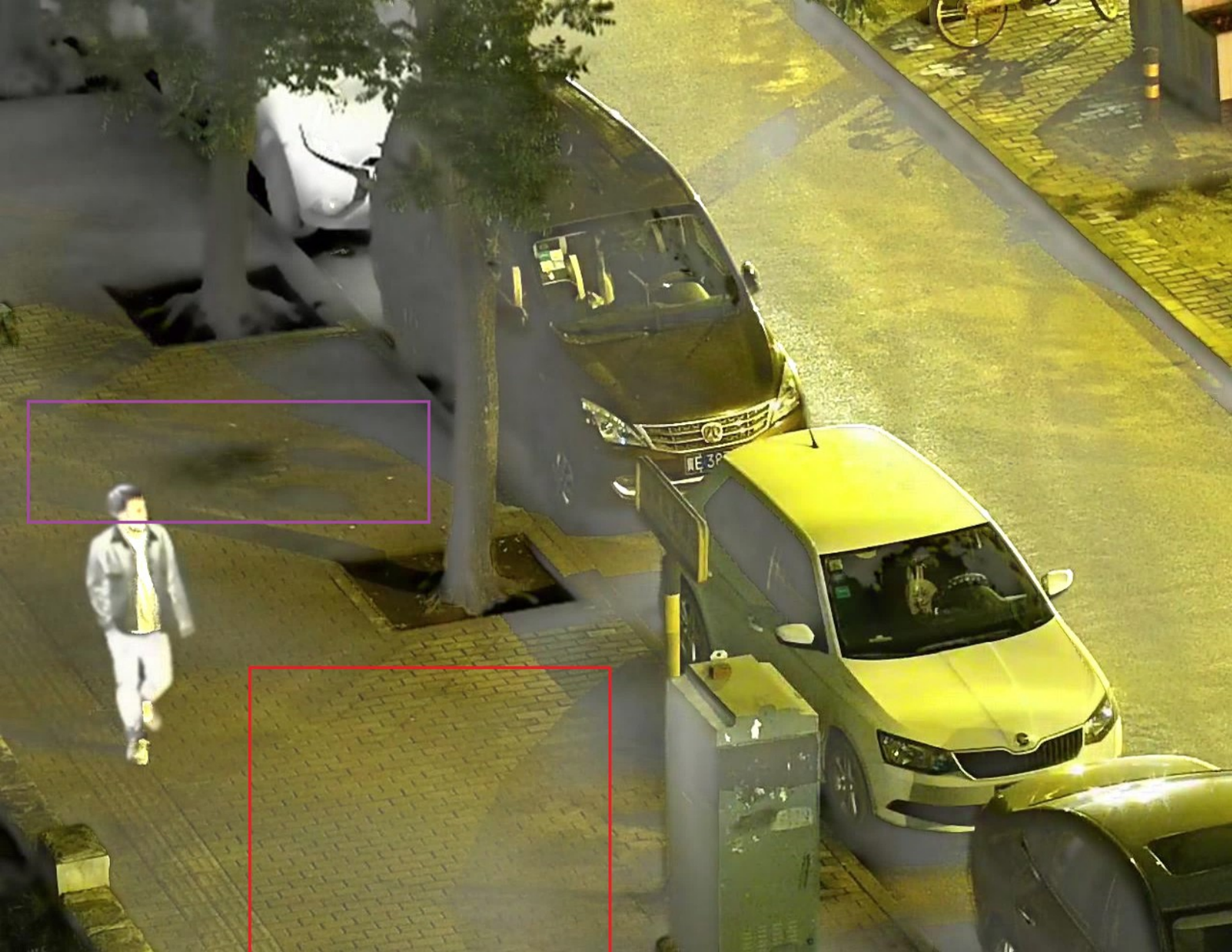}
    {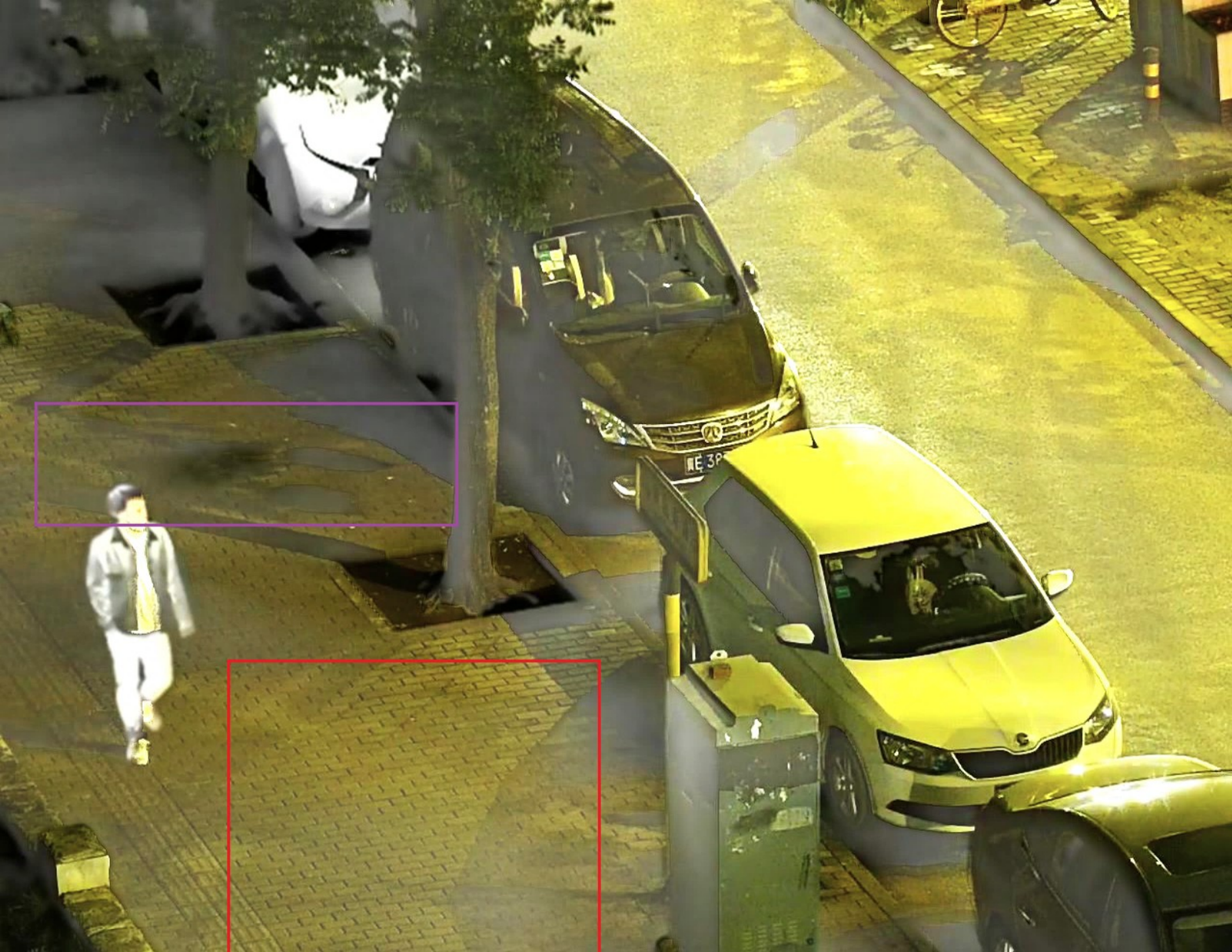}
    {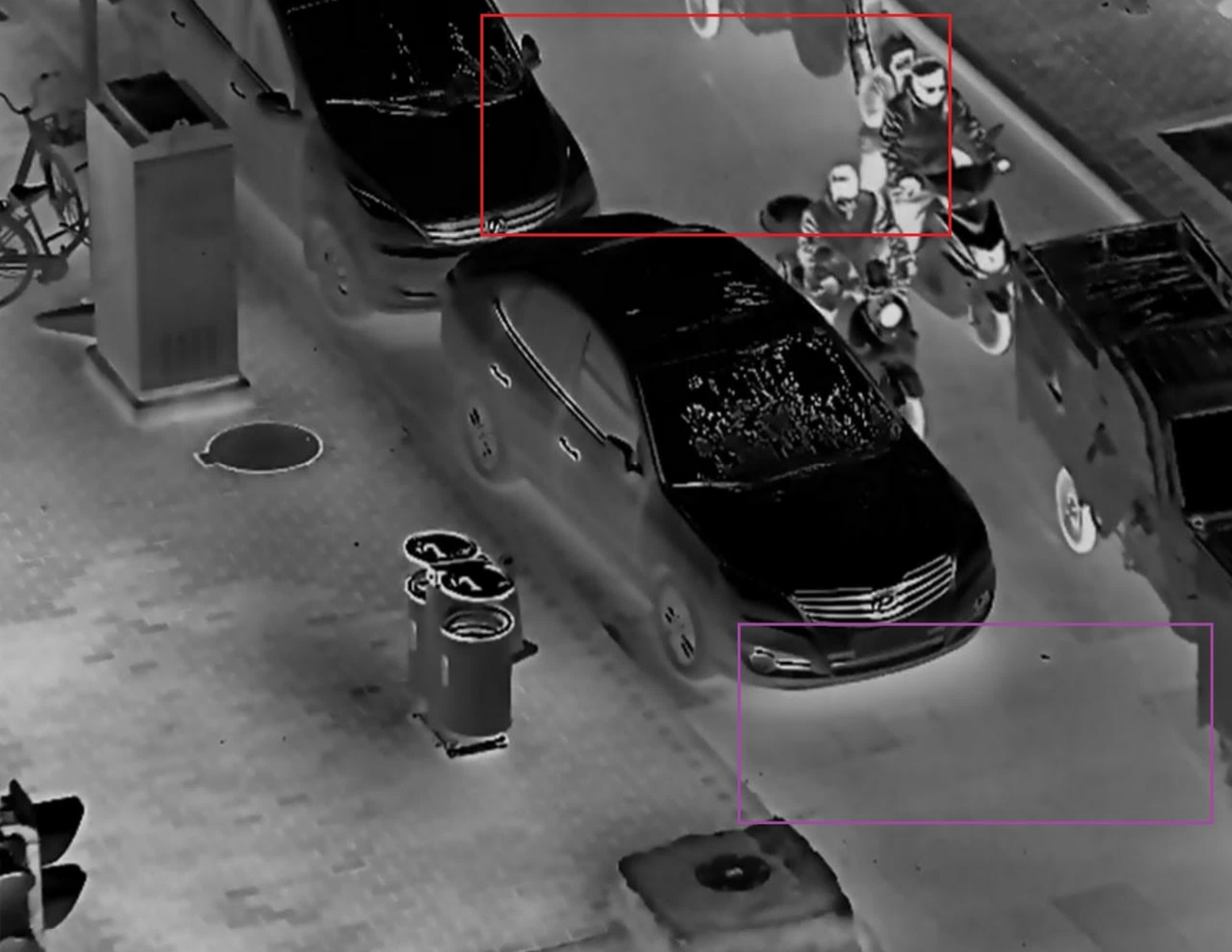}
    {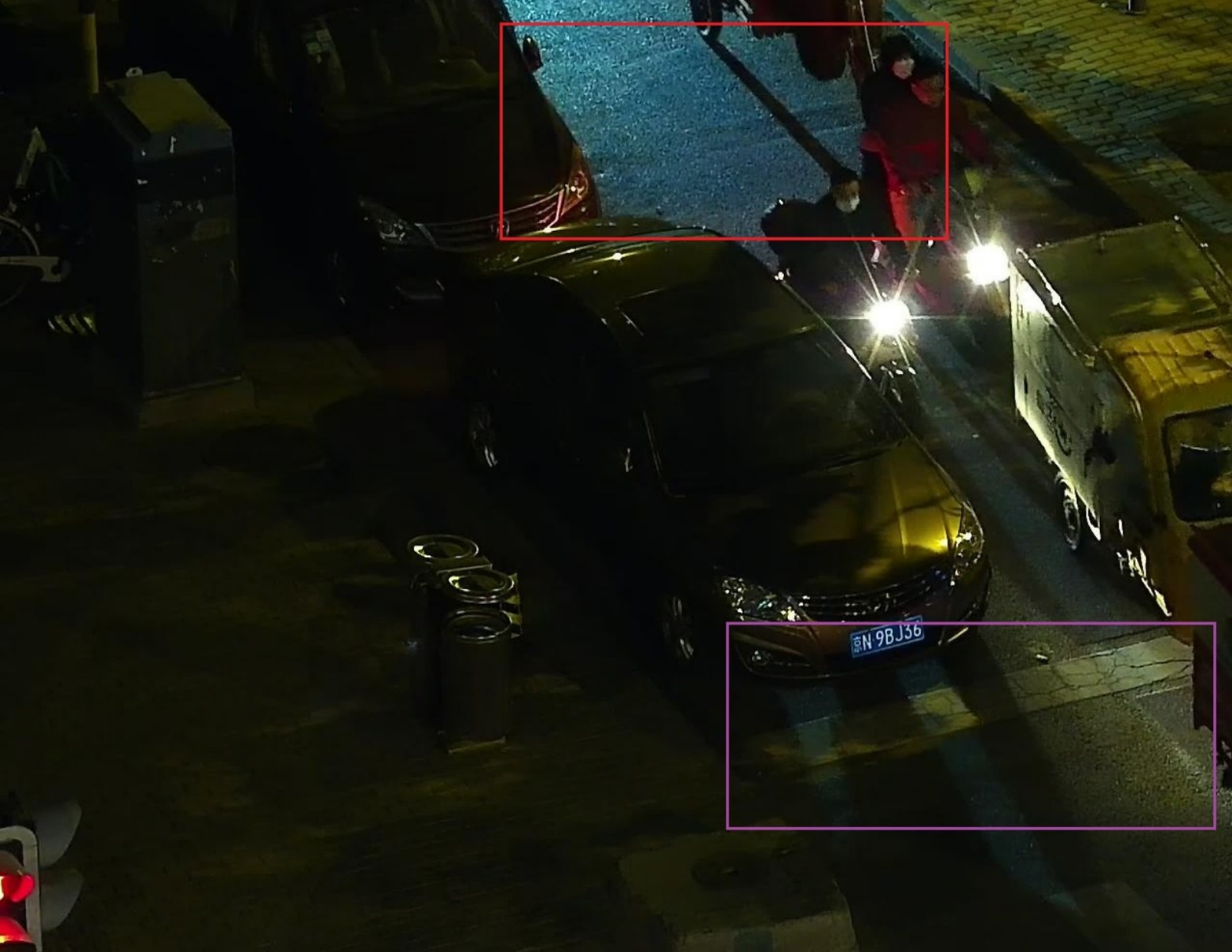}
    {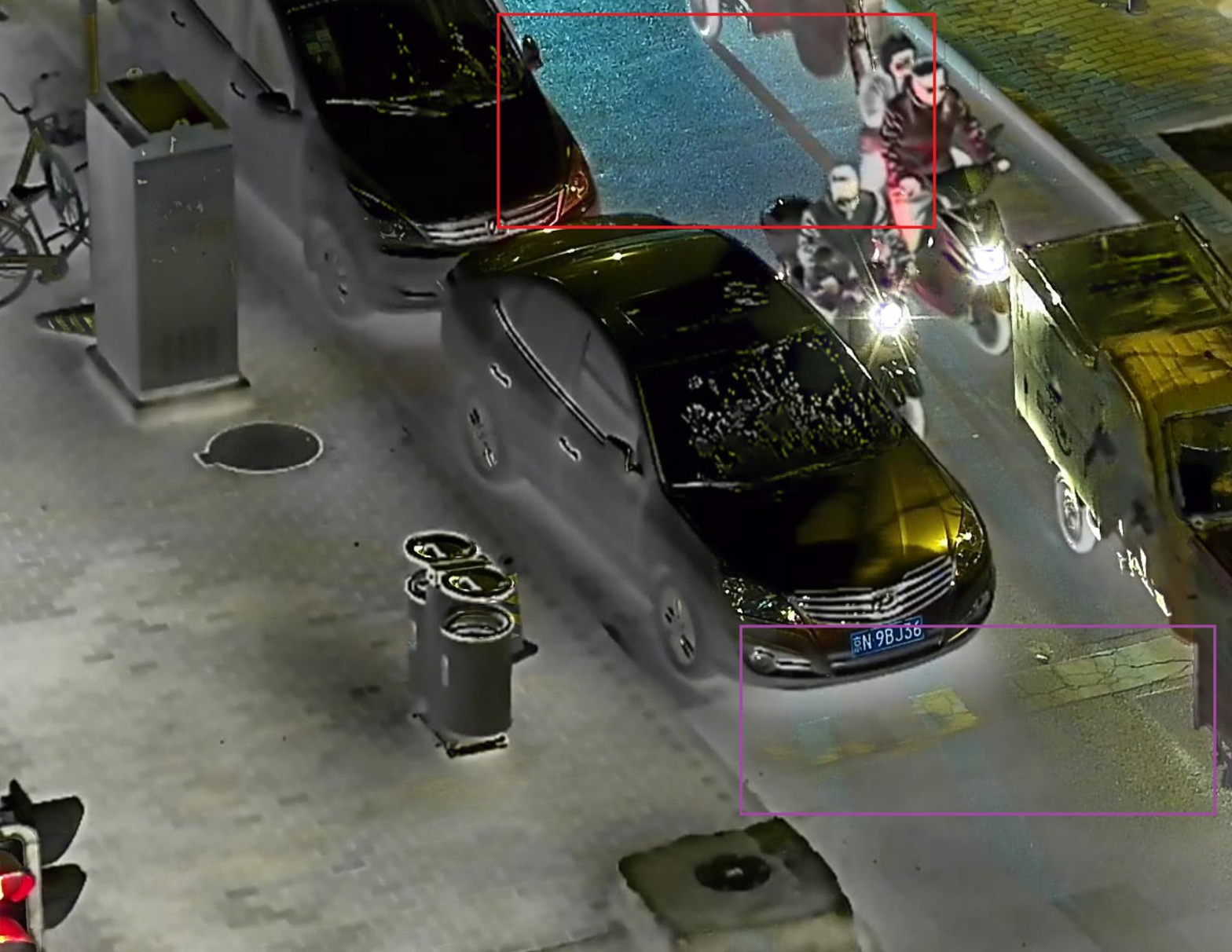}
    {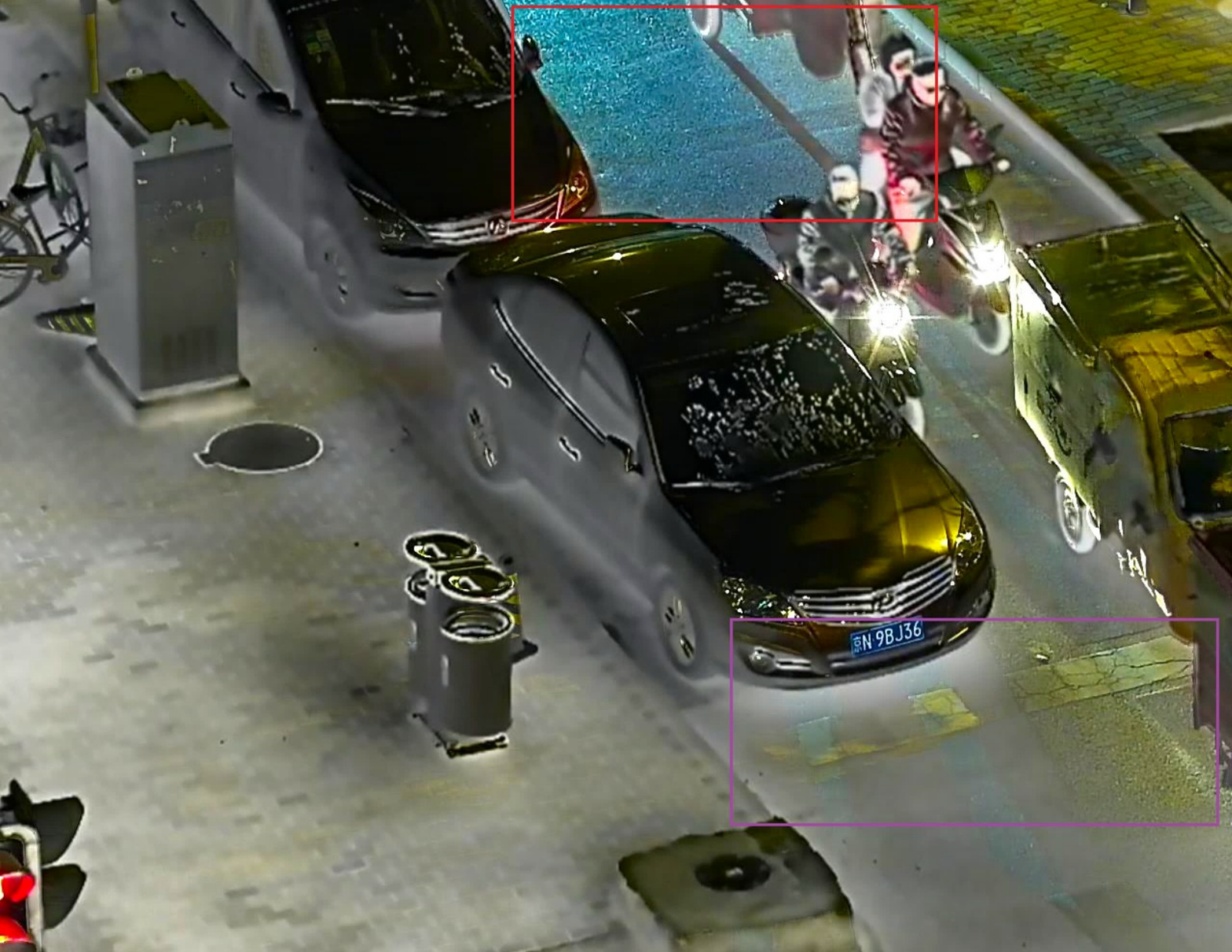}

  \vspace{5pt}

  \twocolrow
    {In this challenge, we focus on the low contrast degradation in the infrared images.}
    {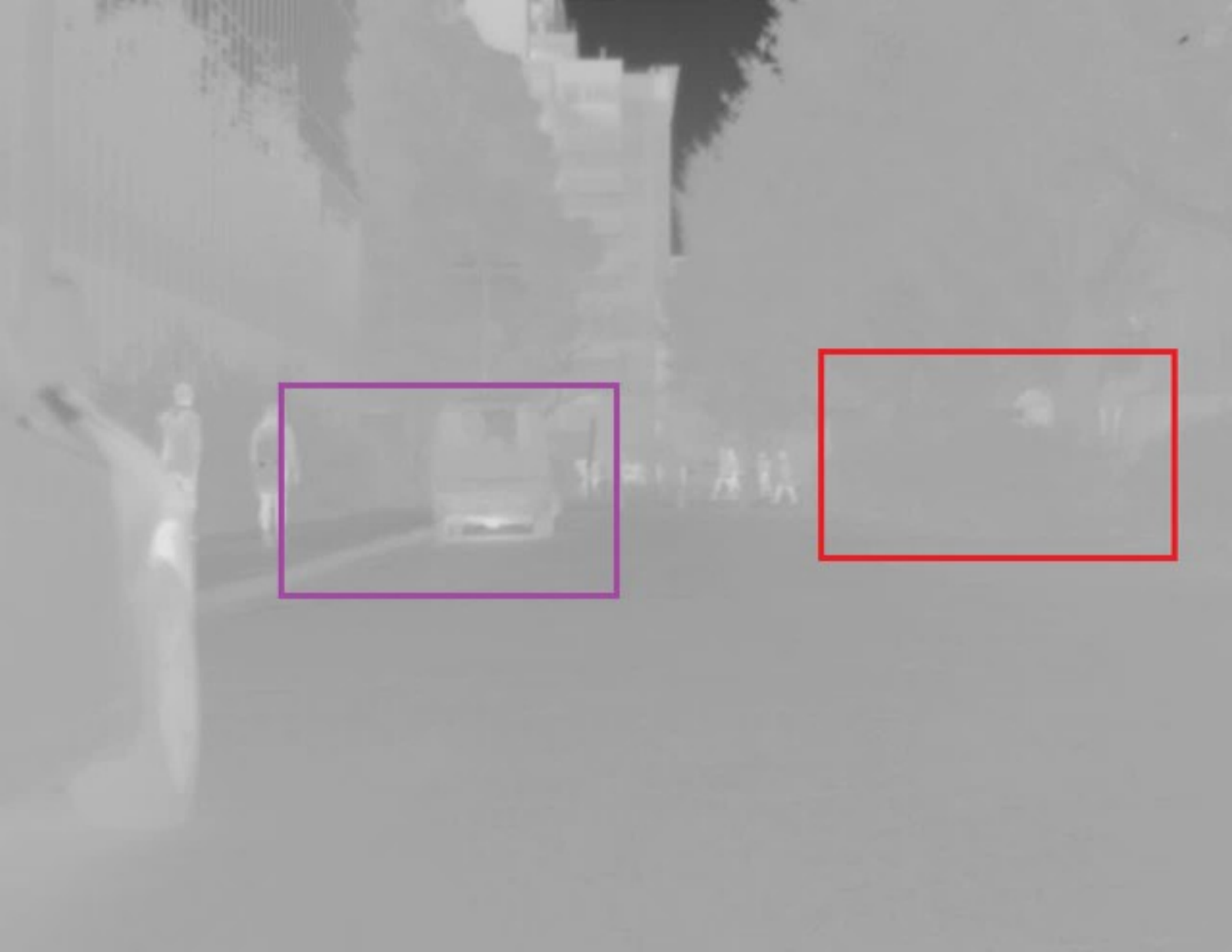}
    {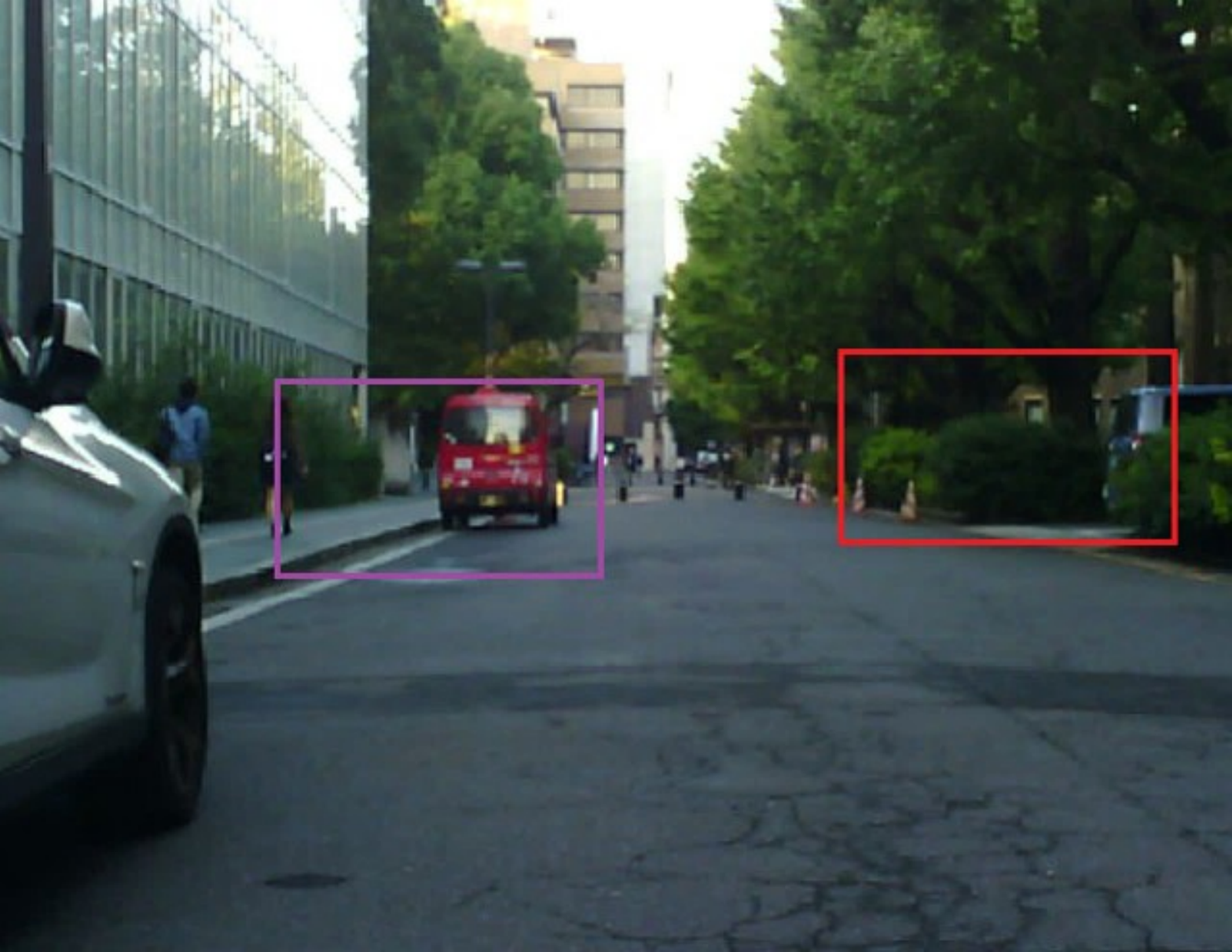}
    {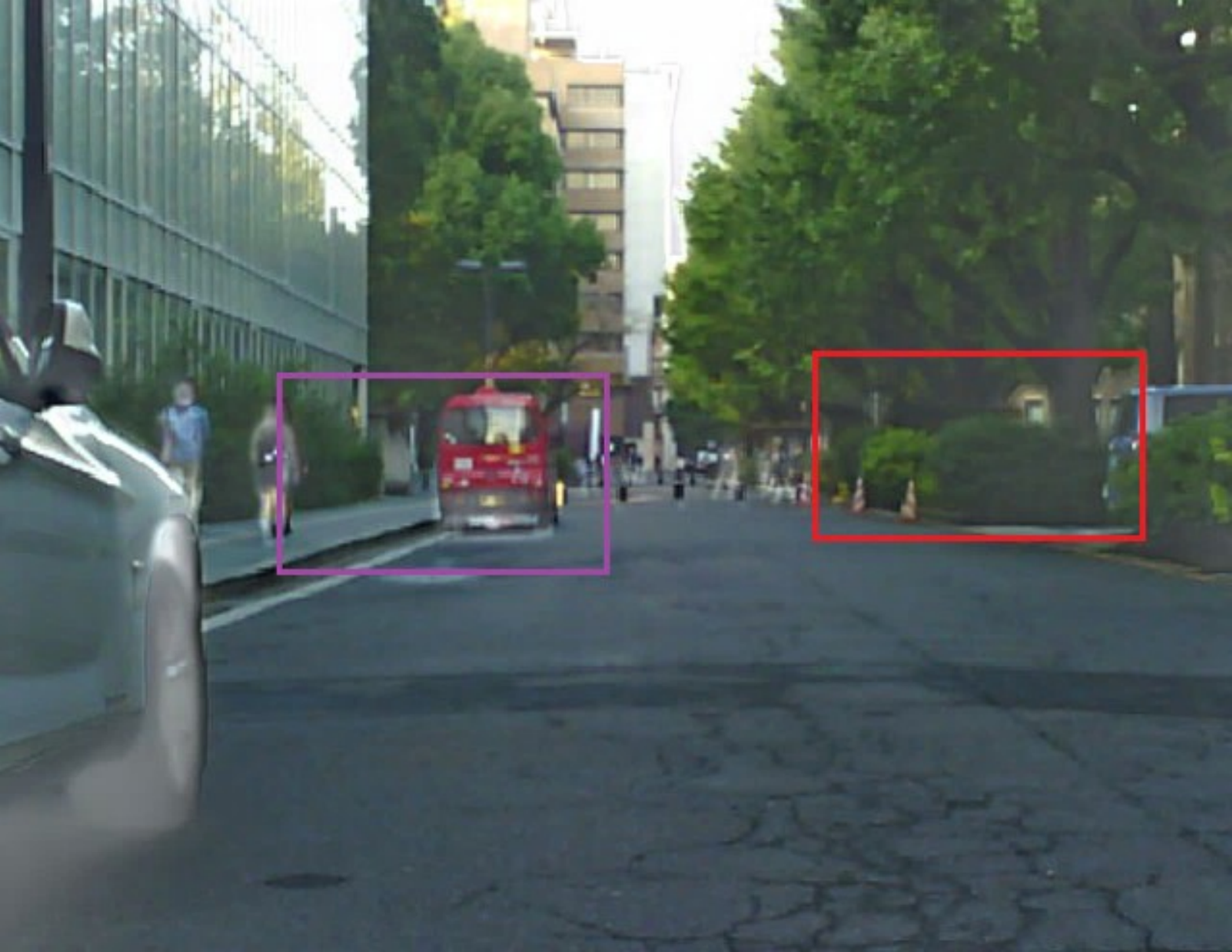}
    {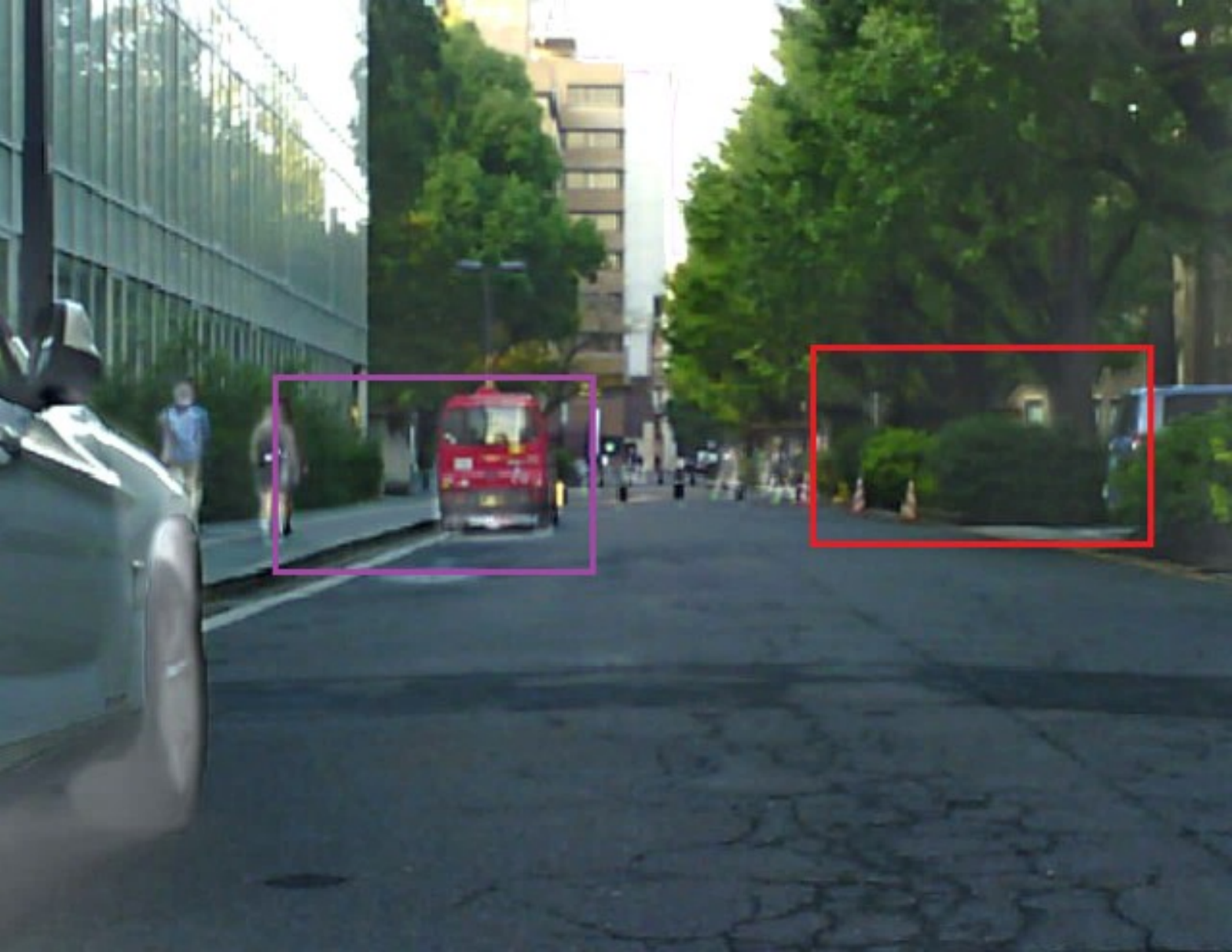}
    {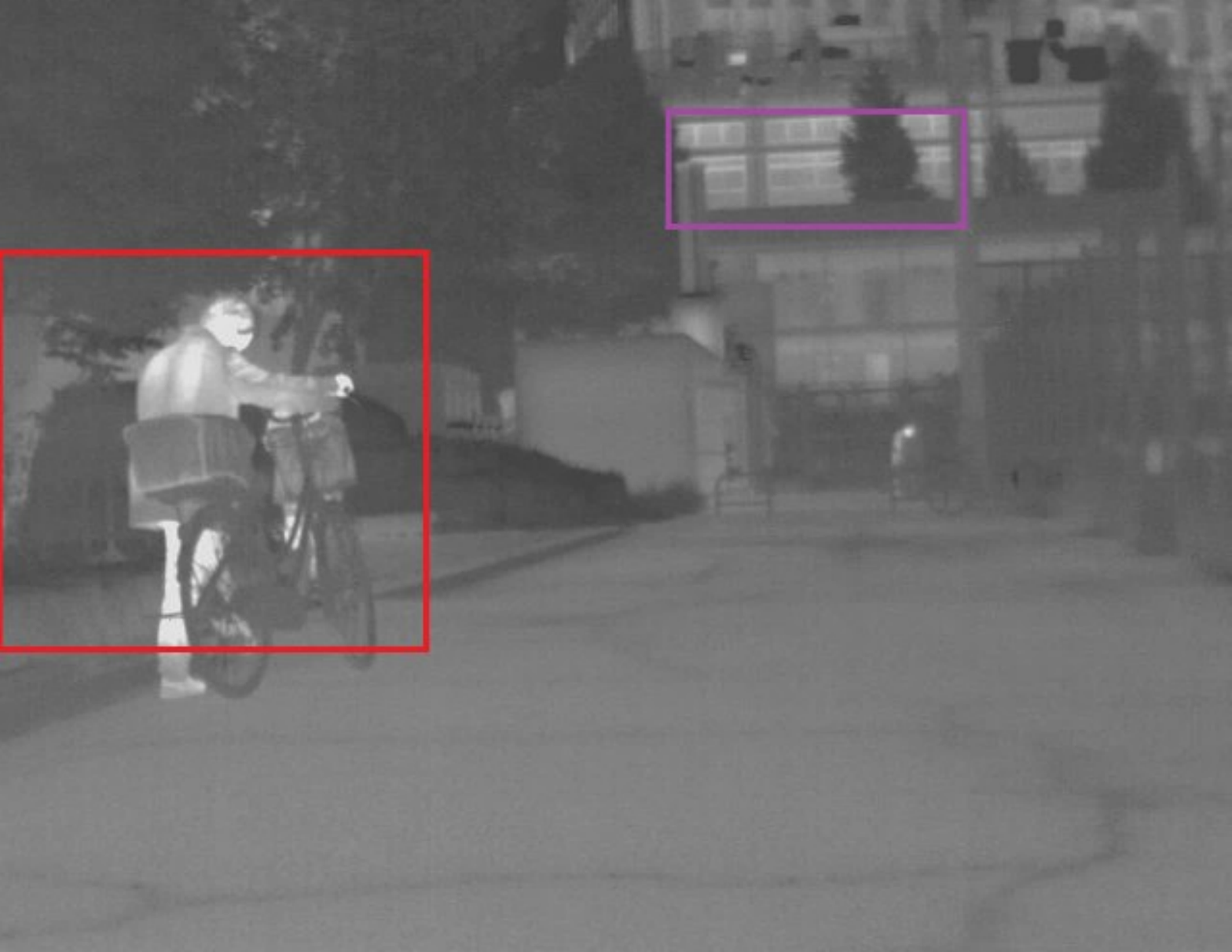}
    {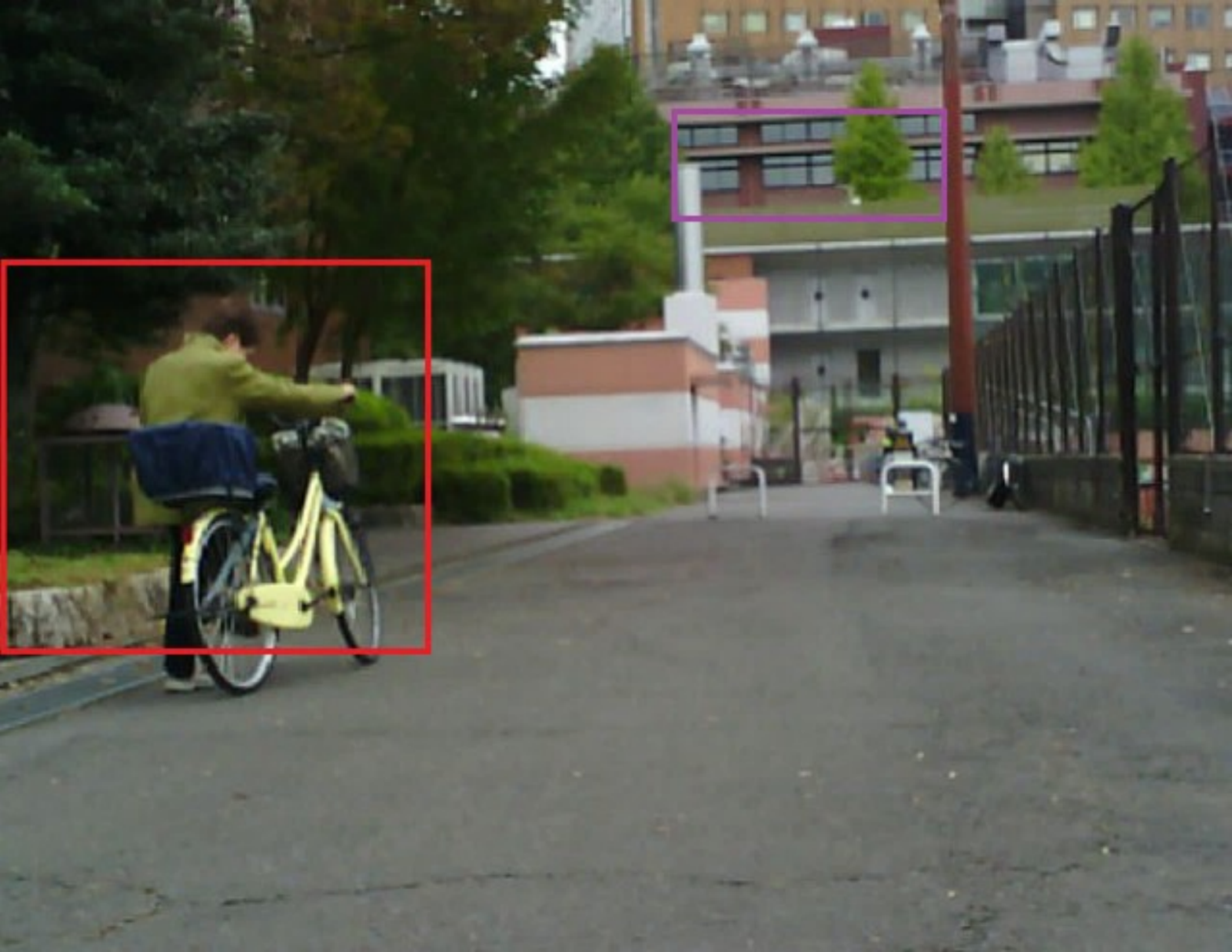}
    {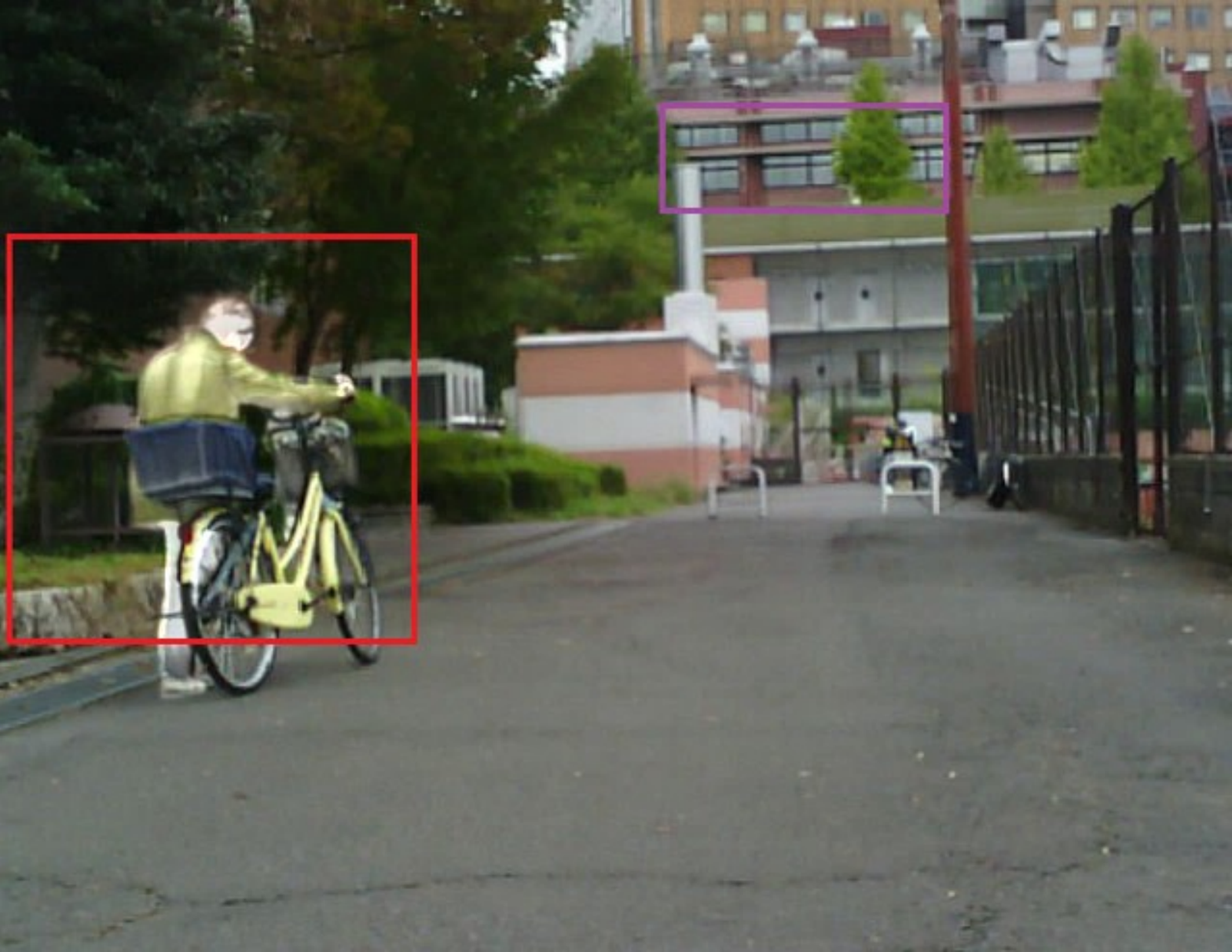}
    {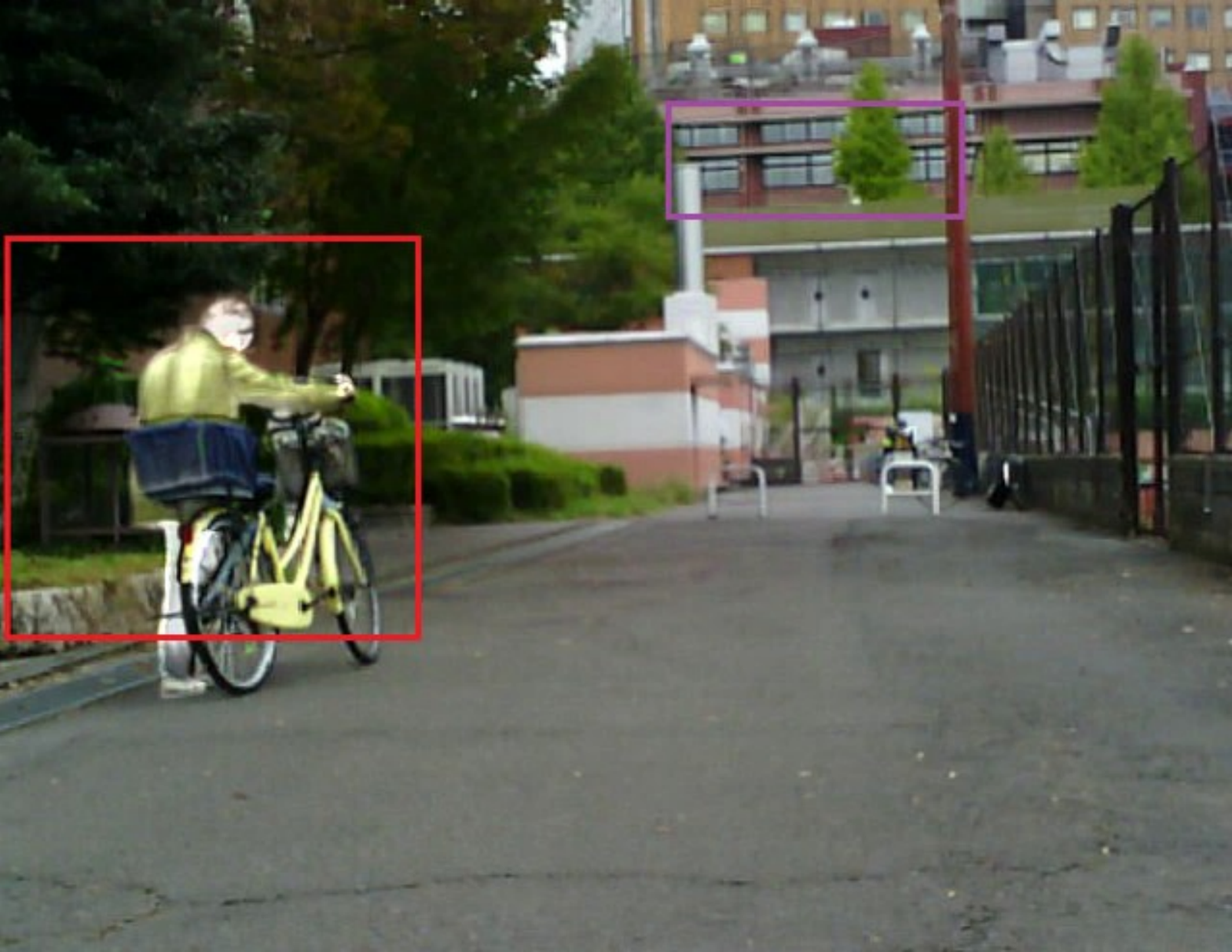}

  \vspace{5pt}

  \twocolrow
    {This pertains to the fusion of infrared and visible light images, with a low light degradation in the visible images.}
    {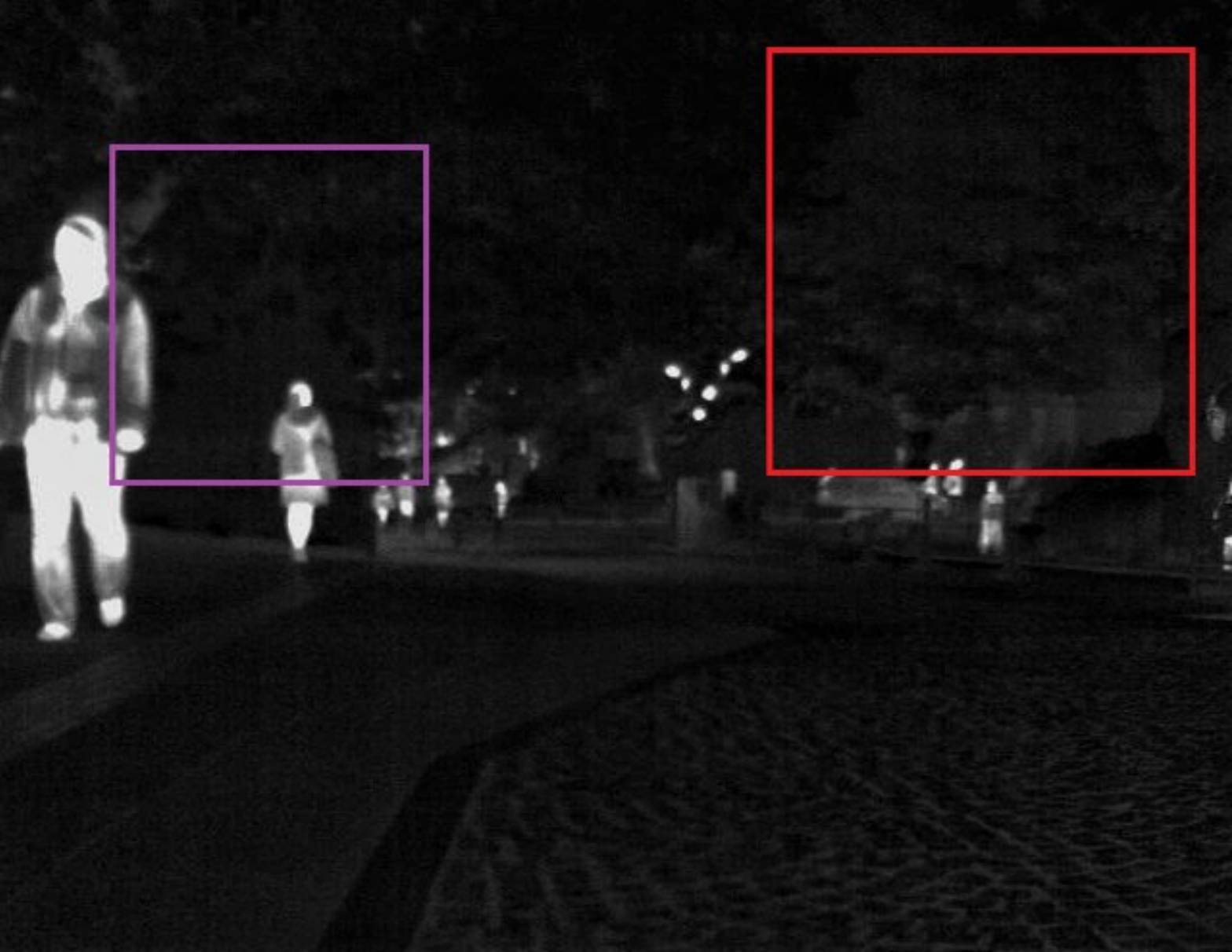}
    {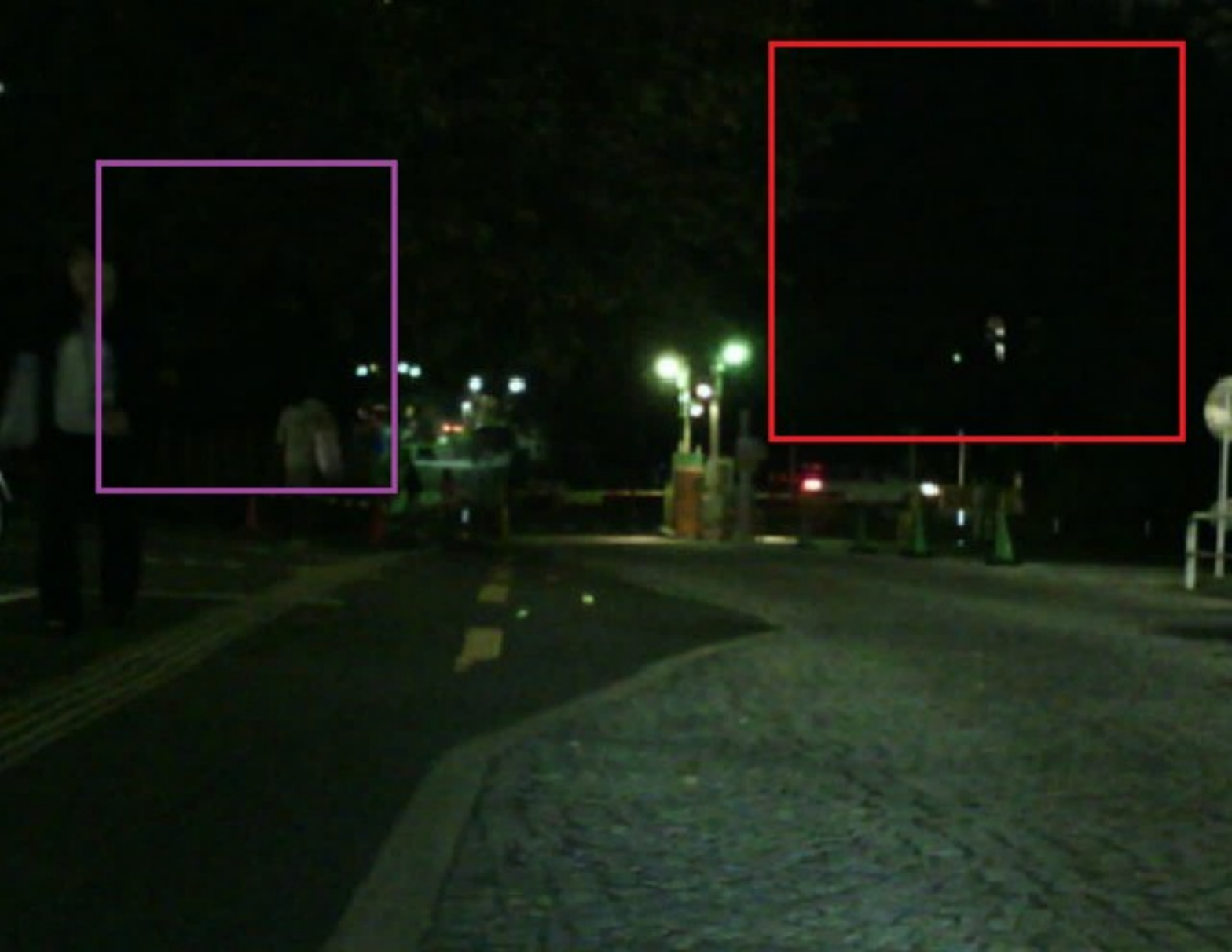}
    {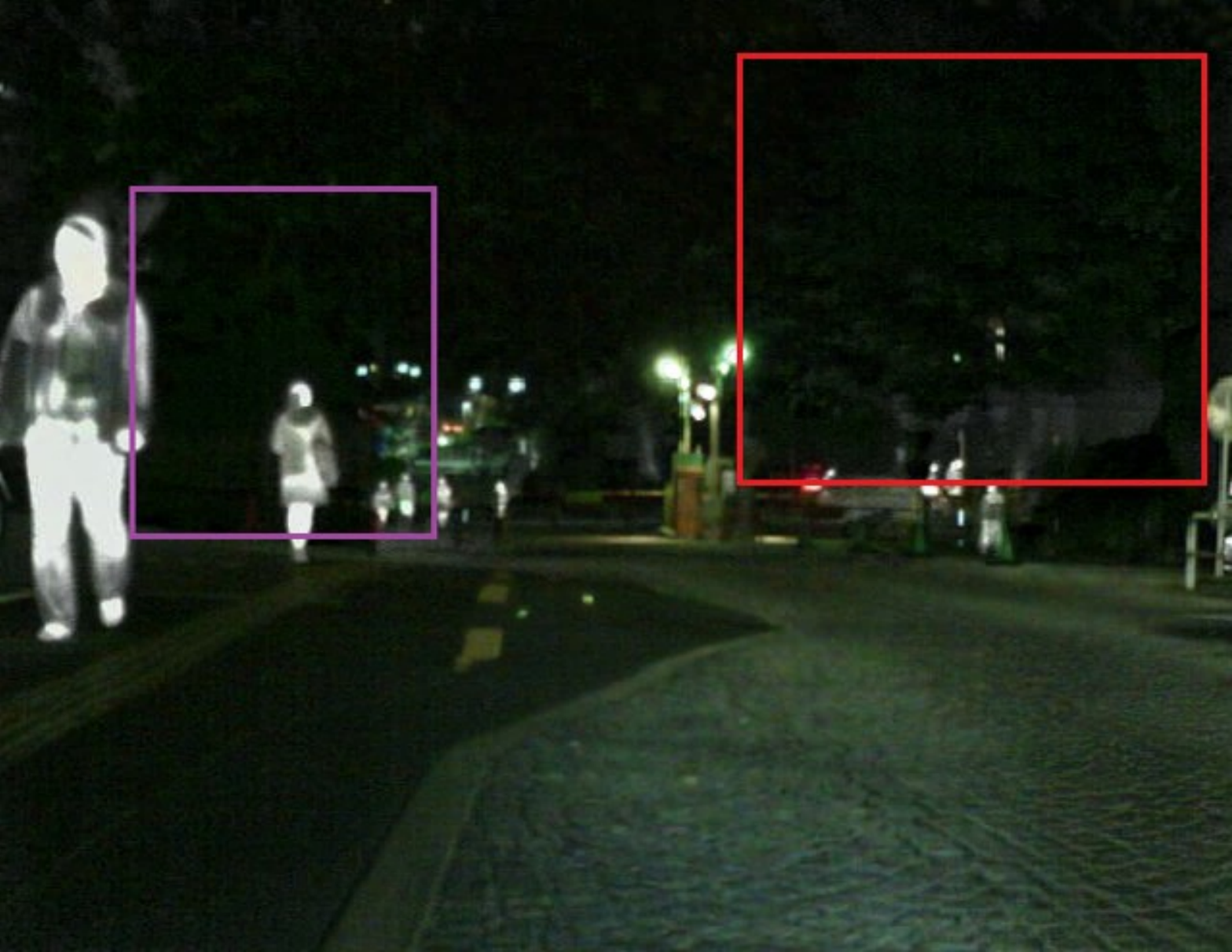}
    {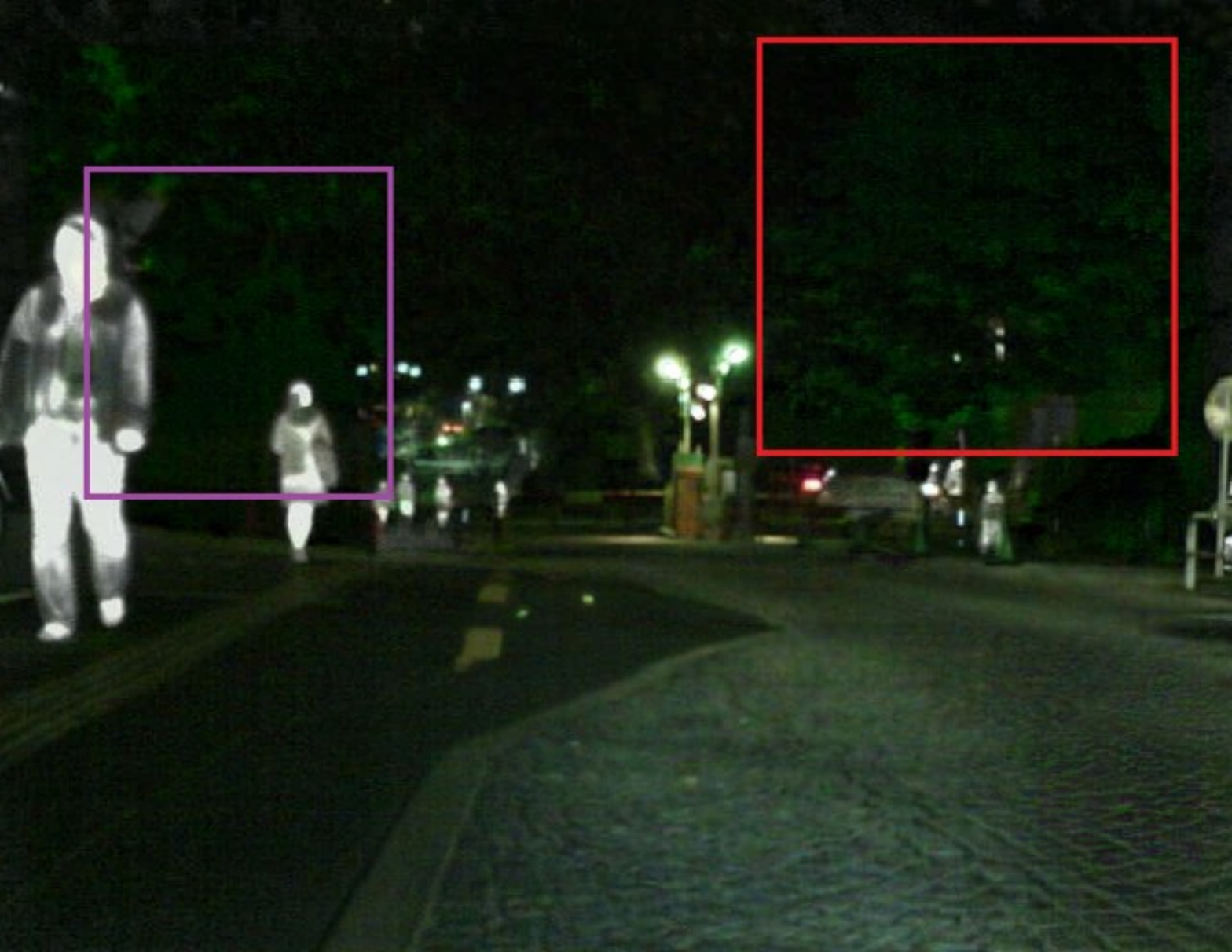}
    {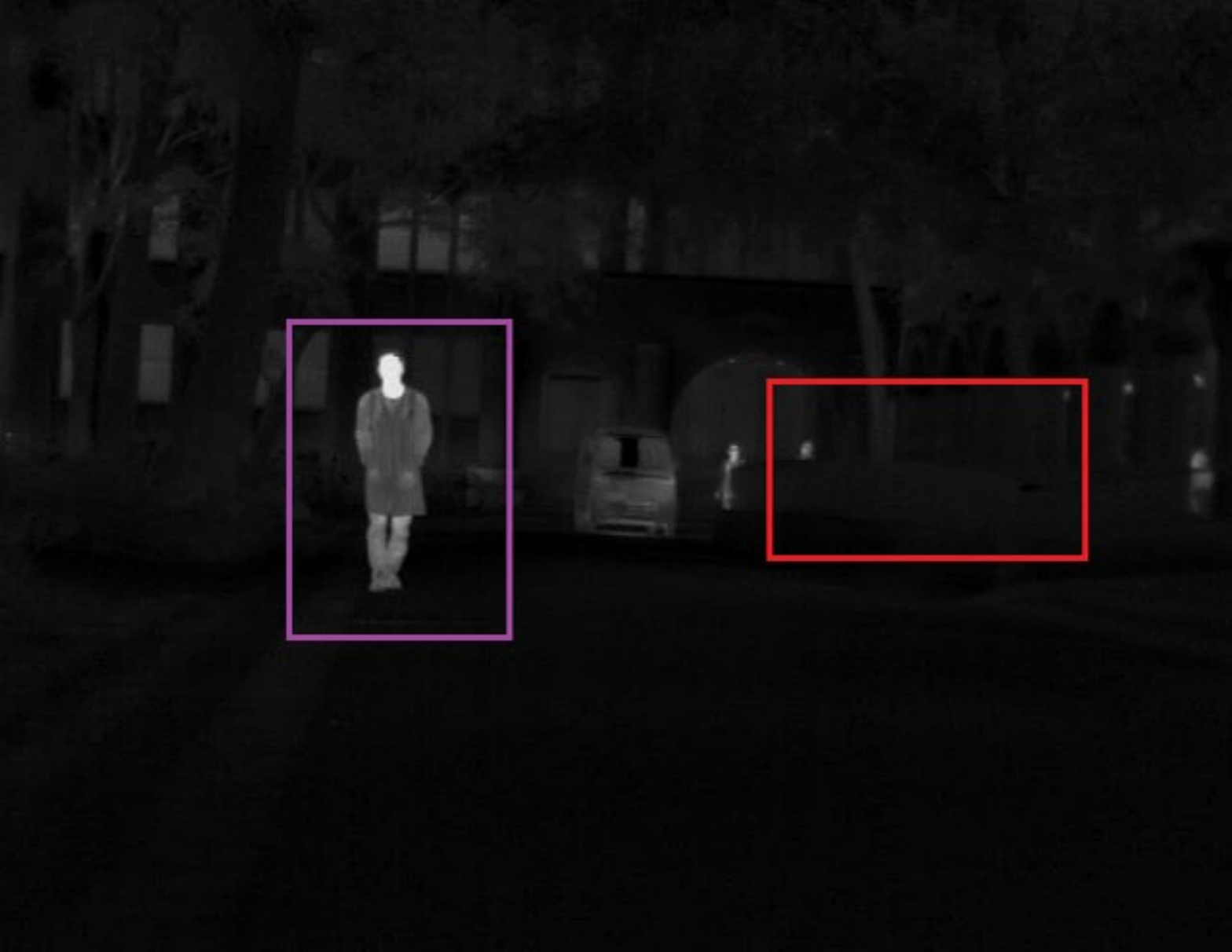}
    {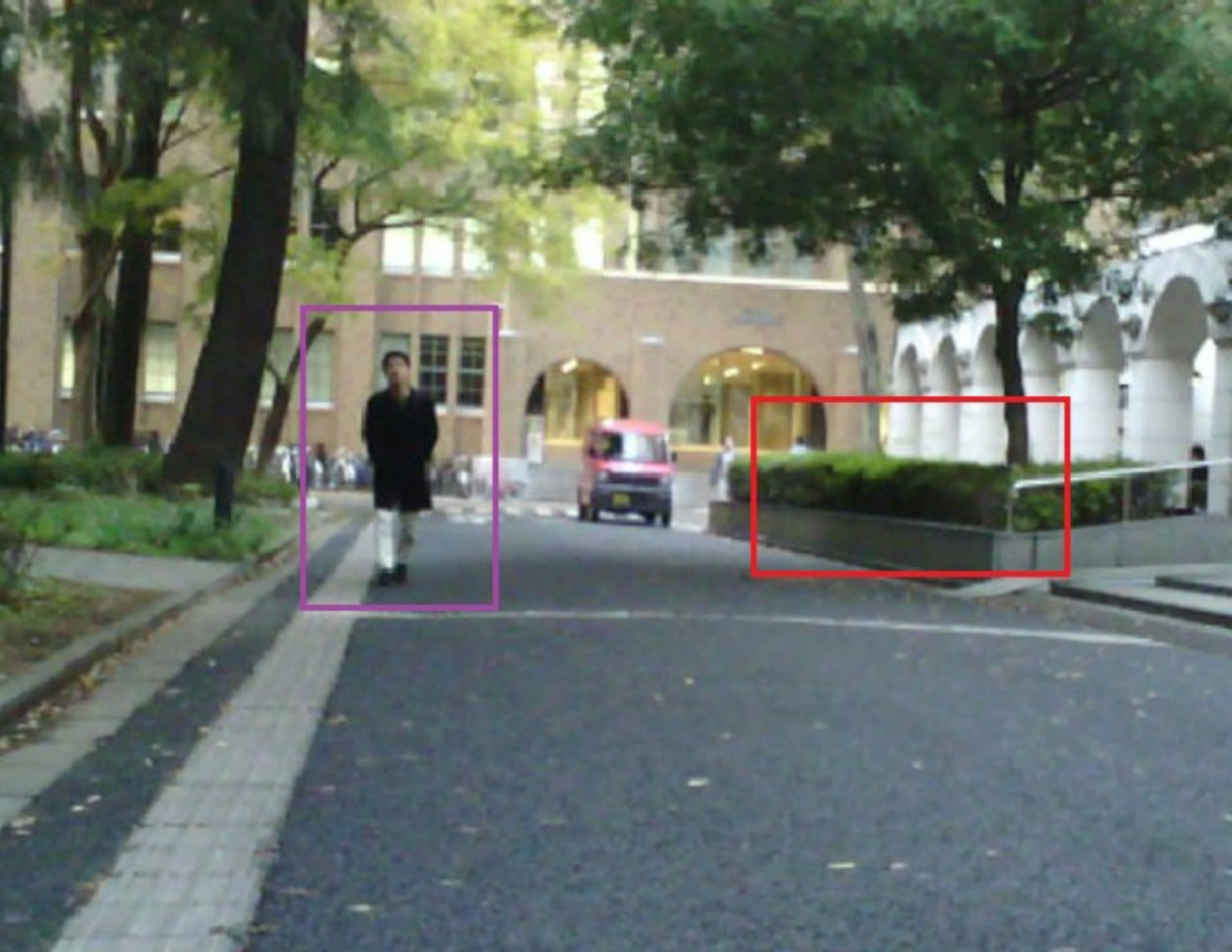}
    {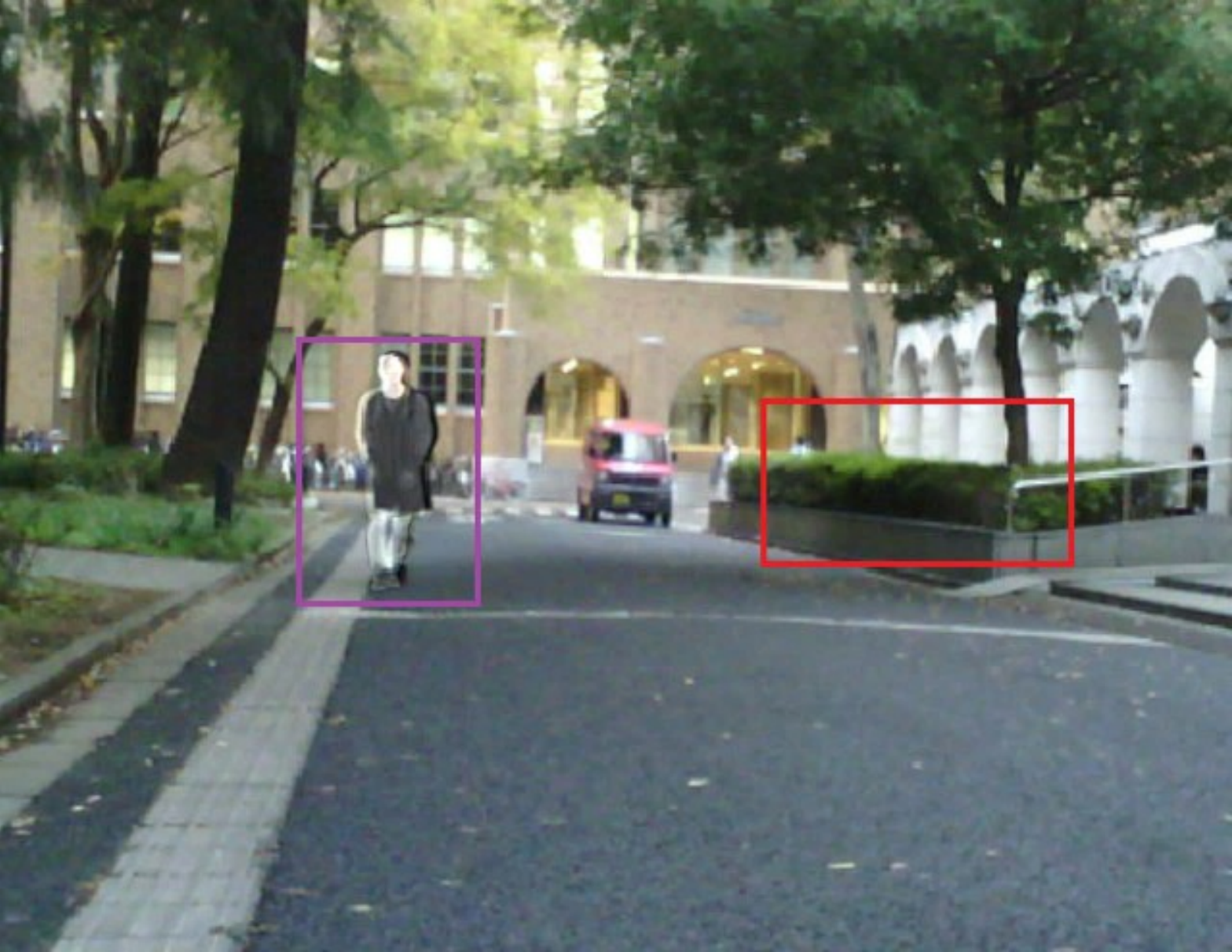}
    {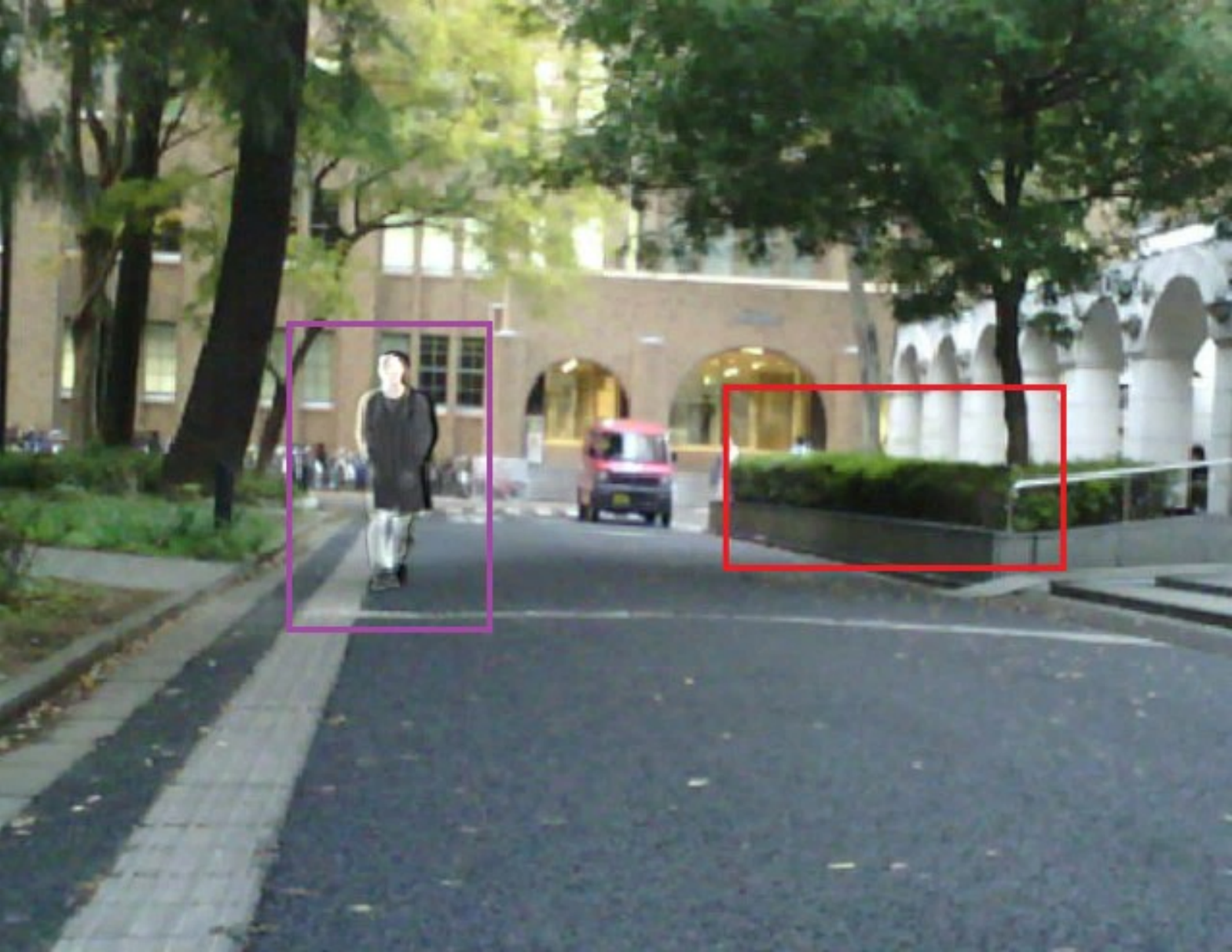}

  \vspace{5pt}

  \twocolrow
    {We're tackling the infrared-visible light image fusion challenge, dealing with visible images suffering from overexposure degradation.}
    {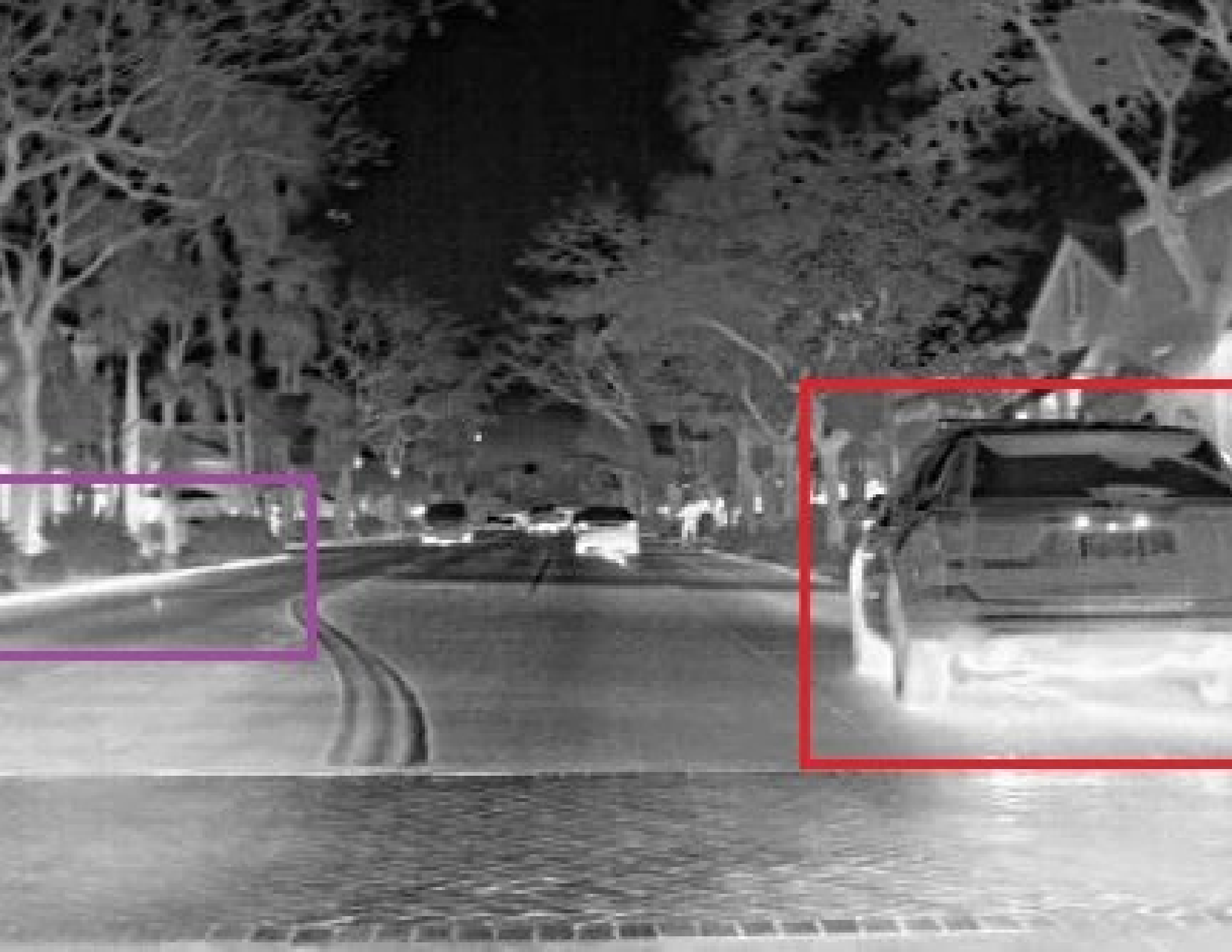}
    {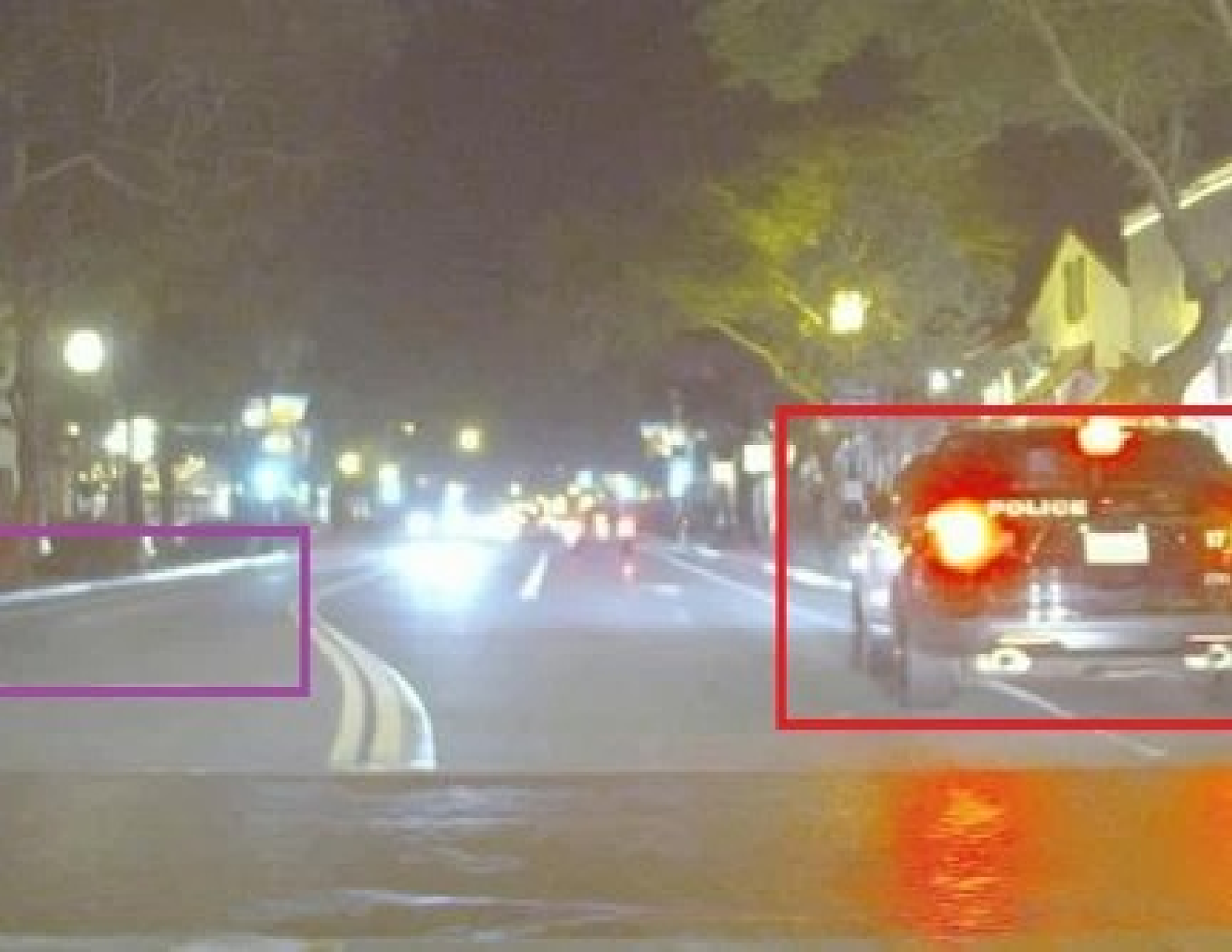}
    {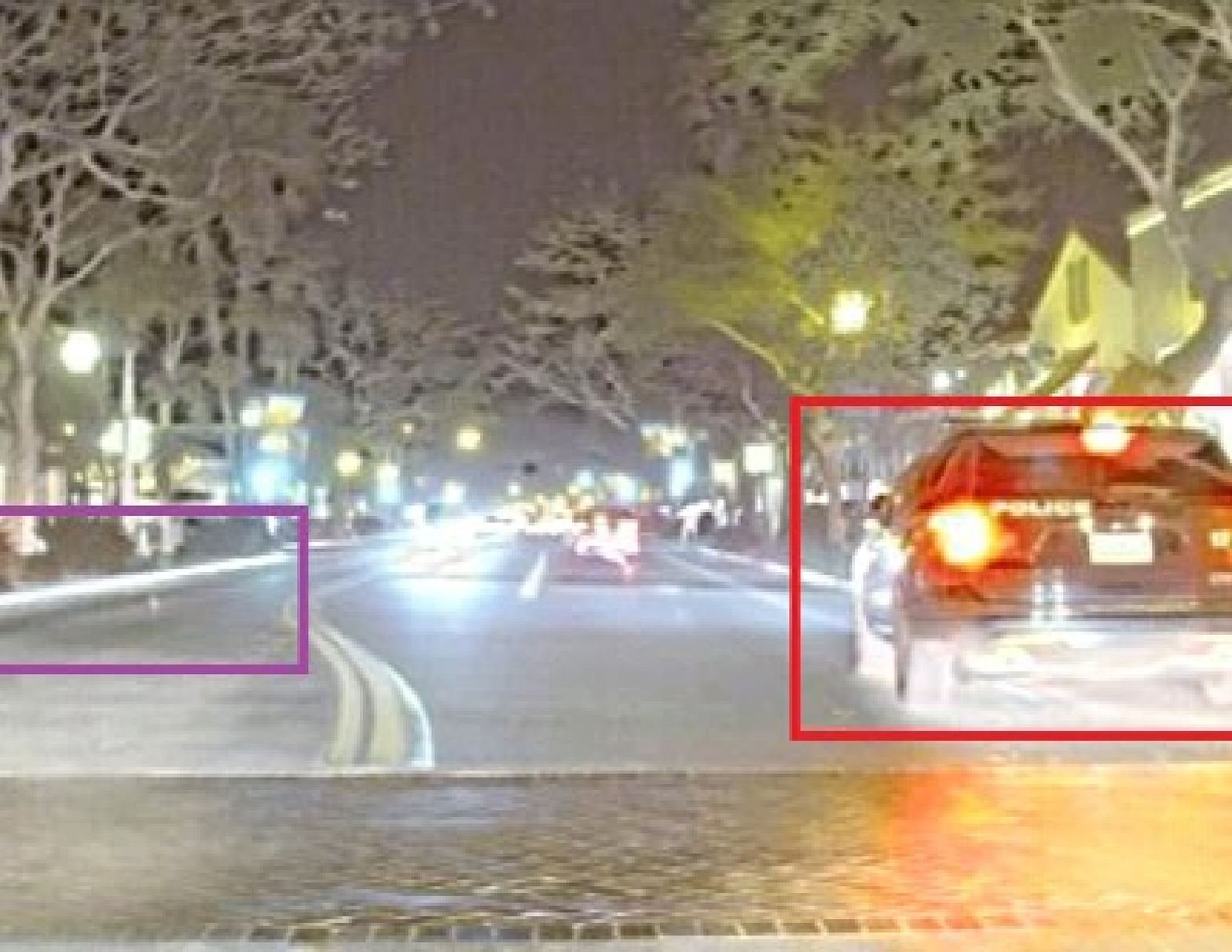}
    {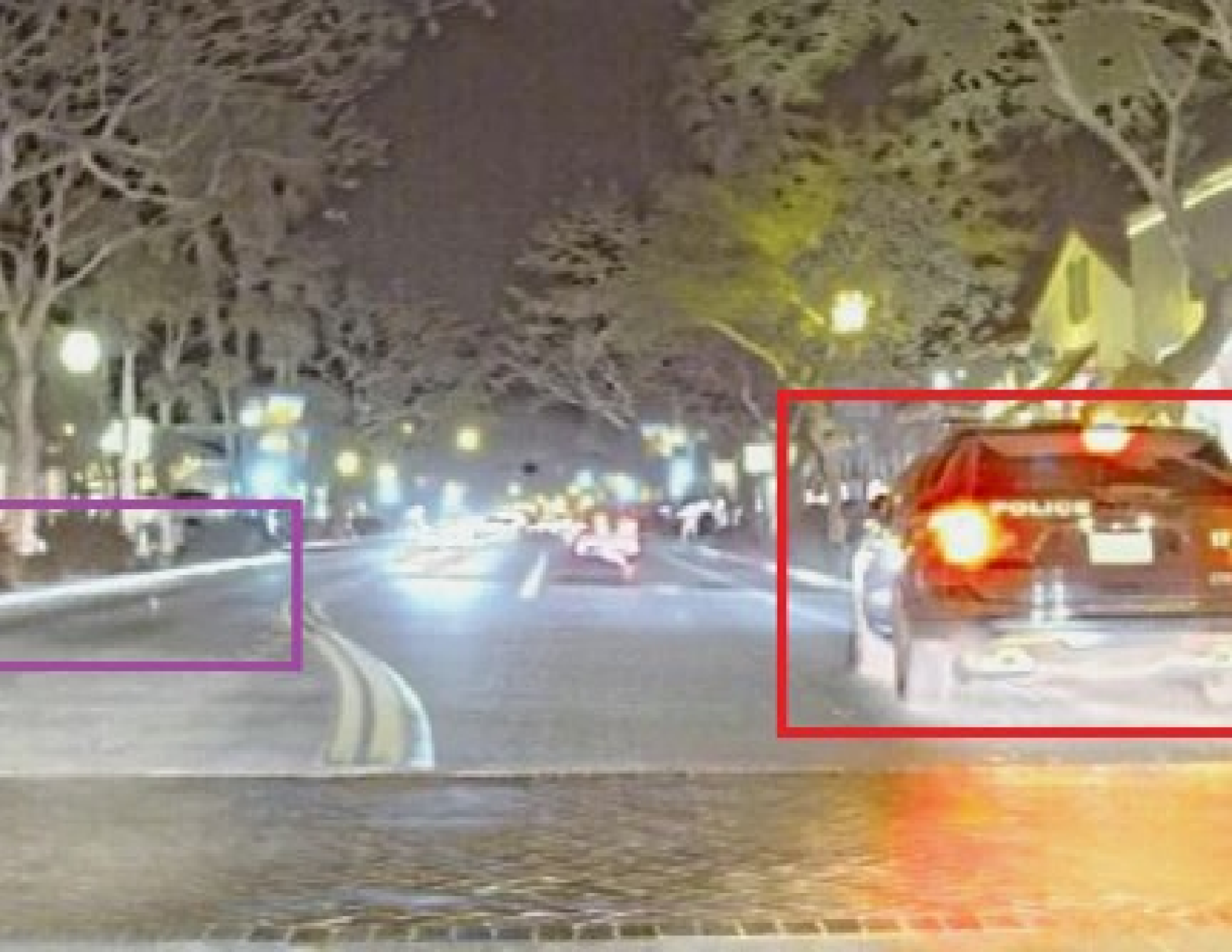}
    {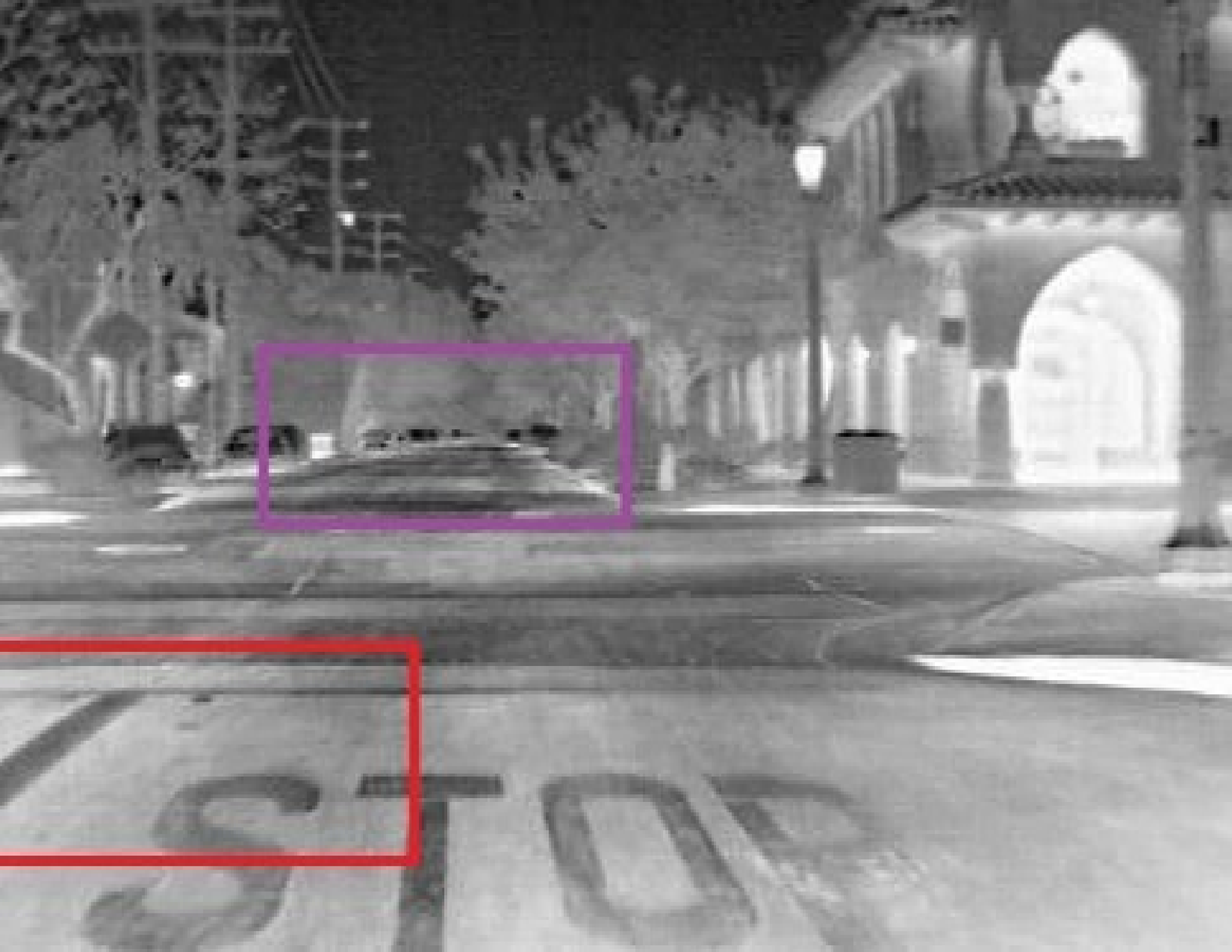}
    {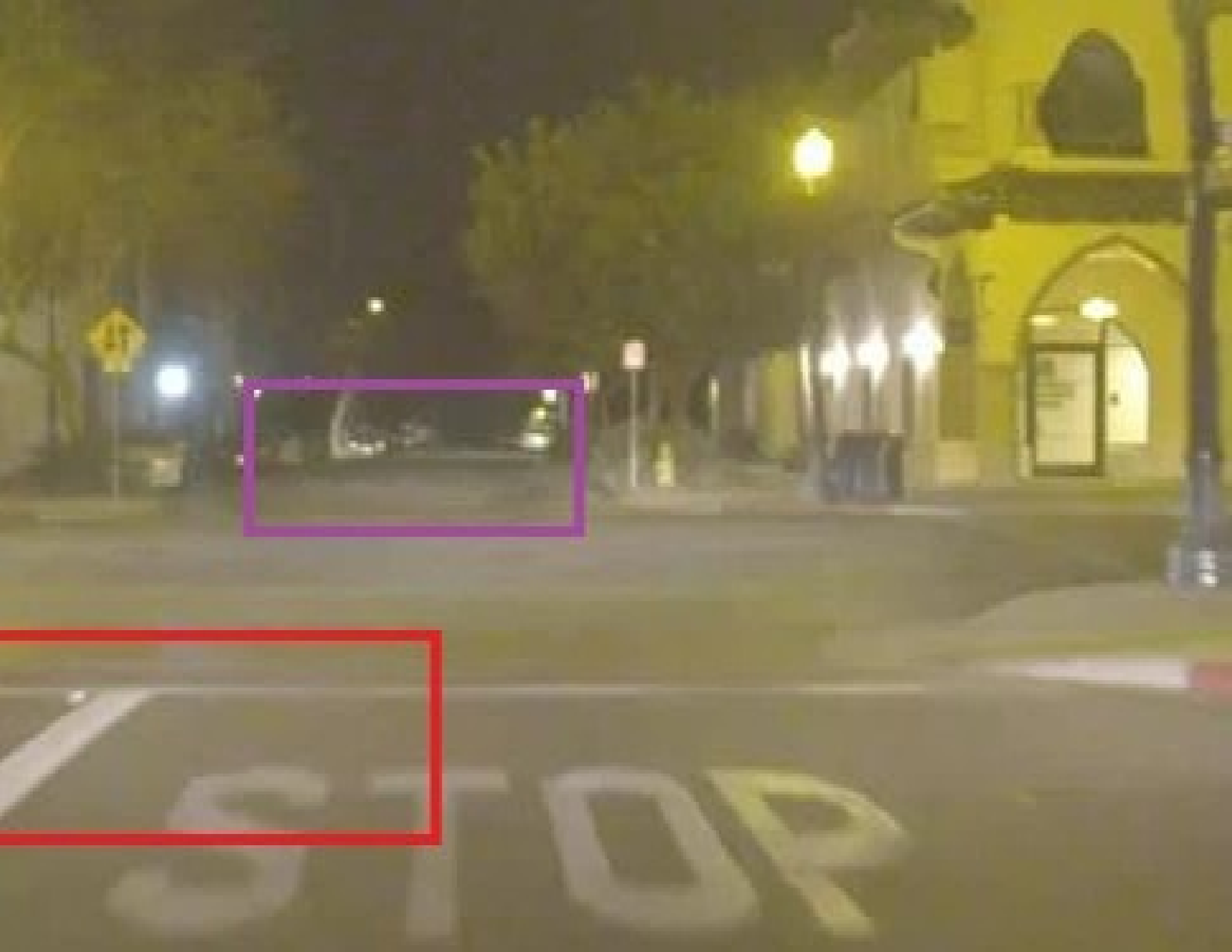}
    {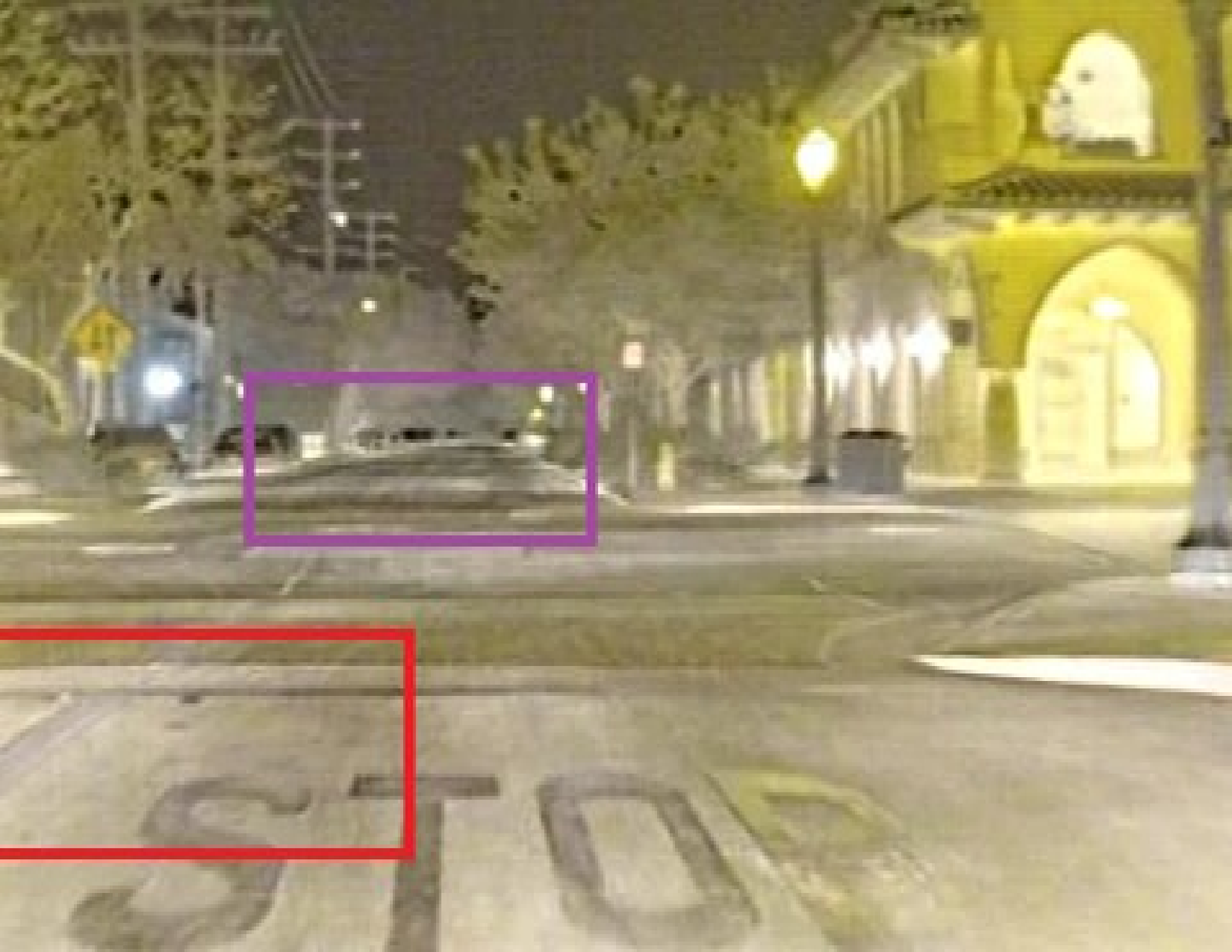}
    {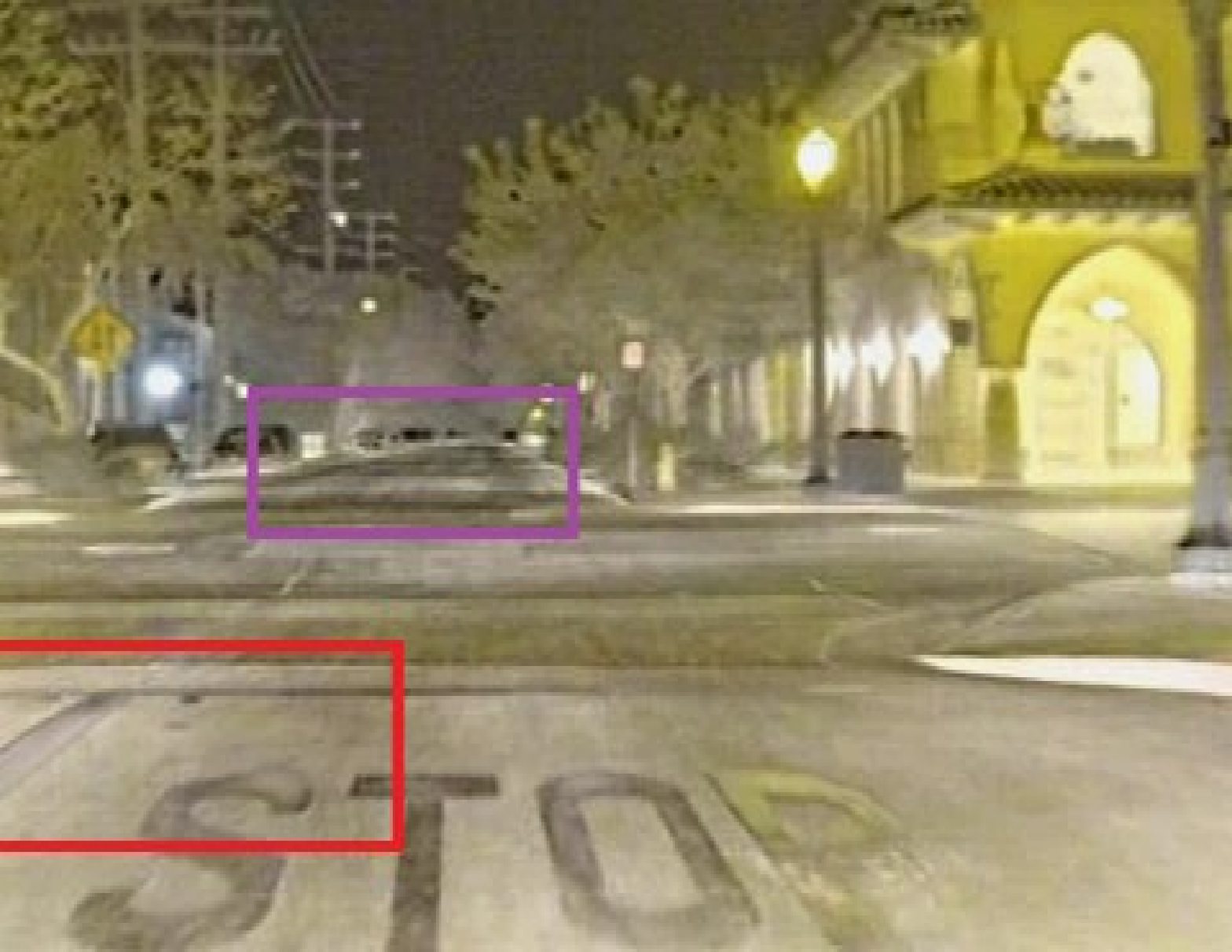}

  \caption{Qualitative comparison with the baseline method Text-IF ~\cite{yi2024text} on the four standard datasets. Degradations from top to bottom: Low Light (LLVIP), Infrared Low Contrast (MFNet), Low Light (MSRS), and Overexposure (RoadScene). For each dataset, Text-IF and our method use the same degradation-aware text prompt for a fair comparison.}
  \label{fig: qualitative results for standard datasets}
\end{figure}

\begin{table}[t]
\centering
\small
\setlength{\tabcolsep}{3.2pt}
\renewcommand{\arraystretch}{1.15}
\caption{Quantitative comparison of our method under degradation-aware text guidance with Text-IF and combinations of existing image restoration (eir.) and fusion methods on source images with various types of degradations (MSRS and LLVIP datasets: low-light visible images; MFNet: low-contrast
infrared images; RoadScene: overexposed visible images). (\textbf{Bold}: best performance)}
\label{tab:comparison_multi_dataset}
\resizebox{\textwidth}{!}{%
\begin{tabular}{c|ccc|ccc|ccc|ccc}
\toprule
\multirow{2}{*}{\textbf{Methods}}
& \multicolumn{3}{c|}{\textbf{MSRS Dataset}}
& \multicolumn{3}{c|}{\textbf{LLVIP Dataset}}
& \multicolumn{3}{c|}{\textbf{MFNet Dataset}}
& \multicolumn{3}{c}{\textbf{RoadScene Dataset}} \\

\cline{2-13}
& CLIP-IQA$\uparrow$ & EN$\uparrow$ & NIQE$\downarrow$
& EN$\uparrow$ & NIQE$\downarrow$ & MUSIQ$\uparrow$
& SD$\uparrow$ & EN$\uparrow$ & MUSIQ$\uparrow$
& SF$\uparrow$ & NIQE$\downarrow$ & BRISQUE$\downarrow$ \\
\midrule
\textit{eir.}+UMF-CMGR     & 0.101 & 6.316 & 3.738 & 7.087 & 3.891 & 47.543 & 23.684 & 5.414 & 34.113 & 11.047 & 3.792 & 32.485 \\
\textit{eir.}+TarDAL       & 0.082 & 5.855 & 4.750 & 7.042 & 3.659 & 41.735 & 33.454 & 6.142 & 25.120 & 11.789 & 3.667 & 32.436 \\
\textit{eir.}+ReCoNet      & 0.117 & 7.216 & 5.769 & 7.109 & 4.695 & 44.187 & 41.654 & 5.161 & 29.299 & 10.312 & 4.785 & 37.775 \\
\textit{eir.}+MURF         & 0.111 & 5.872 & 4.199 & 6.757 & 4.177 & 50.589 & 23.741 & 5.601 & 35.626 & 15.605 & 3.779 & 30.594 \\
\textit{eir.}+U2Fusion     & 0.127 & 6.724 & 3.997 & 7.439 & 3.969 & 48.481 & 33.940 & 5.740 & 34.255 & 18.006 & 4.215 & 34.577 \\
\textit{eir.}+MetaFusion   & 0.106 & \textbf{7.302} & 3.584 & 7.495 & 3.722 & 49.620 & 42.026 & 6.665 & 34.762 & \textbf{26.653} & 3.473 & 29.500 \\
\textit{eir.}+DDFM         & 0.094 & 6.723 & \textbf{3.465} & 7.150 & 5.184 & 35.933 & 30.465 & 6.480 & 26.902 & 10.493 & 3.717 & 32.334 \\
Text-IF ~\cite{yi2024text} & 0.132 & 7.172 & 3.708 & 7.391 & 3.502 & 48.625 & 43.933 & \textbf{6.683} & \textbf{35.650} & 17.766 & 3.342 & 29.021 \\
\textbf{IT-TextFusion (Ours)}  & \textbf{0.145} & 6.689 & 3.906 & \textbf{7.502} & \textbf{3.336} & \textbf{54.067} & \textbf{65.141} & 4.137 & 33.942 & 15.332 & \textbf{3.330} & \textbf{27.571} \\
\bottomrule
\end{tabular}%
}
\end{table}

\subsection{Comparison with a Fixed Generic Prompt}
To evaluate the effect of degradation-specific prompt content, we first conduct experiments using a fixed generic prompt, i.e., \emph{``This is an infrared and visible image fusion task.''} Under this setting, the textual branch remains active, while the input prompt contains no degradation-aware or scene-specific semantic information. This setting is also consistent with our primary baseline Text-IF~\cite{yi2024text}. Therefore, rather than treating this experiment as a text-free visual-only baseline, we regard it as an experiment using a fixed generic prompt and use it for controlled comparison. In this way, we can investigate whether the proposed architecture itself provides improved fusion capability before introducing explicit degradation-aware text guidance.

Tab.~\ref{tab:quant_msrs_llvip_roadscene} shows that the proposed method achieves competitive performance when using a fixed generic prompt, with different advantages across datasets and metrics. On MSRS, the proposed method improves SCD, EN, VIFF, and $Q^{AB/F}$ compared with Text-IF, although its SD is lower. On LLVIP, improvements are observed in SD, EN, VIFF, and $Q^{AB/F}$, whereas SCD decreases slightly. On RoadScene, the proposed method obtains higher SD and $Q^{AB/F}$, while Text-IF retains higher SCD and VIFF and both methods achieve the same EN. These results indicate that the architectural modifications alter the balance among different fusion properties rather than providing uniform improvements on every metric.

Compared with Text-IF, the proposed method introduces an iterative text-image interaction mechanism, where the same global text embedding modulates features across the four hierarchical decoding stages and is injected again during residual refinement.
Even with the global text embedding produced from a fixed generic prompt, this repeated conditioning enables global semantic regulation at multiple feature resolutions and strengthens the interaction between progressively decoded fusion features and reintroduced encoder features. These results suggest that the performance improvement of the proposed framework does not rely solely on degradation-aware textual semantics; the hierarchical text-conditioned interaction and subsequent residual refinement themselves also contribute to the fusion performance.

\subsection{Comparison with Degradation-Aware Text Guidance}
\subsubsection{Evaluation on Standard Datasets}
\label{subsec: eva standard_datasets}
After evaluating the model using a fixed generic prompt, we further investigate its performance with degradation-aware text guidance, where the input prompts explicitly describe the degradation characteristics of the source images. In real-world scenarios, source images often suffer from various types of degradations, such as poor illumination, noise, and low contrast. Among the compared methods introduced above, Text-IF ~\cite{yi2024text} is the only framework that explicitly incorporates textual guidance and is designed to handle diverse degradation conditions in a unified manner. 
In contrast, the other comparison methods are conventional image fusion approaches that rely solely on visual information and are not specifically designed to address degraded inputs.
Therefore, to ensure a fair comparison under degradation scenarios, we follow common practice by combining existing image fusion methods with appropriate image restoration models as preprocessing steps. SOTA image restoration models for different degradations
include URetinex-Net ~\cite{wu2022uretinex} for low-light image enhancement,
AirNet ~\cite{li2022all} for contrast enhancement, GDID ~\cite{chen2023masked} for
denoising, and LMPEC ~\cite{afifi2021learning} for overexposure correction.

\textbf{Quantitative Comparison.}
The quantitative results on the four standard infrared-visible fusion datasets are reported in Tab. ~\ref{tab:comparison_multi_dataset}.
Overall, the proposed method exhibits different advantages and trade-offs across datasets and evaluation metrics, rather than uniformly outperforming all competing methods.

On MSRS, IT-TextFusion achieves the highest CLIP-IQA score (0.145), indicating superior performance under this no-reference perceptual-quality metric.
However, its EN and NIQE are not the best among the compared methods, indicating a trade-off between semantic/perceptual quality and conventional information or naturalness measures.

On LLVIP, IT-TextFusion achieves the highest EN ($7.502$) and MUSIQ
($54.067$), together with the lowest NIQE ($3.336$). These results
indicate improvements in information content and selected
no-reference perceptual-quality criteria, although the proposed method
does not achieve the best result for every fusion metric.

On MFNet, IT-TextFusion obtains the highest SD ($65.141$), indicating
stronger global intensity variation. However, its EN ($4.137$) is
lower than that of Text-IF ($6.683$), and its MUSIQ ($33.942$) is also
lower than the best competing score. Therefore, the main
advantage on MFNet is reflected in global contrast rather than in
entropy or no-reference perceptual quality.

On RoadScene, the proposed method achieves the lowest NIQE ($3.330$)
and BRISQUE ($27.571$), whereas its SF ($15.332$) is lower than the
best competing score. Thus, the method shows advantages in
selected no-reference image-quality criteria but does not dominate all
fusion metrics.

\begin{figure}[!t]
  \centering

  \begin{minipage}[t]{0.495\textwidth}
    \onegroupEMS{This is the infrared-visible light fusion task, where visible images are affected by blur degradation.}
      {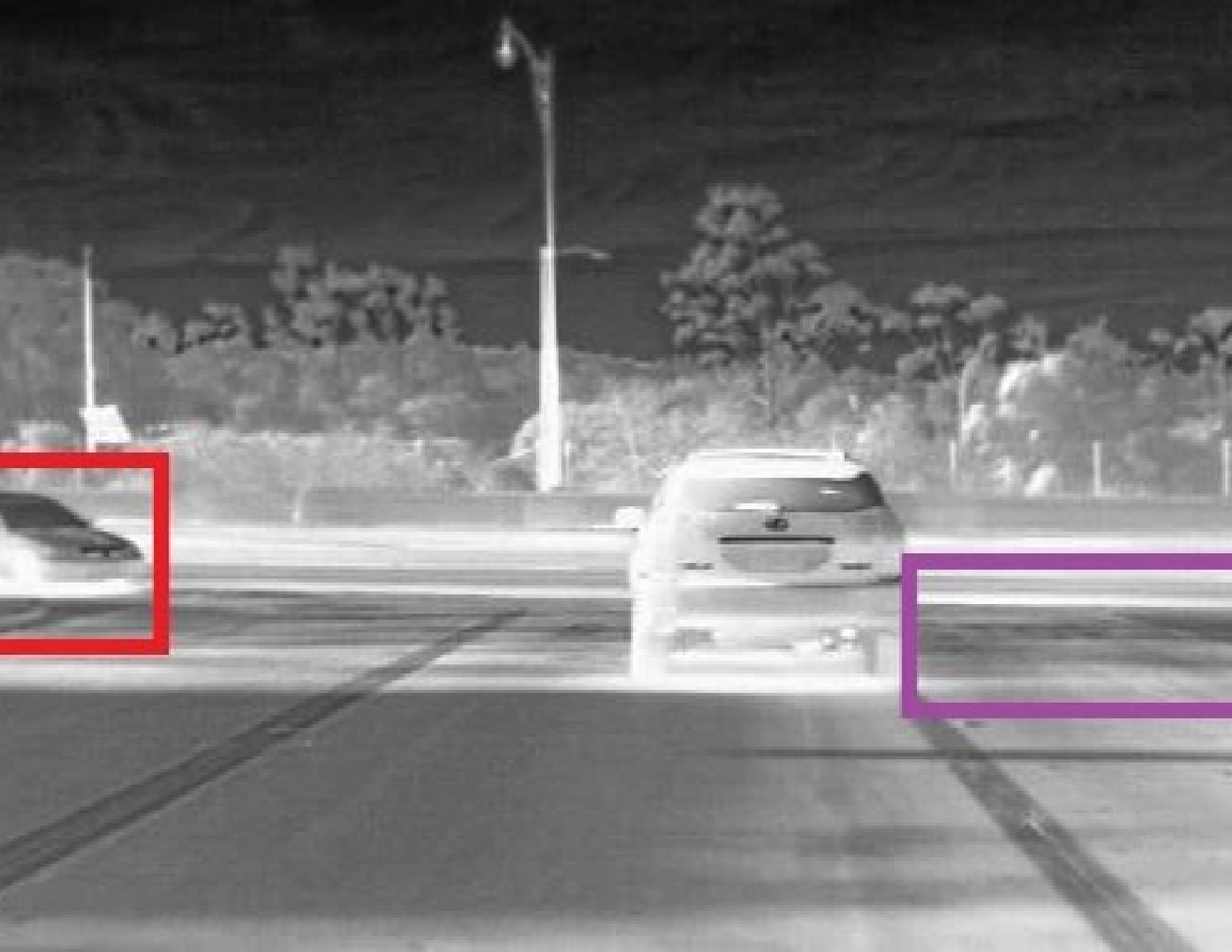}
      {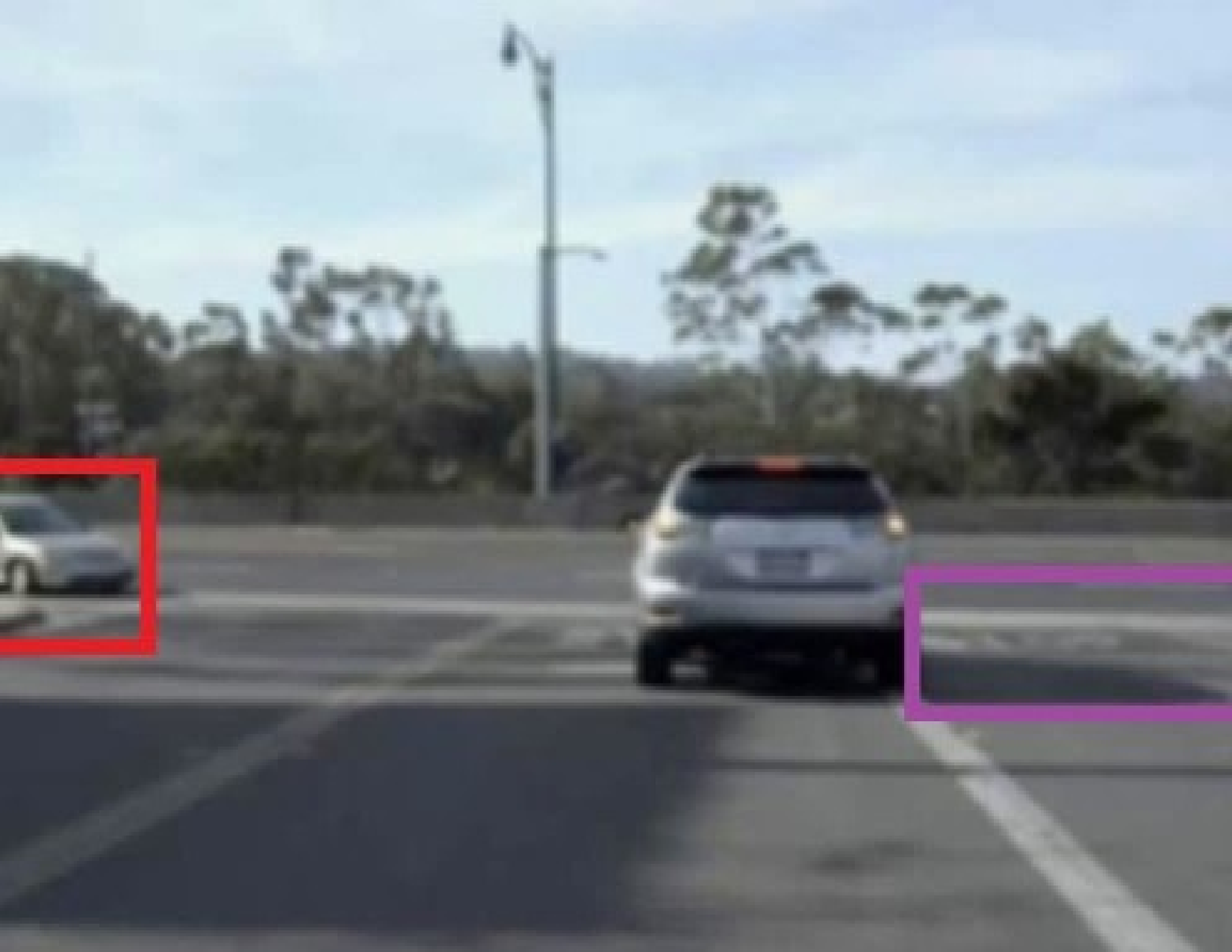}
      {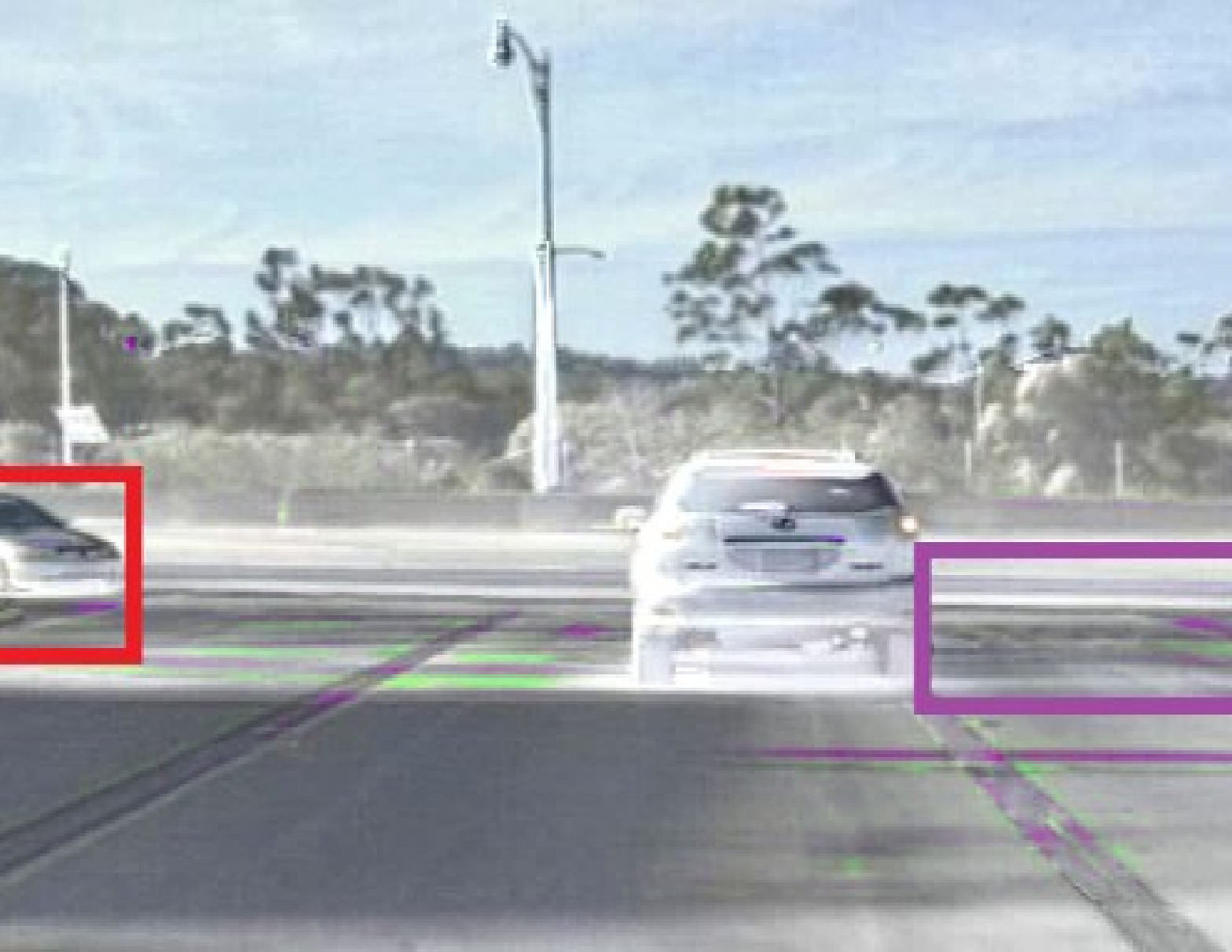}
      {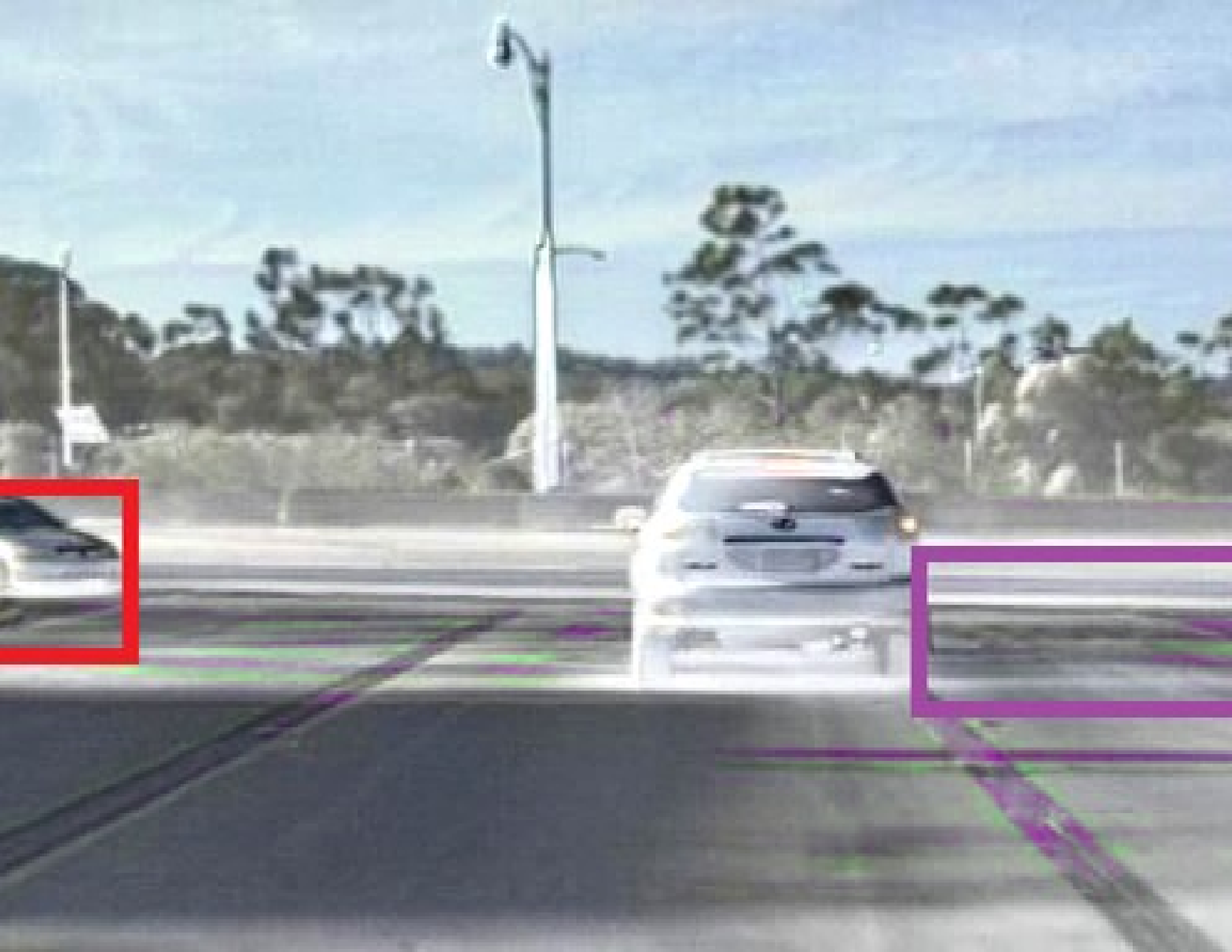}

    \vspace{14pt}

    \onegroupEMS{This is the infrared visible light fusion task. Visible images have the overexposure degradation.}
      {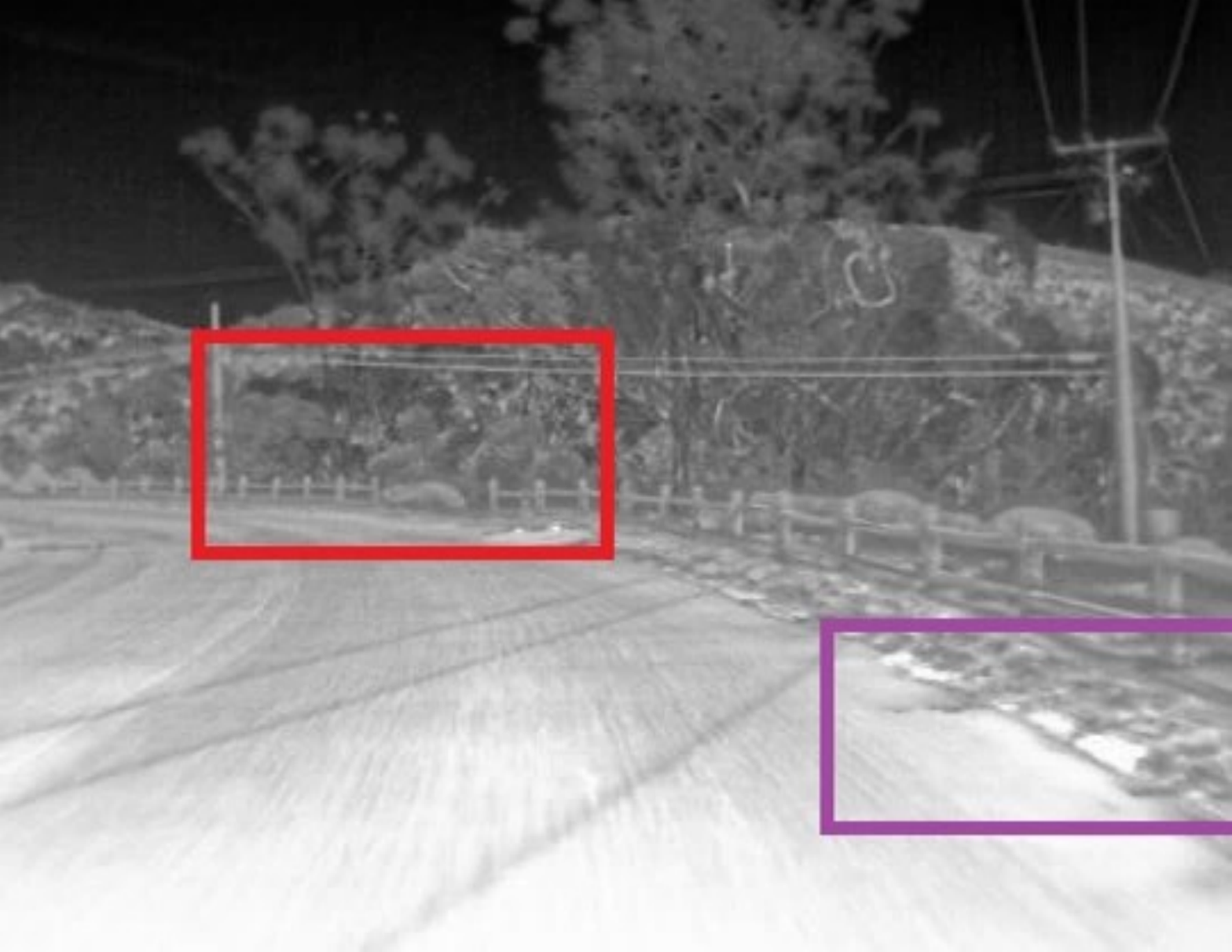}
      {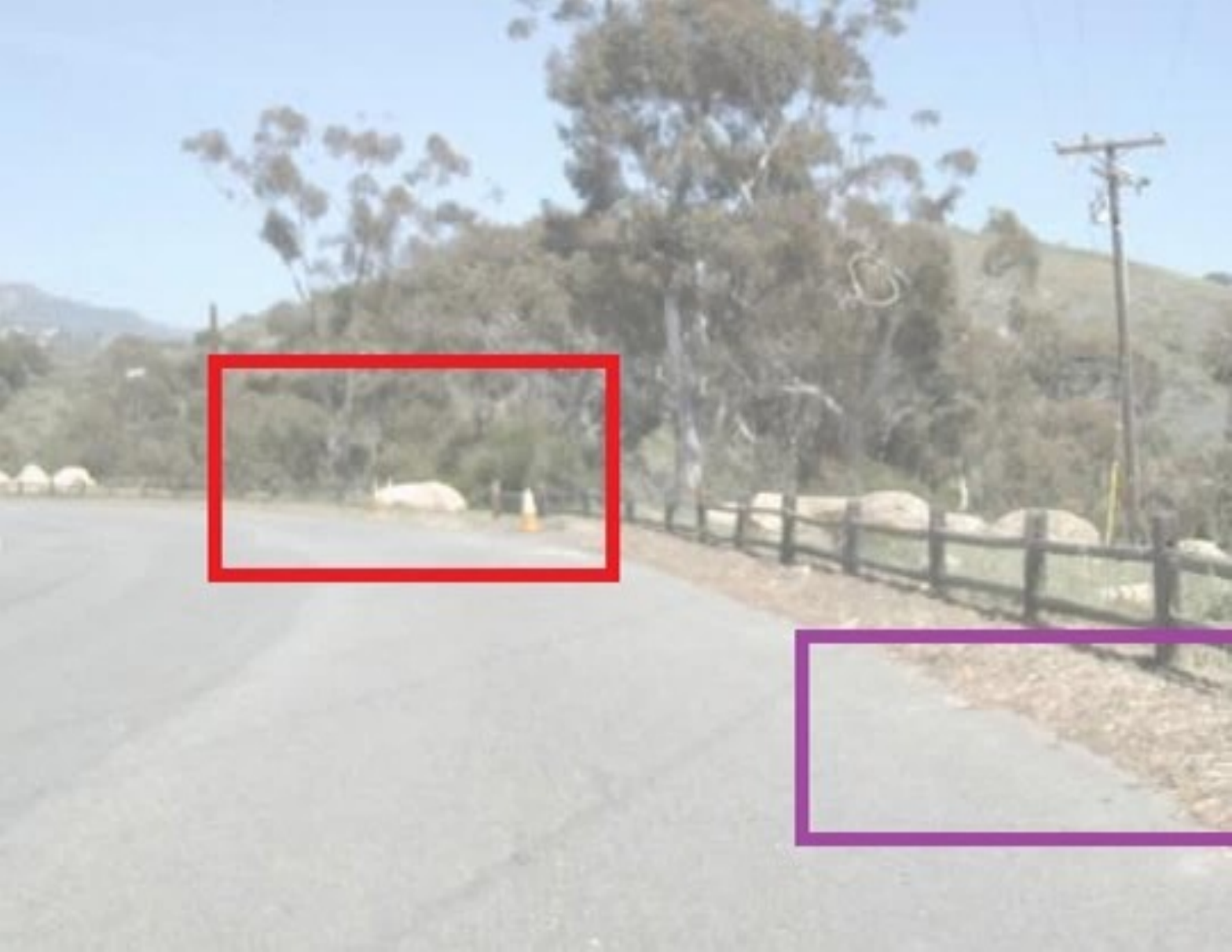}
      {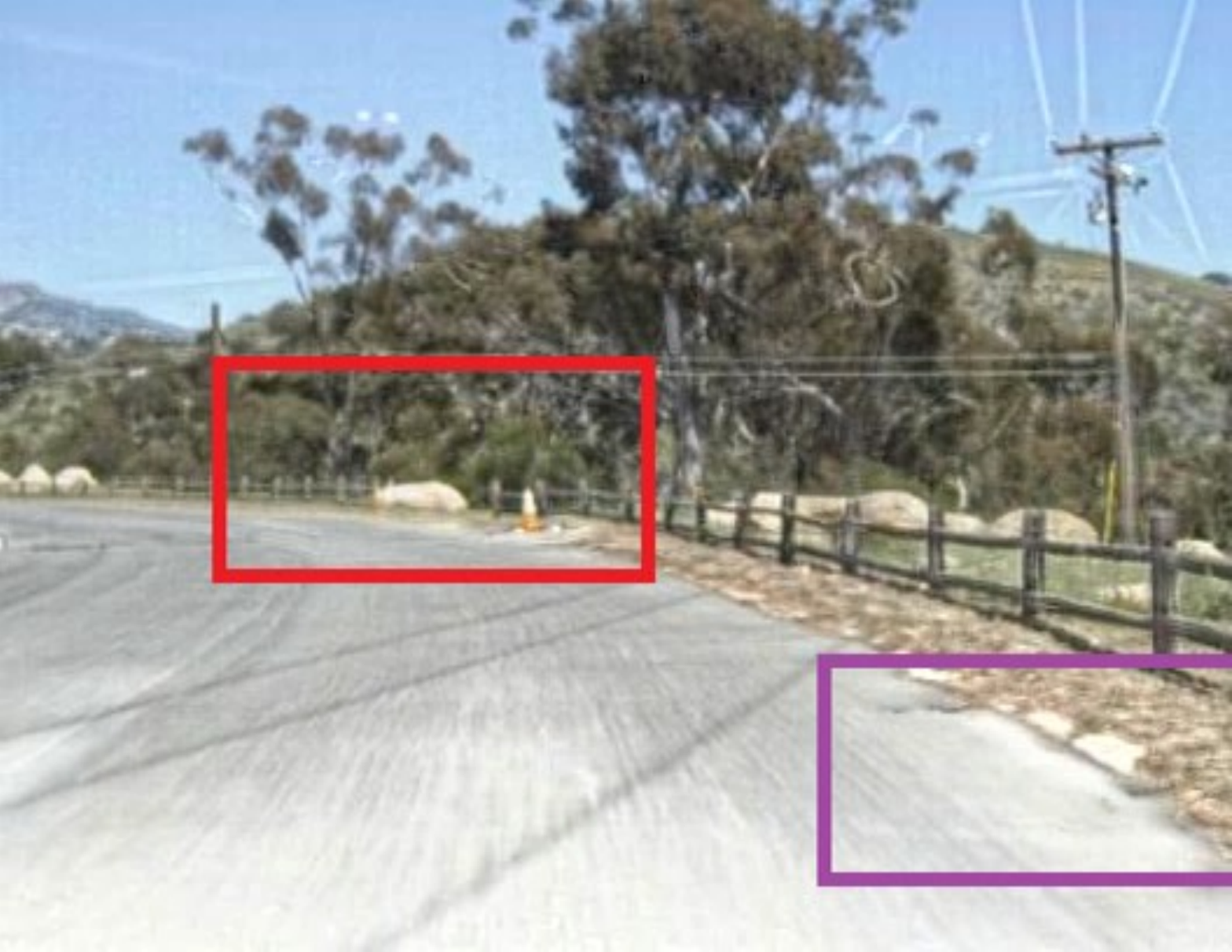}
      {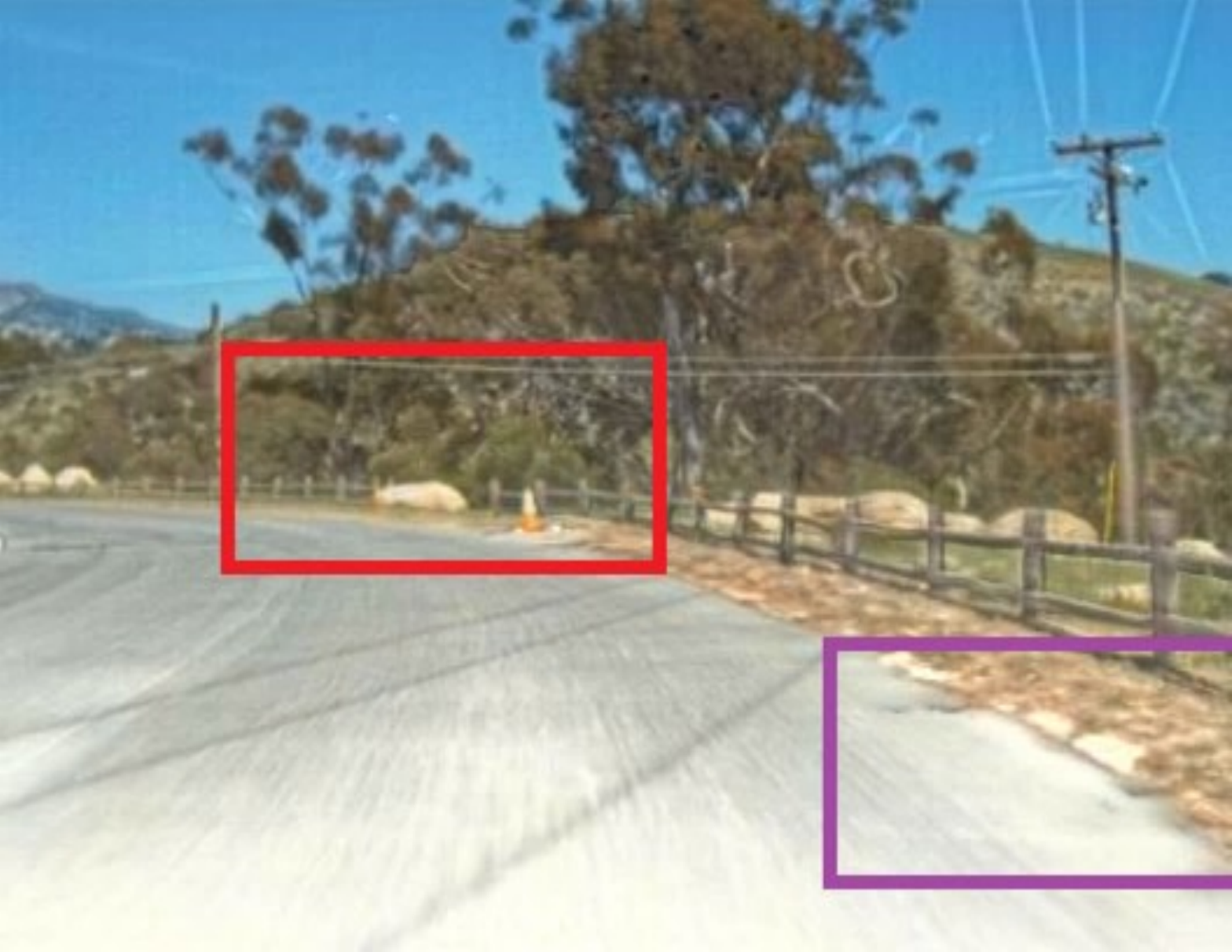}

    \vspace{14pt}

    \onegroupEMS{This is the infrared-visible light fusion task, where visible images are affected by haze degradation.}
      {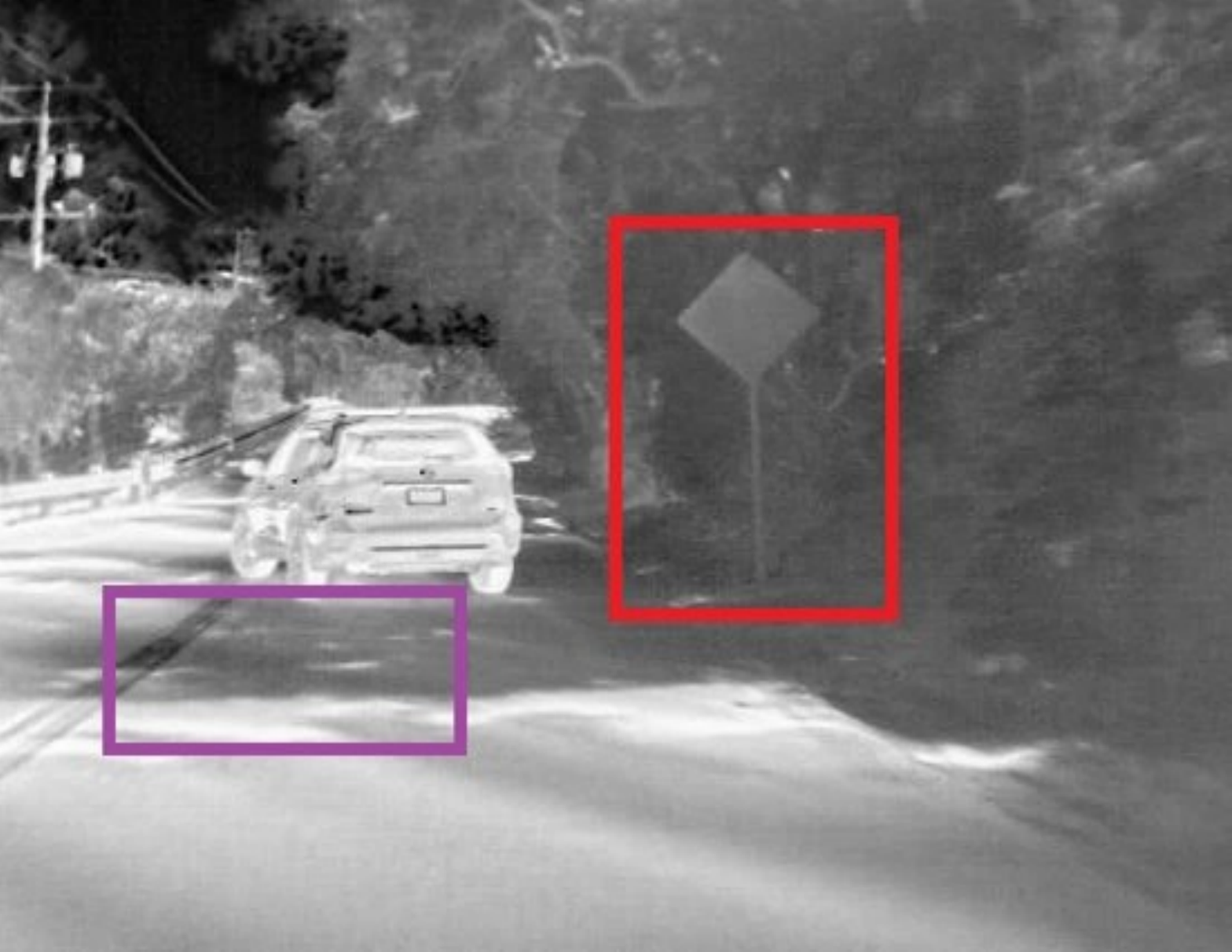}
      {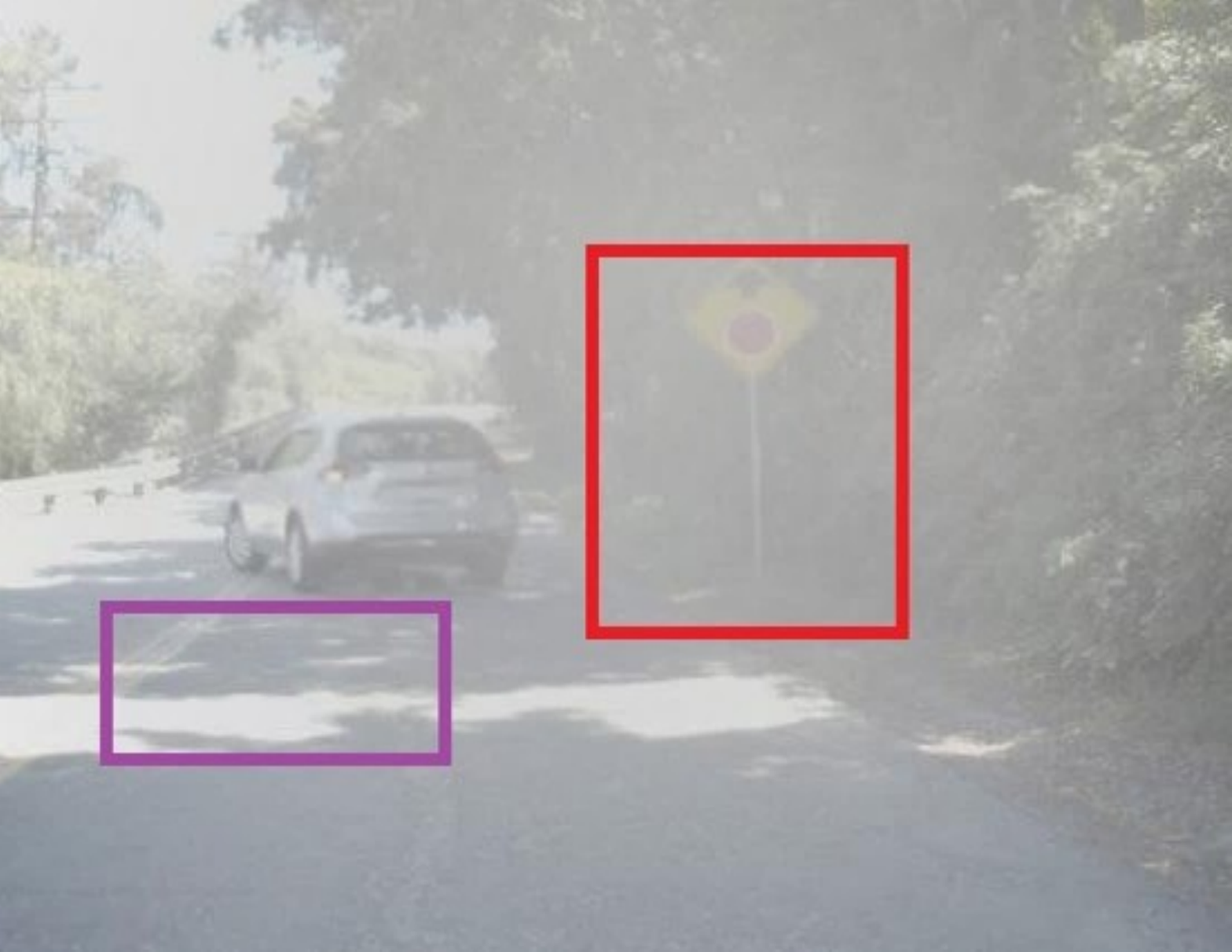}
      {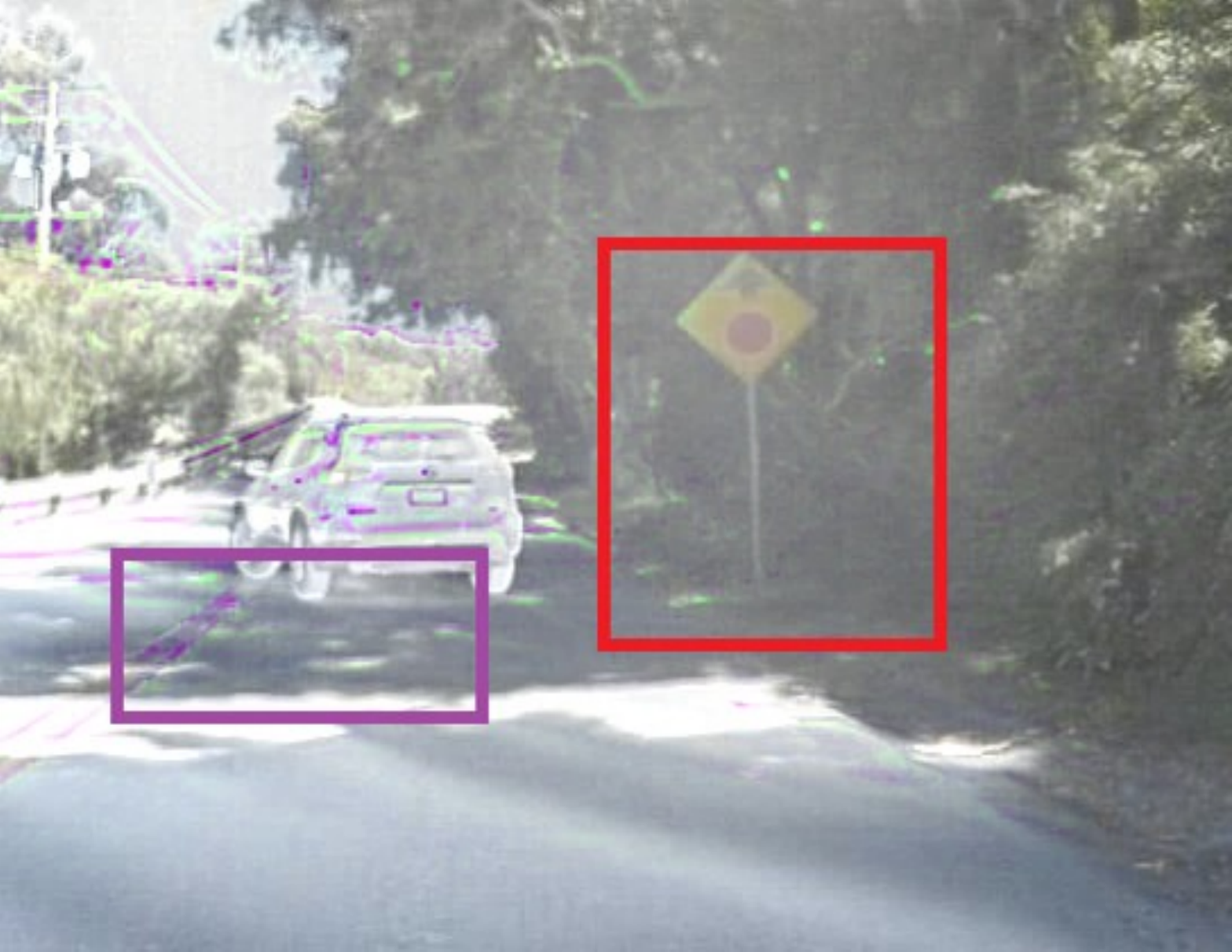}
      {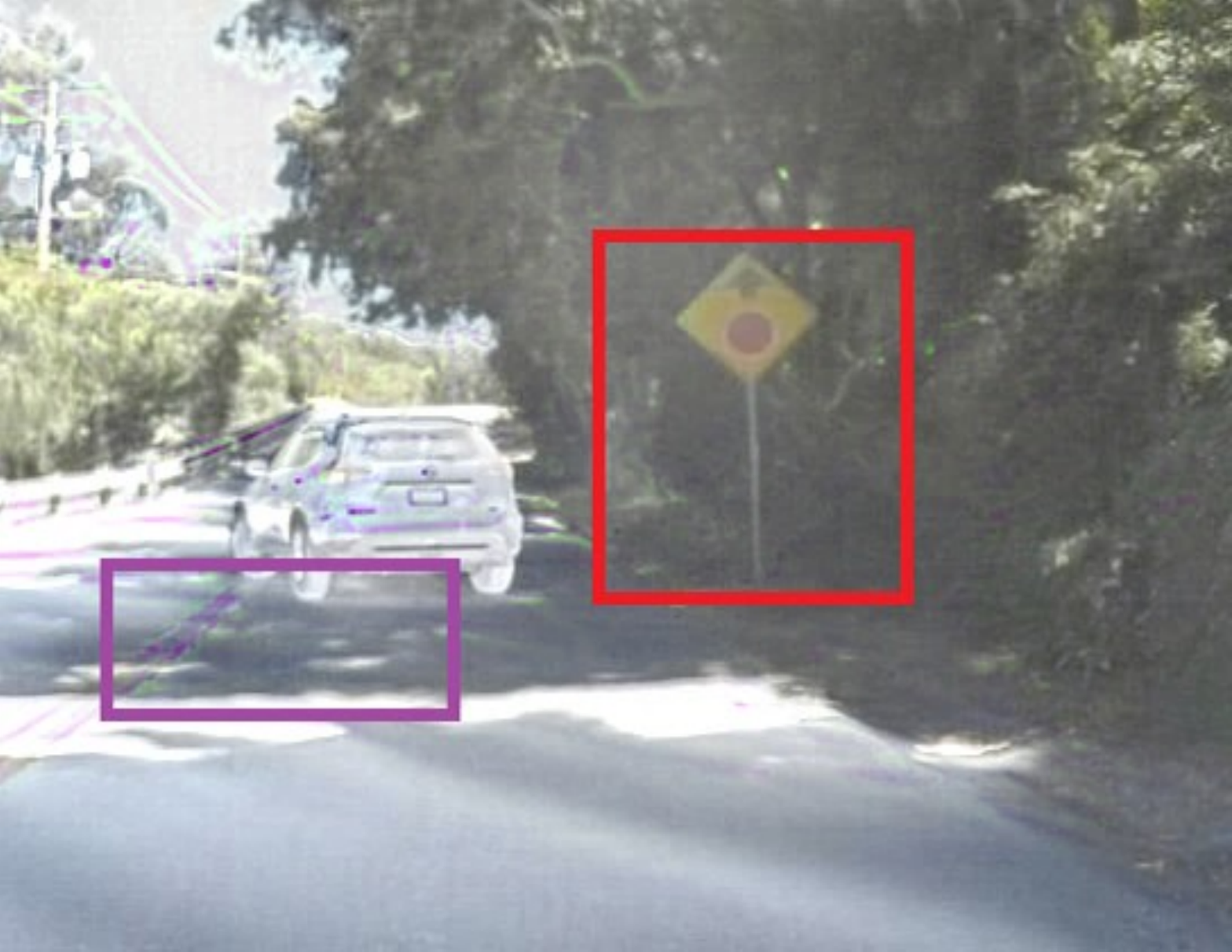}
  \end{minipage}%
  \hspace{0.01\textwidth}%
  \begin{minipage}[t]{0.495\textwidth}
    \onegroupEMS{This is the infrared visible light fusion task. Infrared images have the low contrast degradation.}
      {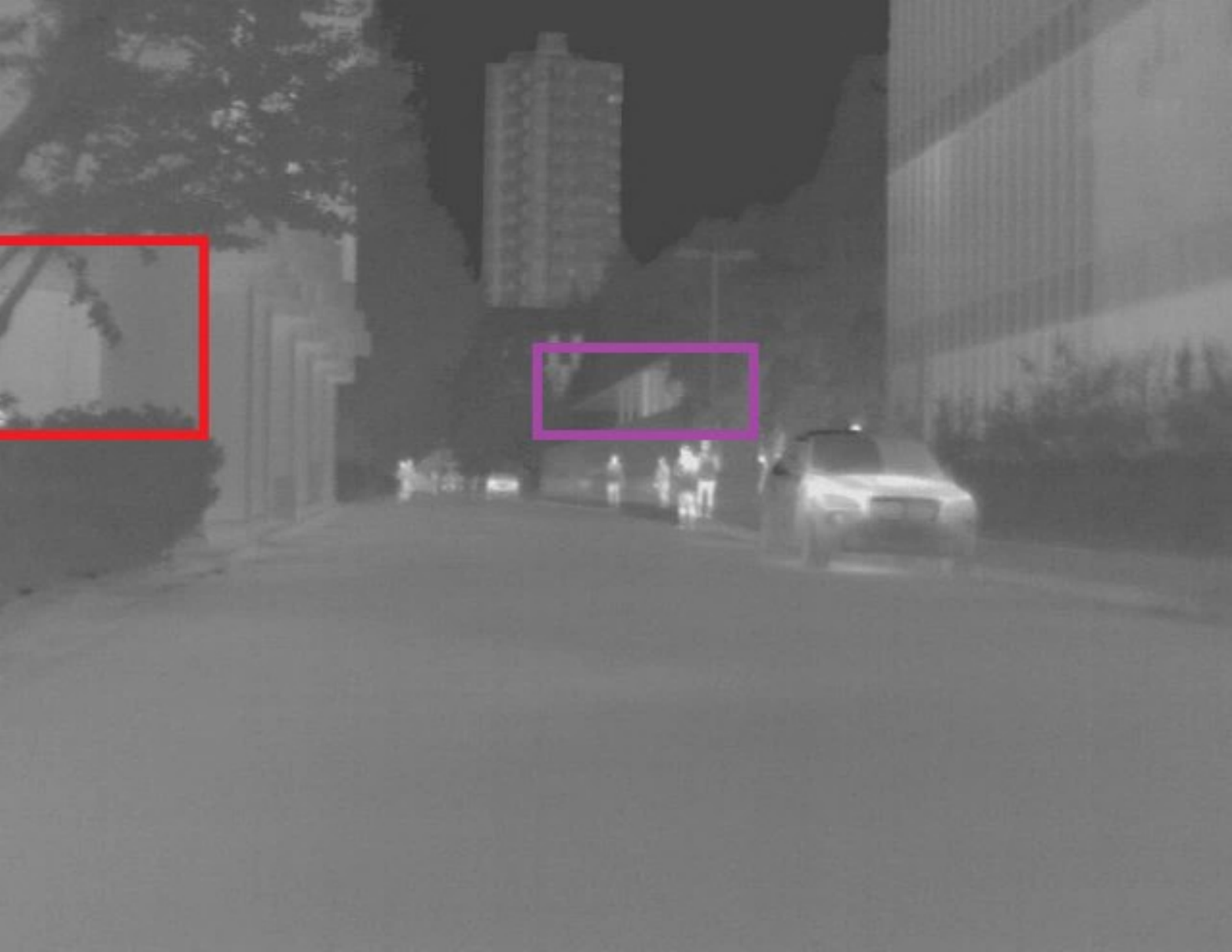}
      {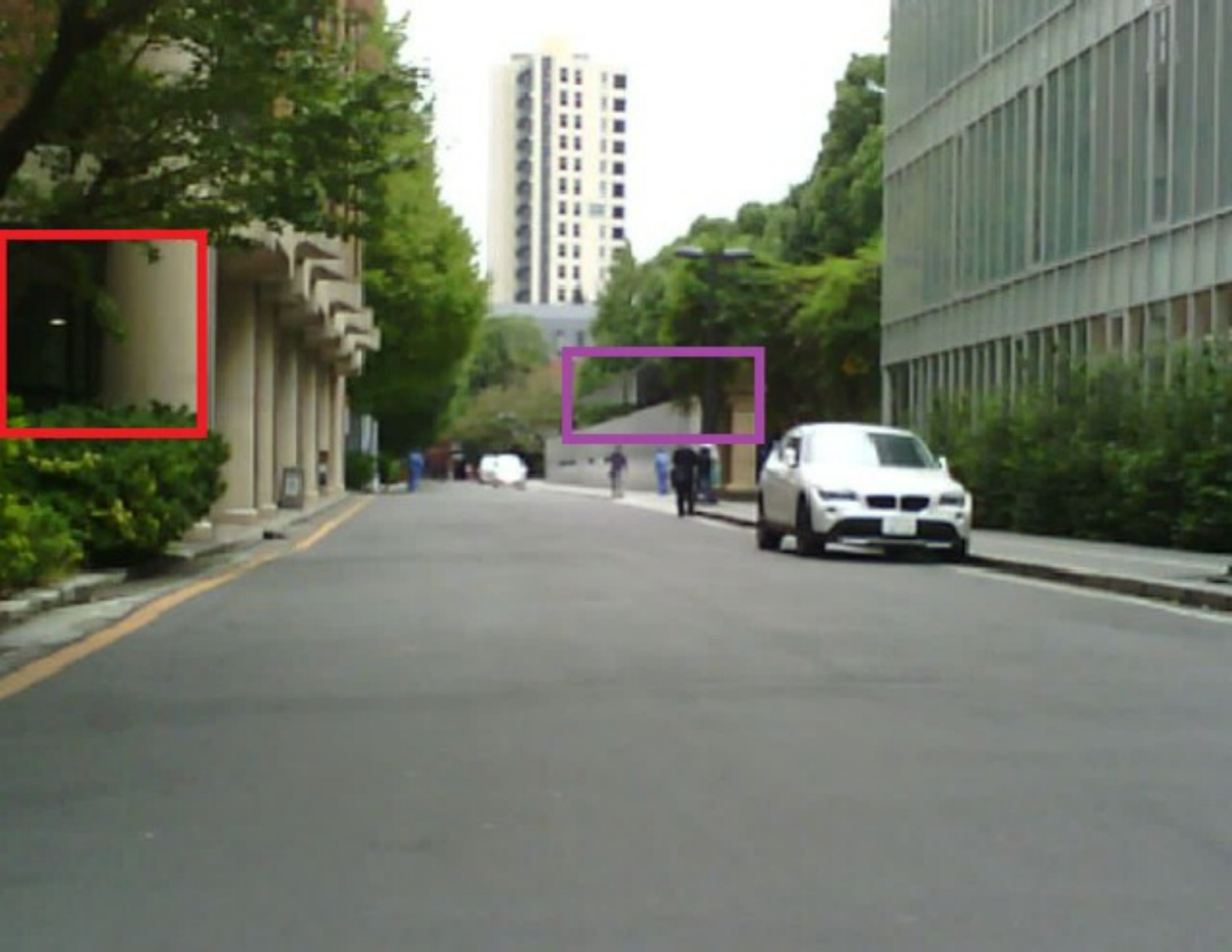}
      {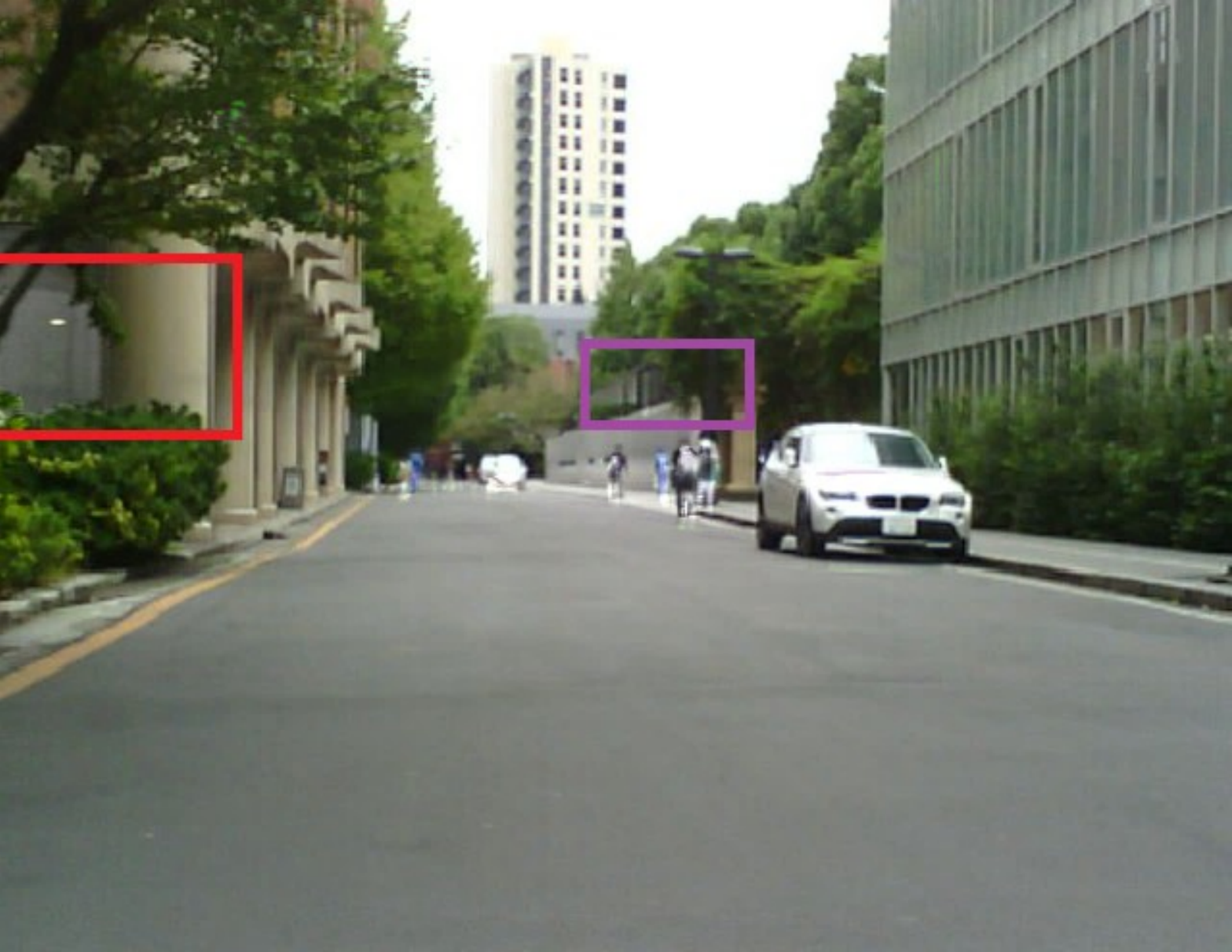}
      {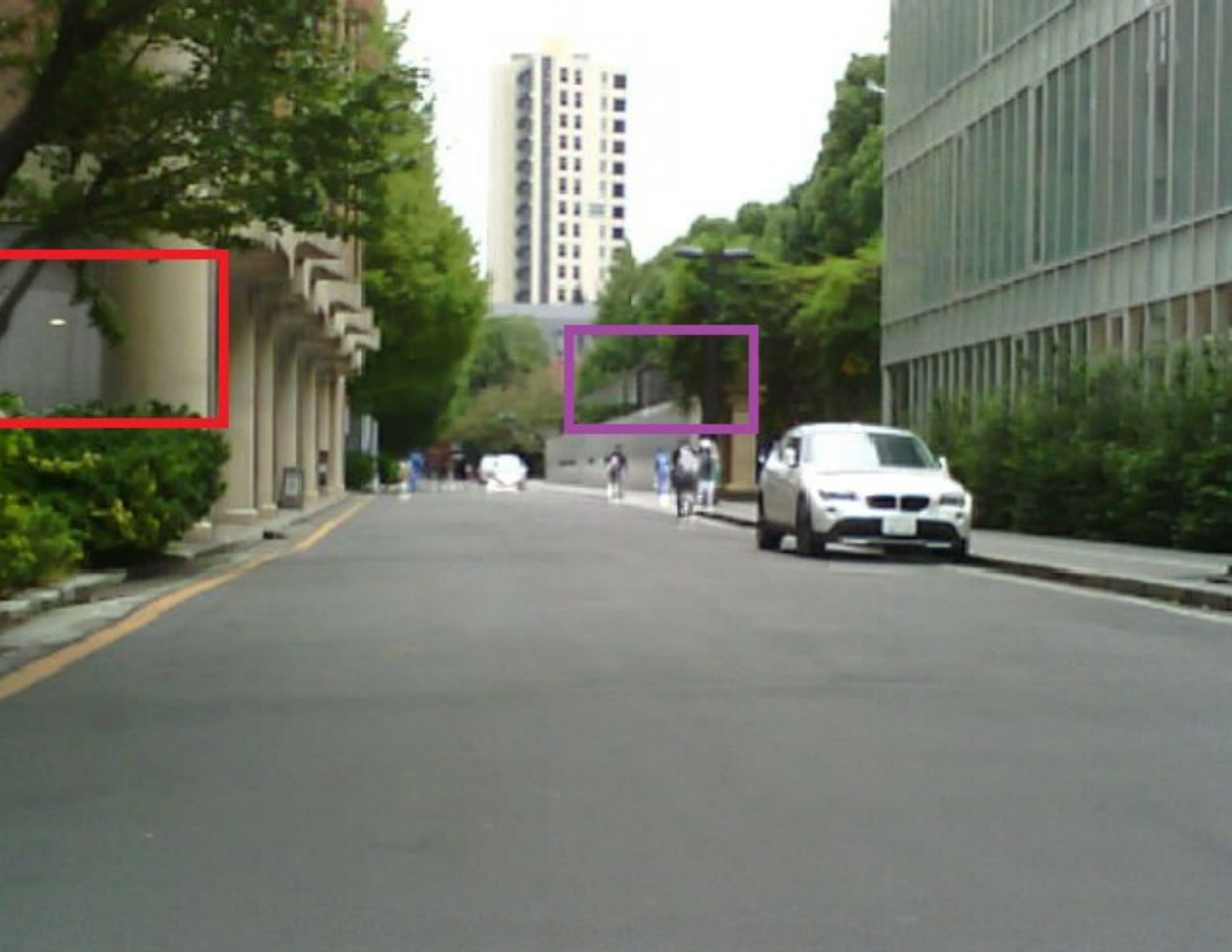}

    \vspace{14pt}

    \onegroupEMS{This is the infrared visible light fusion task. Visible images have the low light degradation.}
      {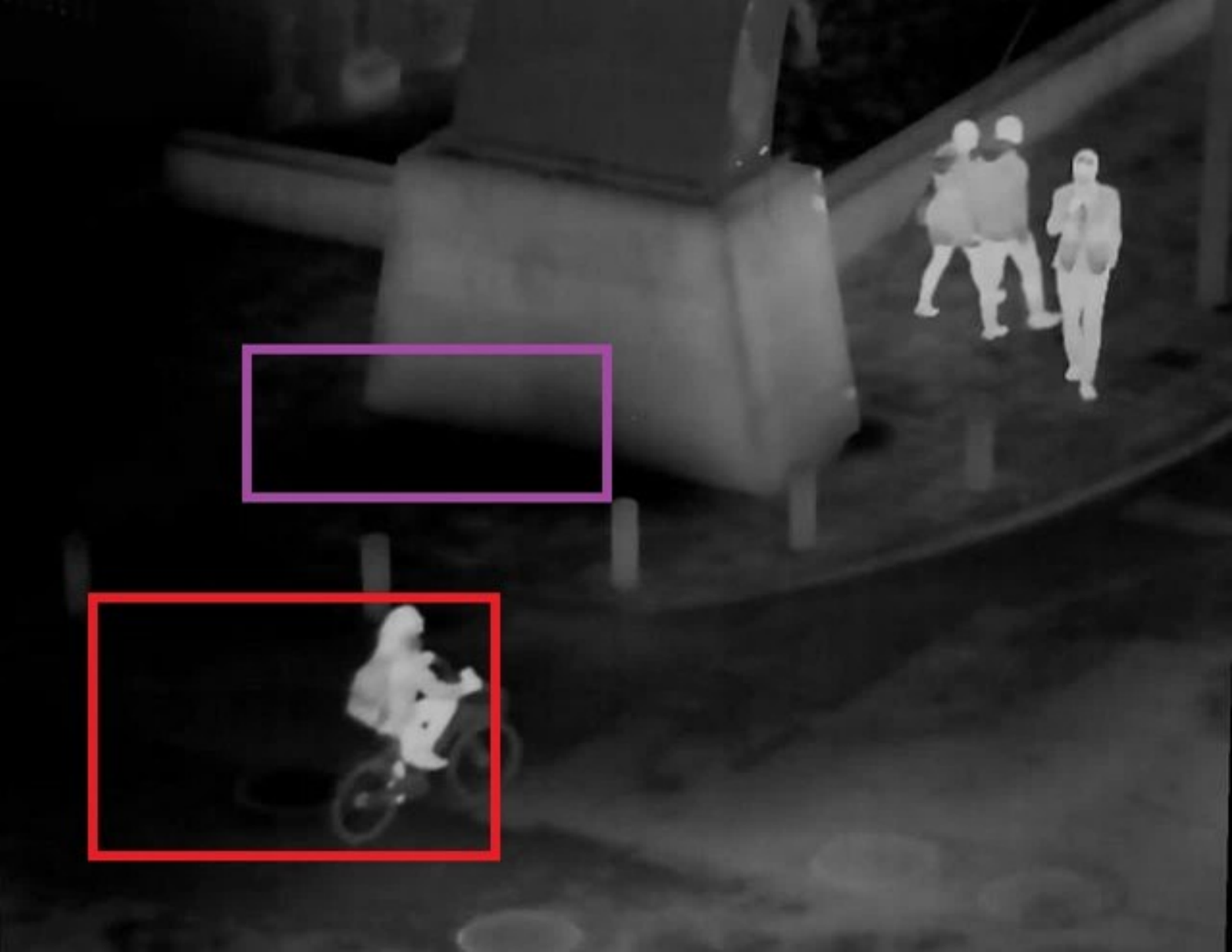}
      {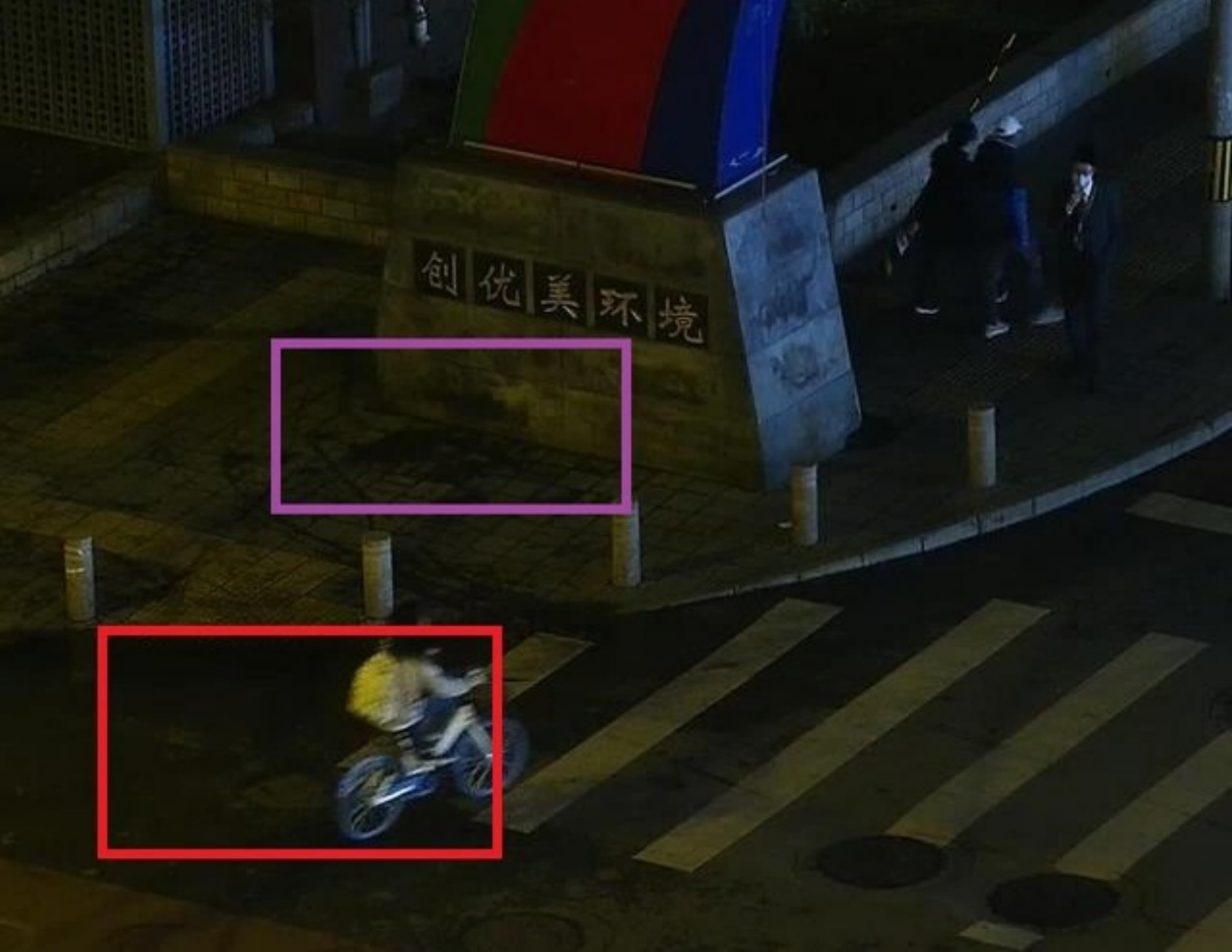}
      {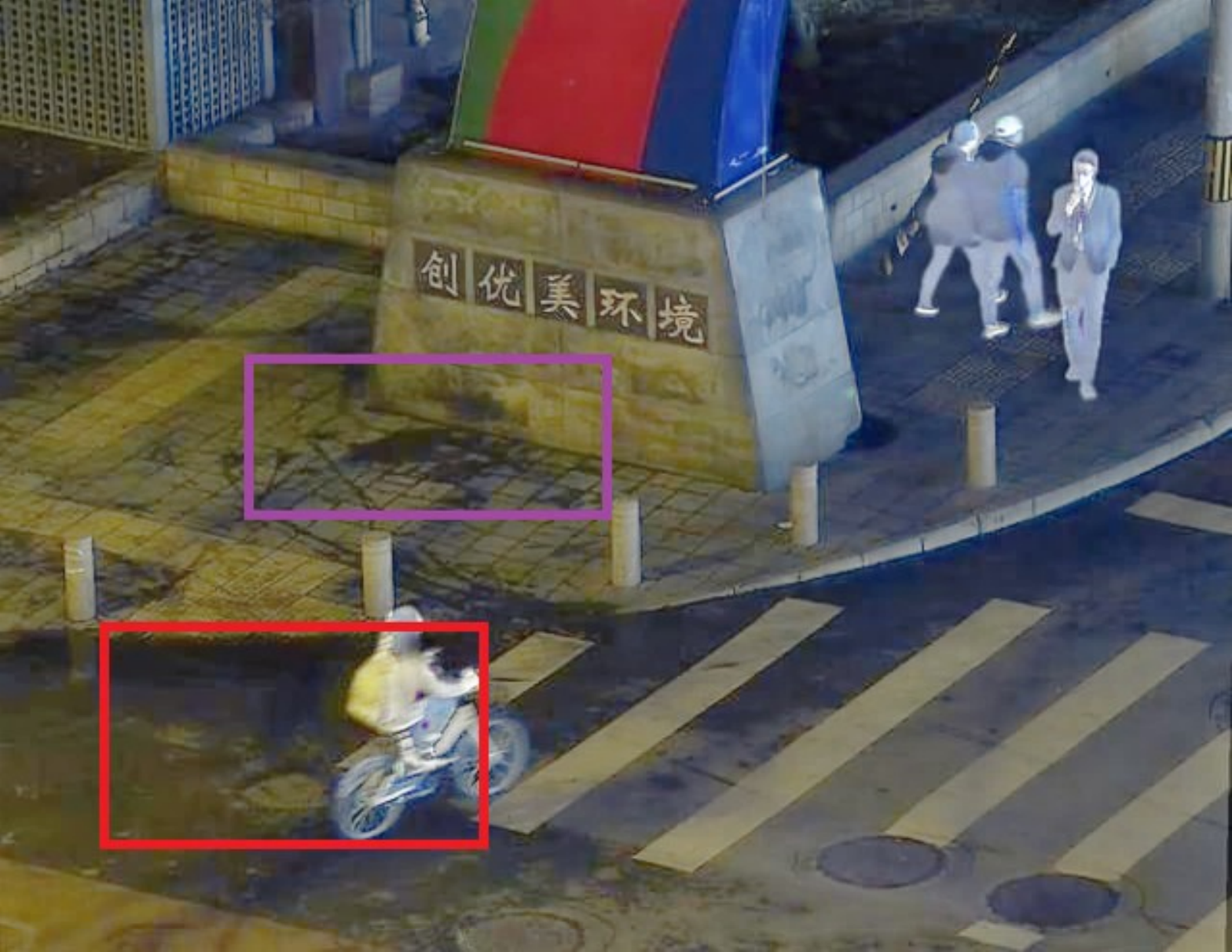}
      {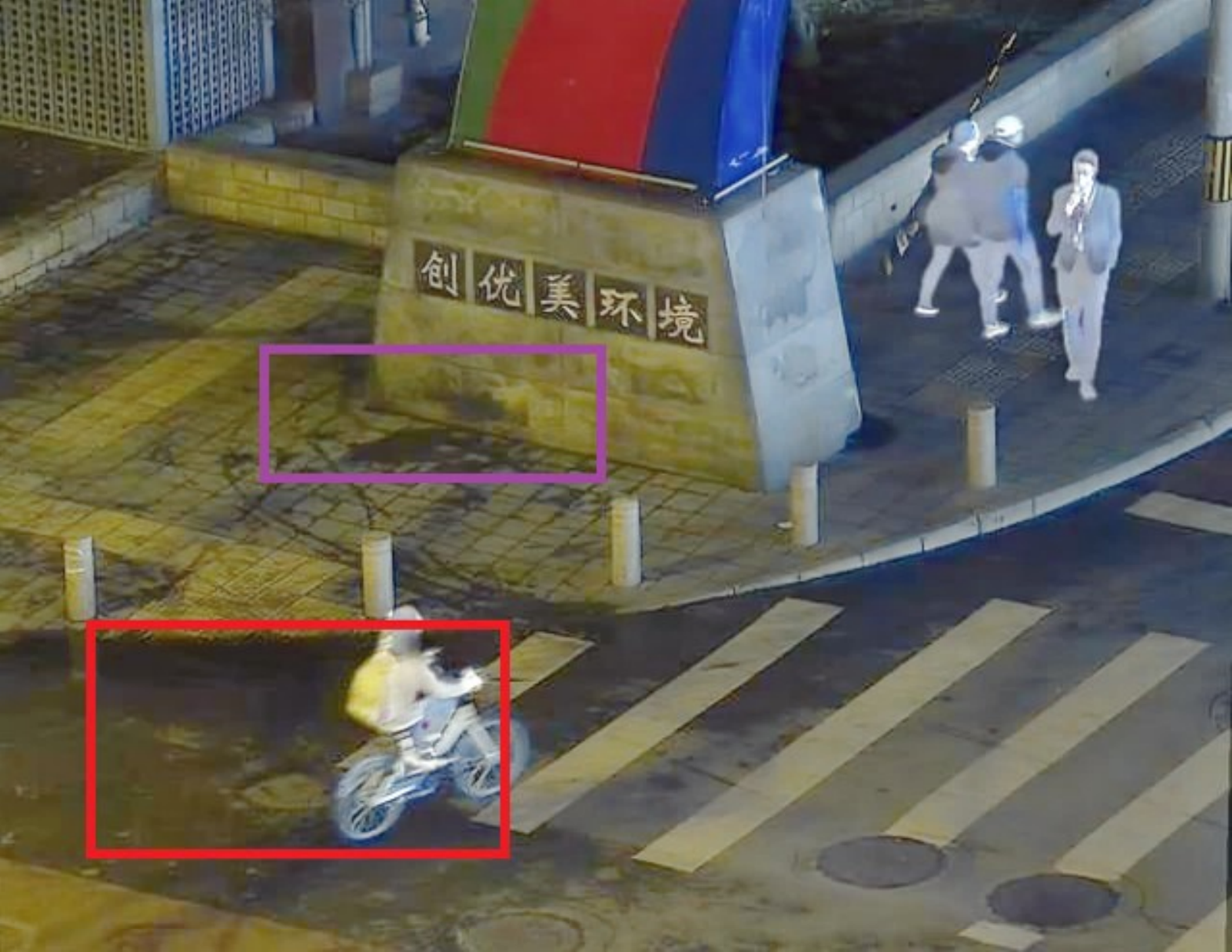}

    \vspace{14pt}

    \onegroupEMS{This is the infrared-visible light fusion task, where visible images are affected by rain degradation.}
      {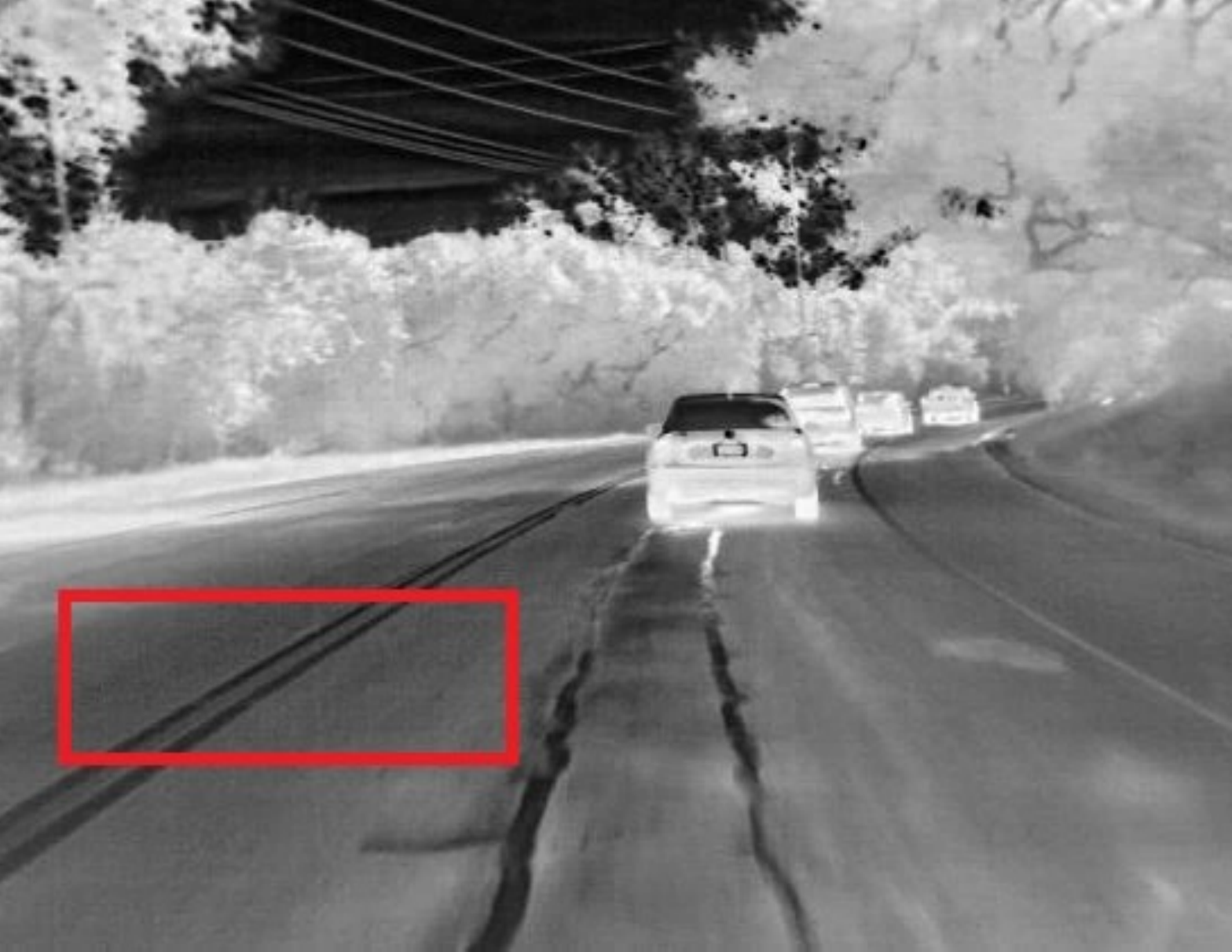}
      {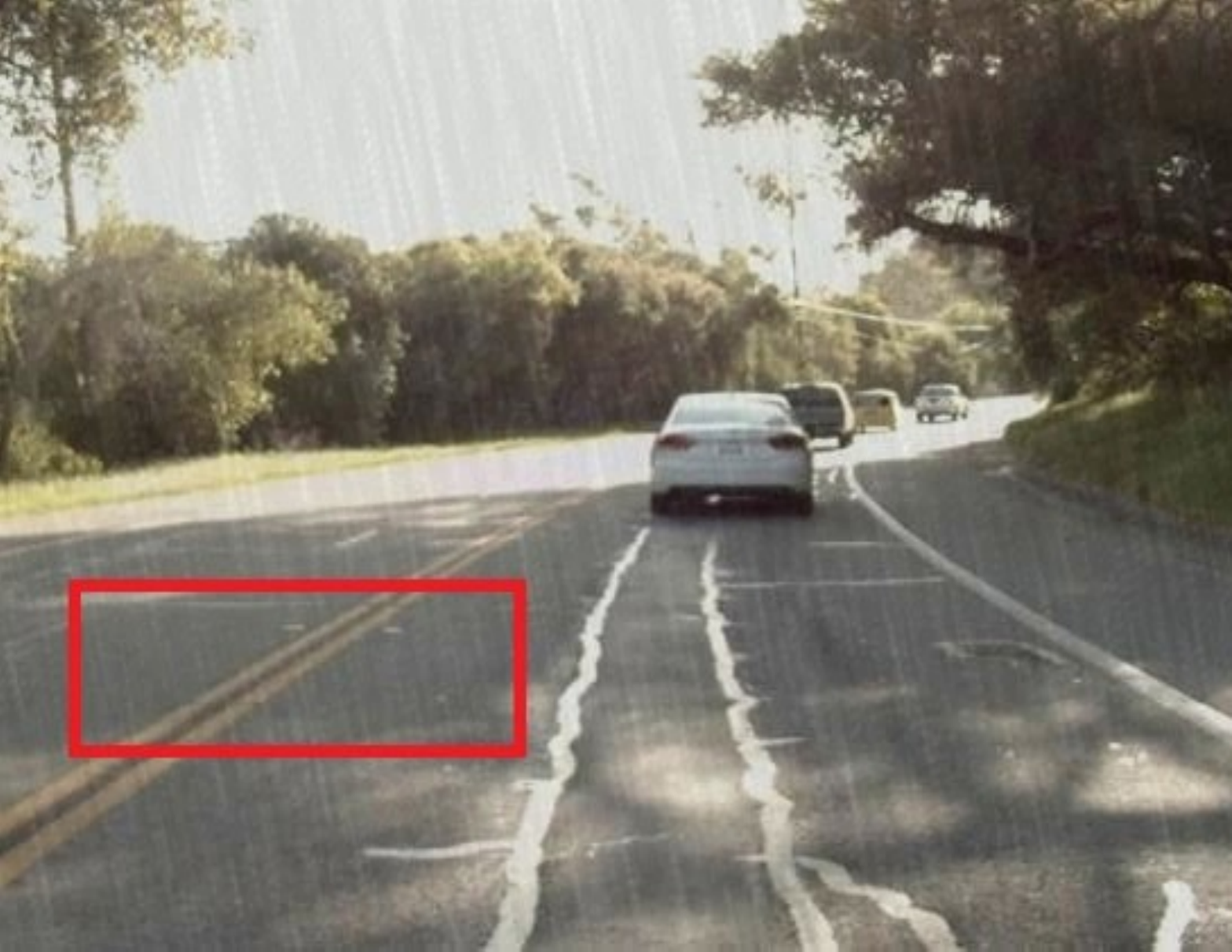}
      {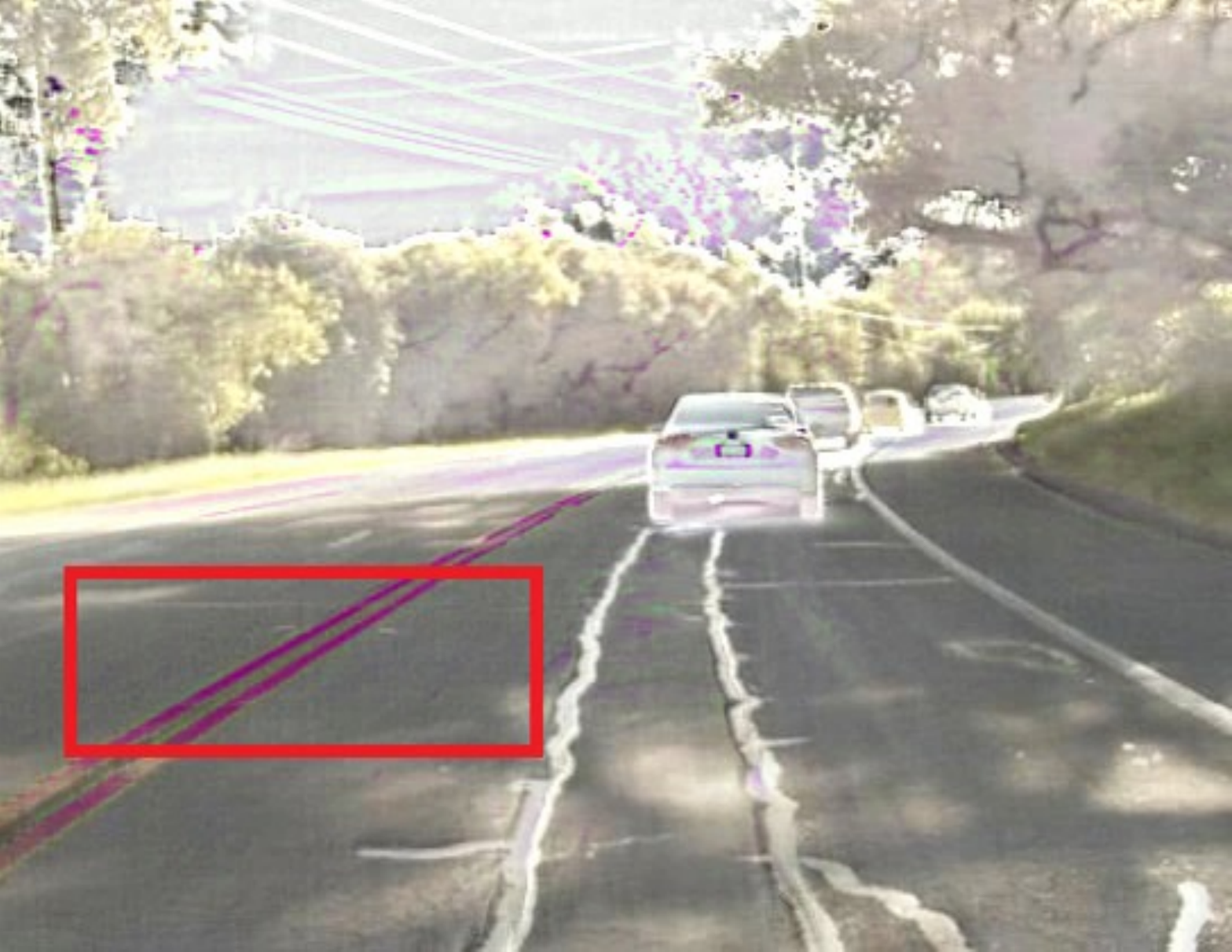}
      {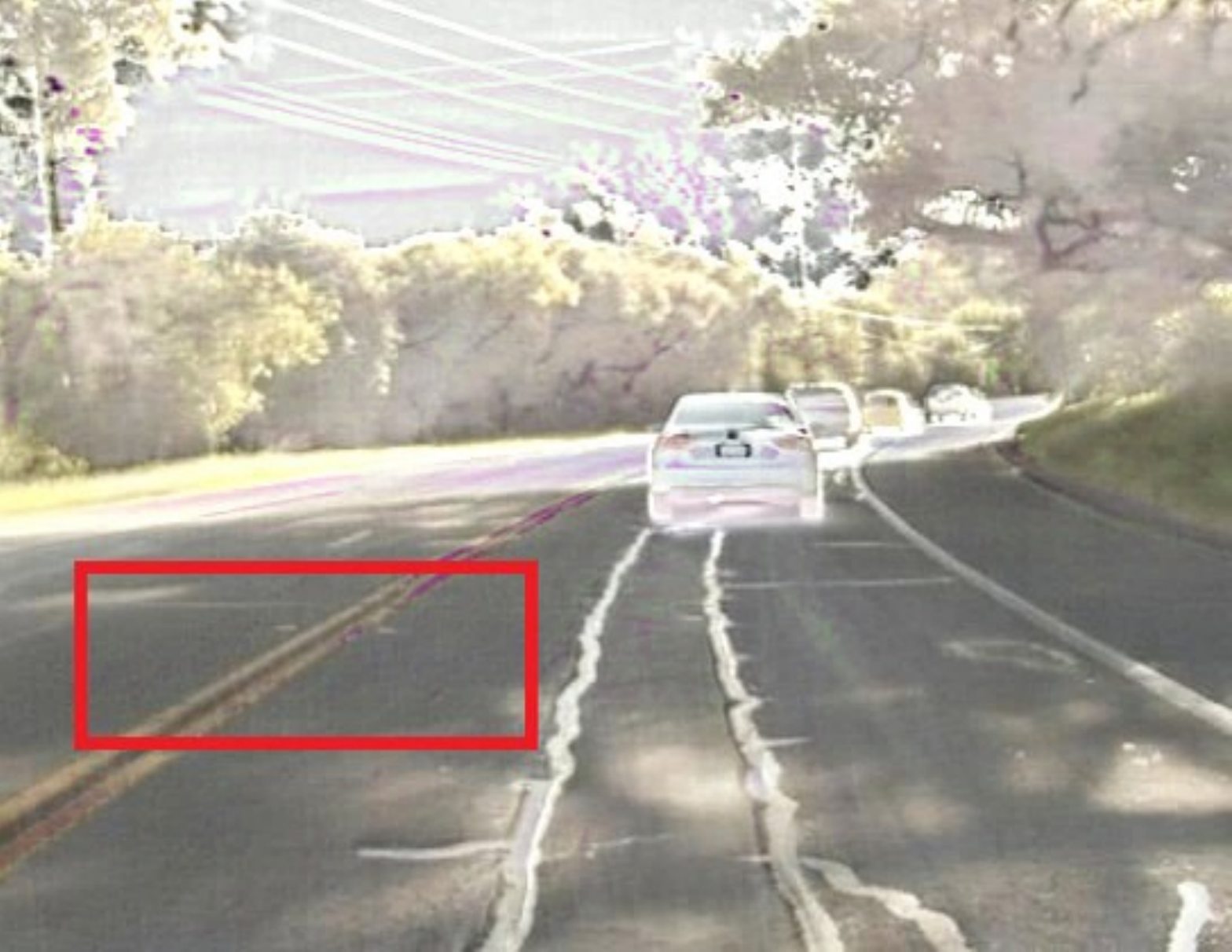}
  \end{minipage}

  \caption{Qualitative comparison on the EMS dataset under six degradation types. Each group contains the infrared input, visible input, the result of Text-IF ~\cite{yi2024text}, and our result. The sentence above each group denotes the degradation-aware text prompt used for text-guided fusion.}
  \label{fig:ems_qualitative_6groups}
\end{figure}

\textbf{Qualitative Comparison.} As shown in Fig.~\ref{fig: qualitative results for standard datasets}, the proposed method demonstrates clear qualitative improvements over Text-IF ~\cite{yi2024text} on three standard datasets (LLVIP, MFNet, and RoadScene), while the visual difference on MSRS is relatively limited. These observations are generally consistent with the quantitative results in Tab.~\ref{tab:comparison_multi_dataset}. On LLVIP, where visible images suffer from severe low-light degradation, our method produces fusion results with more appropriate brightness, improved color saturation, and clearer structural details compared with Text-IF.
For MFNet, which mainly contains low-contrast infrared images, the improvement is particularly evident in terms of target-background separation.
This visual enhancement is consistent with the significant increase in SD, indicating that our model effectively enhances global contrast and intensity variation, making thermal targets more distinguishable.
On RoadScene, our method better alleviates overexposure effects by expanding the dynamic range of the fusion results, leading to more balanced brightness and clearer textures.
In contrast, on MSRS, both methods produce visually comparable fusion results, suggesting that the performance margin under this degradation setting is relatively limited.
\begin{table}[t]
\centering
\caption{
Average quantitative comparison of our method with Text-IF ~\cite{yi2024text} on the EMS dataset.
For each metric, results are first averaged within each degradation type and then equally averaged across the nine degradation types.
$\uparrow$ indicates that
a higher value is better, and $\downarrow$ indicates that a lower
value is better.
(\textbf{Bold}: better result)
}
\label{tab:ems_avg}
\resizebox{\textwidth}{!}{
\small
\setlength{\tabcolsep}{4.6pt}
\renewcommand{\arraystretch}{1.2}
\begin{tabular}{l|cccccccc|cccc}
\toprule
\multirow{2}{*}{Method}
& \multicolumn{8}{c|}{\textbf{Fusion Metrics}}
& \multicolumn{4}{c}{\textbf{Perceptual Metrics}} \\
\cmidrule(lr){2-9} \cmidrule(lr){10-13}
& EN$\uparrow$ & SD$\uparrow$ & SF$\uparrow$ & MI$\uparrow$
& SCD$\uparrow$ & VIFF$\uparrow$ & $Q^{AB/F}\uparrow$ & $\mathrm{SSIM}_{\mathrm{sum}}\uparrow$
& NIQE$\downarrow$ & MUSIQ$\uparrow$ & BRISQUE$\downarrow$ & CLIP-IQA$\uparrow$ \\
\midrule
Text-IF ~\cite{yi2024text} & 7.287 & 48.372 & 15.899 & \textbf{14.481} & 1.447 & 0.784 & 0.566 & \textbf{1.198} & 3.405 & 46.999 & 26.929 & \textbf{0.224} \\
\textbf{IT-TextFusion (Ours)} & \textbf{7.296} & \textbf{48.875} & \textbf{16.402} & 14.474 & \textbf{1.525} & \textbf{0.809} & \textbf{0.577} & 1.195 & \textbf{3.373} & \textbf{47.125} & \textbf{25.747} & 0.212 \\
\bottomrule
\end{tabular}
}
\end{table}
\subsubsection{Evaluation on EMS Dataset}
\label{subsec: eva EMS Dataset}
We evaluate our method on the EMS dataset under diverse degradation conditions. 
Following the evaluation protocol adopted by Text-IF ~\cite{yi2024text}, we report eight fusion-related metrics: EN, SD, SF, MI, SCD, VIFF,
$Q^{AB/F}$, and $\mathrm{SSIM}_{\mathrm{sum}}$. These metrics
evaluate complementary properties of the fusion results, including
information content, contrast, spatial activity, source-information
preservation, structural consistency, and visual fidelity.

\textbf{Quantitative Comparison.} As shown in Tab.~\ref{tab:ems_avg}, our method outperforms Text-IF on several fusion-related metrics, including EN, SD, SF, SCD, $Q^{AB/F}$, and VIFF. The improvements in SCD, VIFF, and $Q^{AB/F}$ suggest that the proposed method can preserve complementary source information and structural details more effectively under diverse degradation conditions. However, the gains are metric-dependent: Text-IF achieves slightly
higher MI and $\mathrm{SSIM}_{\mathrm{sum}}$, while the difference
in EN is relatively small. Therefore, the results should be interpreted as a set of metric-specific improvements rather than uniform superiority across all fusion criteria.

NIQE, MUSIQ, BRISQUE, and CLIP-IQA assess no-reference perceptual
quality and image naturalness from different statistical or learned
perspectives and are not direct measures of multi-modal information
preservation. In particular, although CLIP-IQA uses CLIP-derived
representations, the adopted implementation receives only the fused
image and therefore does not directly measure alignment between the
fusion result and the input degradation-aware prompt. Their values may
also be affected by the statistical characteristics of natural images
and may not fully reflect the effectiveness of infrared-visible image
fusion. Accordingly, these metrics are reported as complementary
evaluation criteria and are not combined with the fusion-related
metrics into a single aggregate score. Compared with Text-IF, our
method obtains lower NIQE and BRISQUE values and a higher MUSIQ score,
while Text-IF achieves a higher CLIP-IQA score. These results indicate
that the proposed method improves several aspects of perceptual quality
but also exhibits trade-offs across different no-reference evaluation
criteria.

\textbf{Qualitative Comparison.} The qualitative results on the EMS dataset under six representative degradation scenarios are shown in Fig.~\ref{fig:ems_qualitative_6groups}. 
Compared with Text-IF ~\cite{yi2024text}, the proposed method produces more stable and visually consistent fusion results across different degradation types.

For the Visible Blur, Low Light, Rain, and Overexposure degradation scenarios, visible images suffer from severe degradation in contrast, brightness, and texture, which significantly increases the difficulty of reliable fusion.
Text-IF can partially enhance scene visibility; however, the fusion results still show constrained brightness, residual saturation, or reduced color vividness in degraded regions, indicating that the loss of visible information is not fully compensated for during fusion.
By contrast, our method produces more balanced fusion results with more appropriate brightness, clearer structural details, and better preserved color consistency, allowing salient targets and scene content to remain distinguishable even under severe degradations.
These representative examples suggest improved visual robustness under
several visible-side degradation conditions.
For low-contrast infrared images, thermal targets are weakly separated from the background, making effective fusion particularly challenging.
In this case, the fusion results of our method are visually comparable to those of Text-IF ~\cite{yi2024text}, and both methods exhibit limited contrast enhancement.
A plausible reason is that the degradation mainly affects the intrinsic contrast of the infrared modality, where discriminative thermal cues are inherently weak and difficult to recover through fusion alone.
Without explicit infrared contrast enhancement priors, both methods tend to preserve the original infrared intensity distribution, resulting in similar fusion performance under low-contrast infrared conditions.

\begin{table}[!t]
\centering
\caption{Ablation study of the proposed method on the LLVIP dataset. (\textbf{Bold} indicates the best performance)}
\label{tab:ablation}

\renewcommand{\arraystretch}{1.15}
\setlength{\tabcolsep}{2.5pt}

\begin{tabular}{c c c c | c c c c c}
\toprule
\makecell{TG\\-RRM} 
& \makecell{Deg.-Aware\\Text\\Eng.} 
& \makecell{Cross-Gate \\Fusion} 
& ITIM 
& EN 
& SF 
& VIFF 
& $Q^{AB/F}$
& NIQE \\
\midrule
$\times$ & $\times$ & $\times$ & $\times$ & 7.443 & 16.153 & \textbf{1.045} & 0.751 & 3.600 \\
$\times$ & $\times$ & $\times$ & $\checkmark$ & 7.444 & 16.296 & 1.040 & 0.752 & 3.472 \\
$\checkmark$ & $\times$ & $\times$ & $\checkmark$ & 7.440 & 16.411 & 1.041 & \textbf{0.765} & 3.353 \\
$\checkmark$ & $\times$ & $\checkmark$ & $\checkmark$ & 7.439 & 16.394 & 1.043 & 0.758 & 3.363 \\
$\checkmark$ & $\checkmark$ & $\checkmark$ & $\checkmark$ & \textbf{7.502} & \textbf{16.464} & 1.037 & \textbf{0.765} & \textbf{3.336} \\
\bottomrule
\end{tabular}
\end{table}

\subsection{Ablation Study}

\begin{figure}[t]
    \centering
    \begin{minipage}[t]{0.32\linewidth}
        \centering
        \includegraphics[width=0.96\linewidth]{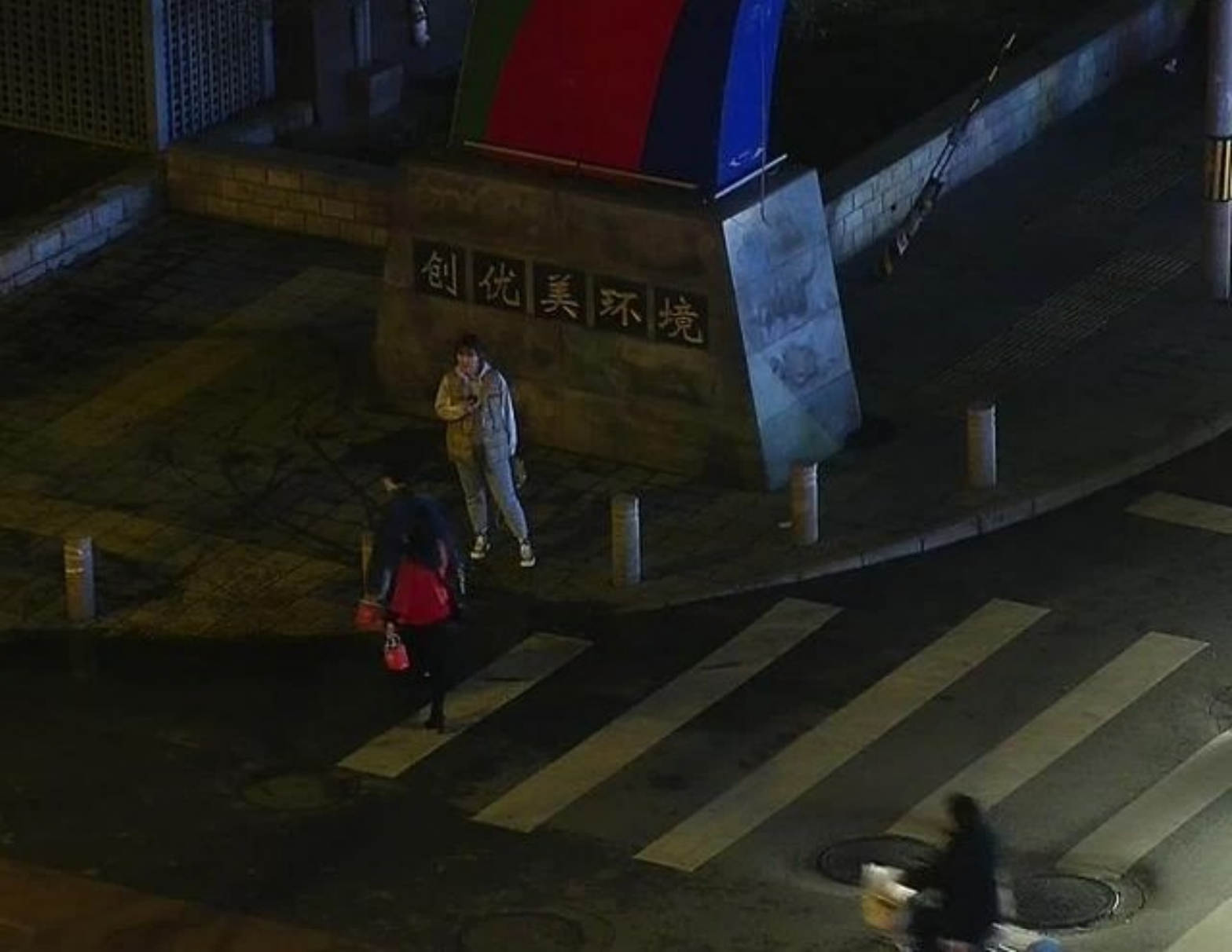}
        \centerline{\footnotesize (a) Visible}
    \end{minipage}
    \hfill
    \begin{minipage}[t]{0.32\linewidth}
        \centering
        \includegraphics[width=0.96\linewidth]{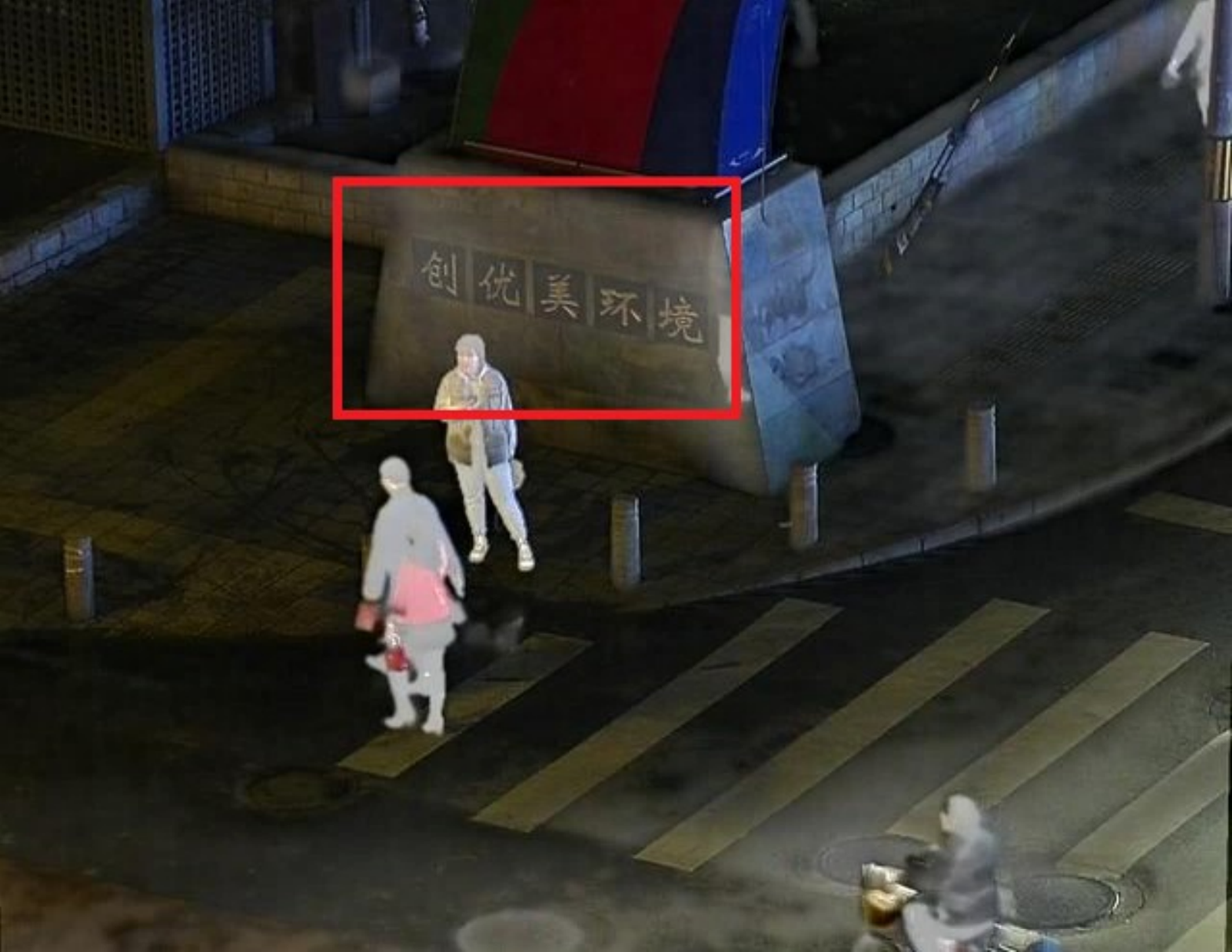}
        \centerline{\footnotesize (b) Original}
    \end{minipage}
    \hfill
    \begin{minipage}[t]{0.32\linewidth}
        \centering
        \includegraphics[width=0.96\linewidth]{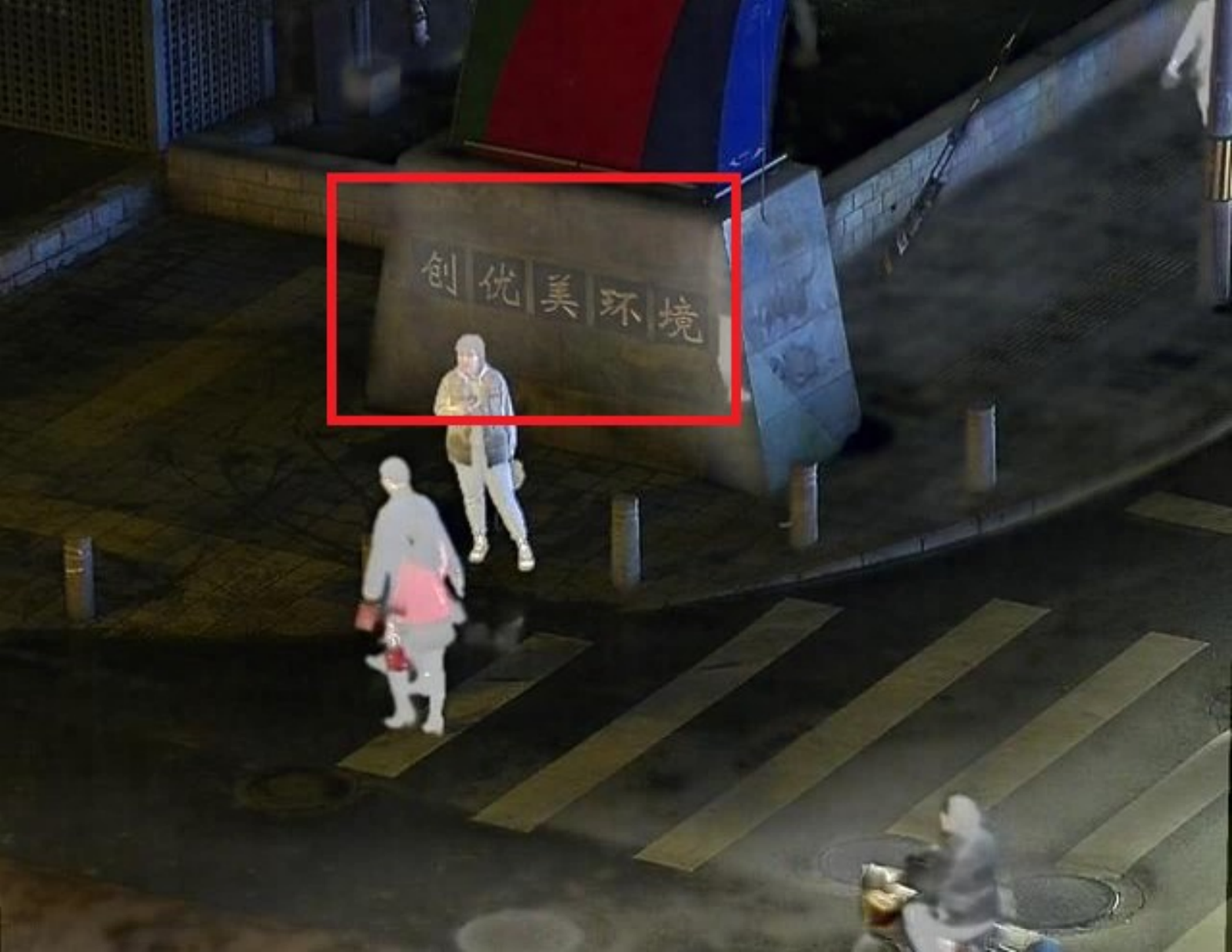}
        \centerline{\footnotesize (c) Only ITIM}
    \end{minipage}

    \vspace{0.5mm} 

    \begin{minipage}[t]{0.32\linewidth}
        \centering
        \includegraphics[width=0.96\linewidth]{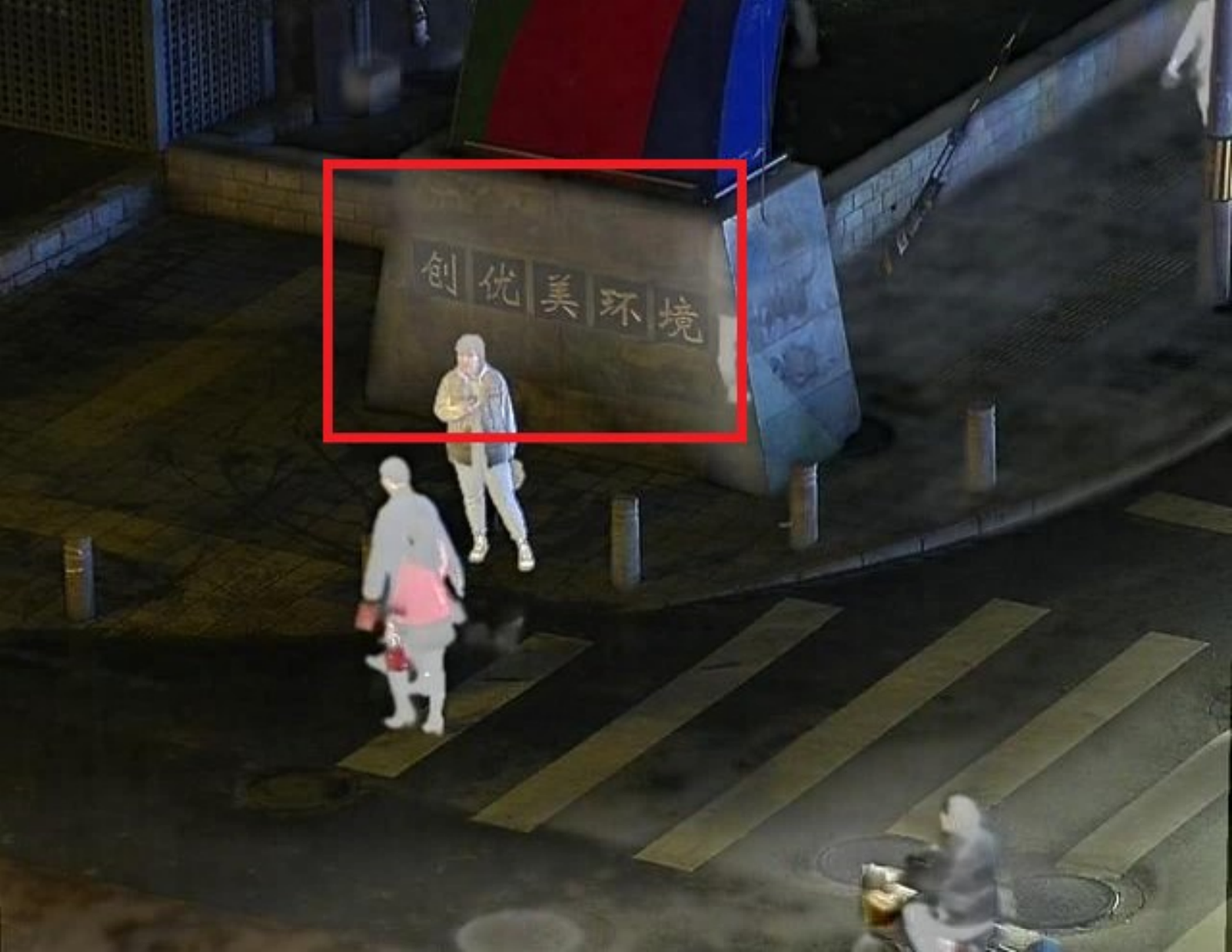}
        \centerline{\footnotesize (d) ITIM+TG-RRM}
    \end{minipage}
    \hfill
    \begin{minipage}[t]{0.32\linewidth}
        \centering
        \includegraphics[width=0.96\linewidth]{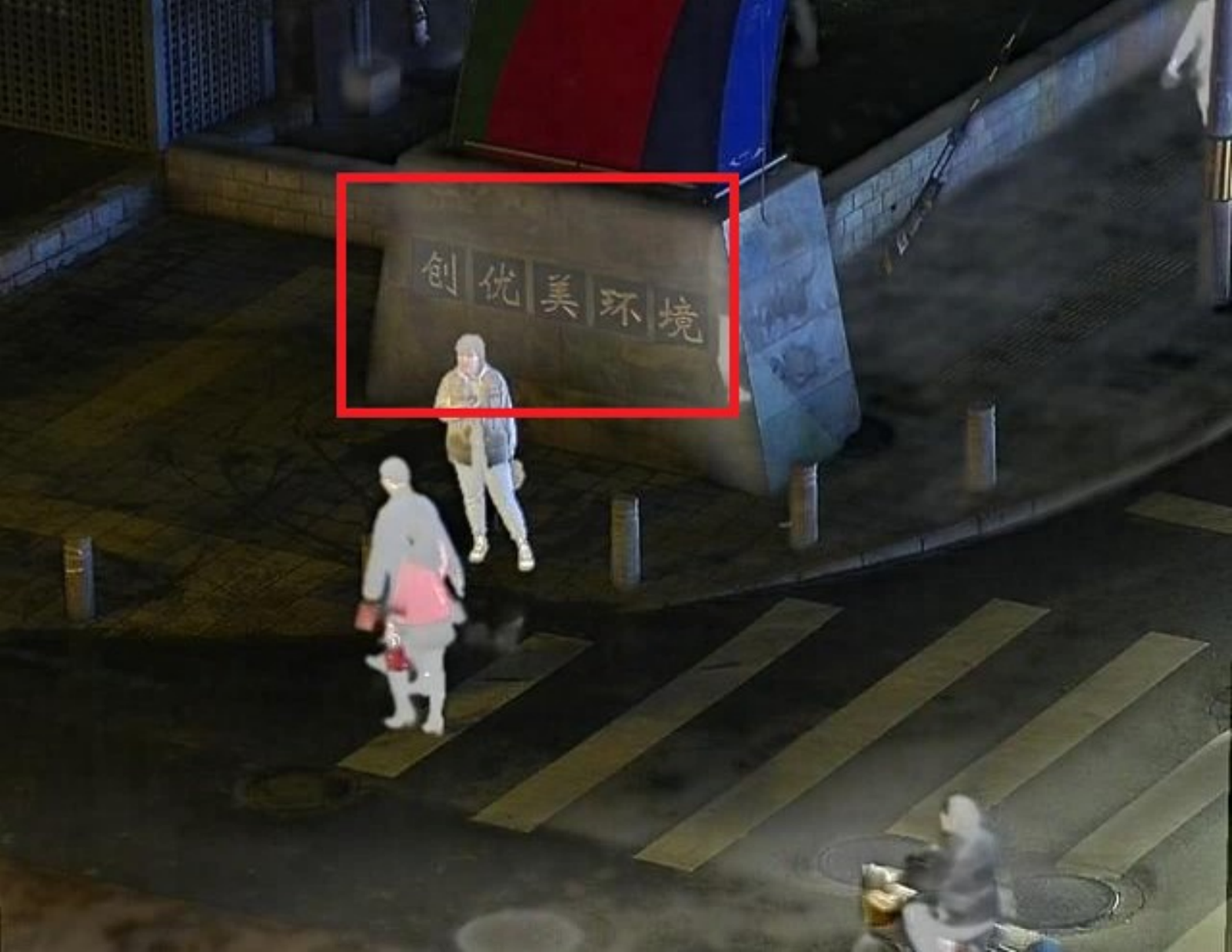}
        \centerline{\footnotesize (e) w/o Deg.-Aware Text Eng.}
    \end{minipage}
    \hfill
    \begin{minipage}[t]{0.32\linewidth}
        \centering
        \includegraphics[width=0.96\linewidth]{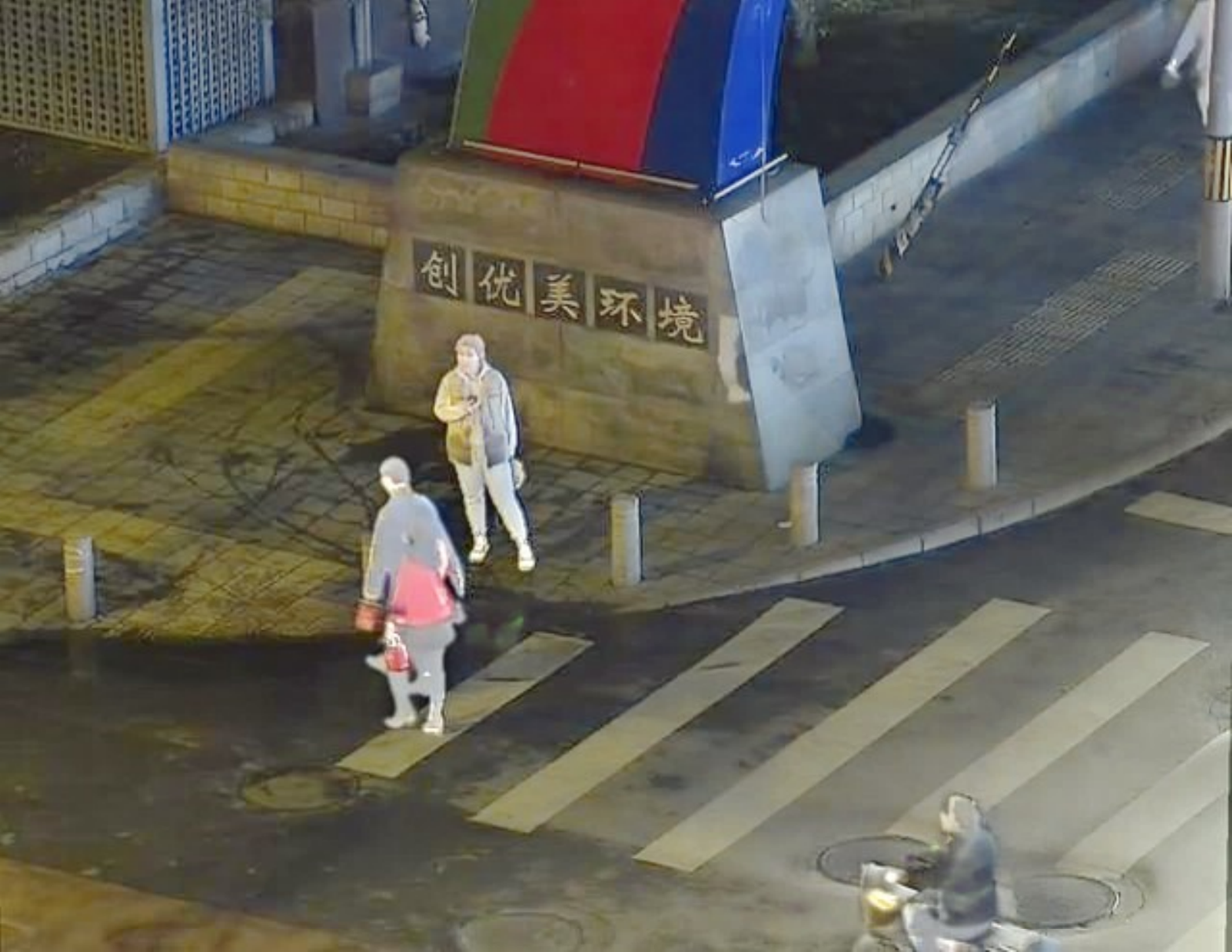}
        \centerline{\footnotesize (f) Complete Model}
    \end{minipage}

    \caption{Qualitative comparison of different ablation variants on the LLVIP dataset.}
    \label{fig:qual_six}
\end{figure}

To further analyze the contribution of each component in the proposed framework, we conduct a series of ablation experiments on the LLVIP dataset.
Starting from a basic fusion backbone, key modules are progressively introduced, including degradation-aware text engineering, ITIM, Cross-Gate Fusion, and TG-RRM. Here, disabling degradation-aware text engineering does not remove the textual branch; instead, the fixed generic prompt is used in place of degradation-aware semantic prompts.
The results are reported in Tab.~\ref{tab:ablation} and Fig.~\ref{fig:qual_six}.

The quantitative results reveal metric-dependent contributions from the proposed components. Under the setting with a fixed generic prompt, ITIM slightly improves EN, SF, and $Q^{AB/F}$ while reducing NIQE, indicating that iterative text-image interaction is beneficial even without degradation-specific semantics. Adding TG-RRM further improves SF and $Q^{AB/F}$ and achieves a lower NIQE, supporting its role in structural refinement and artifact suppression. Cross-Gate Fusion slightly improves VIFF relative to the preceding variant but causes small decreases in several other metrics, indicating a trade-off rather than a uniform gain. Finally, degradation-aware text engineering leads to clear improvements in EN, SF, $Q^{AB/F}$, and NIQE, although VIFF decreases. Consequently, the complete model achieves the best EN, SF, and NIQE and ties for the best $Q^{AB/F}$, but it does not outperform the baseline model in VIFF. These findings suggest that the proposed components jointly balance information preservation, structural fidelity, and perceptual naturalness rather than improving every metric independently.
The qualitative comparisons show that different components exhibit distinct effects on fusion quality, while the complete model produces more balanced results with clearer structures and better visual consistency.
Overall, the complete model achieves the best results on most metrics, indicating that the combined use of these components contributes complementary improvements to overall performance.

%% file: sections/5_Conclusion.tex
\section{Conclusion}

In this paper, we investigate text-guided infrared-visible image fusion under diverse degradation conditions and propose a unified framework that introduces global textual guidance
into hierarchical feature fusion and refinement.
Unlike existing text-guided fusion methods that rely on relatively shallow semantic-visual interaction, our approach
maintains global text conditioning across hierarchical decoding and
residual feature refinement.
Specifically, we enhance prompt expressiveness via task- and degradation-aware text engineering, introduce stage-specific text-conditioned modulation to inject the global text representation into hierarchical decoder
features, and design a Cross-Gate Fusion block for more robust and content-adaptive multi-modal feature integration.
Moreover, we incorporate a lightweight Text-Guided Residual Refinement Module to further improve detail preservation and visual consistency under severe degradations. Extensive experiments on standard fusion benchmarks and the multi-degradation EMS dataset demonstrate that the proposed method achieves competitive overall performance, with improvements on several fusion and perceptual-quality metrics and trade-offs on others. The ablation results indicate that hierarchical text conditioning,
Cross-Gate Fusion, TG-RRM, and degradation-aware prompts can make complementary and synergistic contributions to model performance.

%% file: sections/Acknowledgements.tex
\section*{Data availability statement}
The data that support the findings of this study are available from the corresponding author upon reasonable request.

\section*{Declaration of generative AI and AI-assisted technologies}
During the preparation of this manuscript, the authors used generative AI tools only to improve language and readability. Specifically, the authors used ChatGPT for rewriting in substantial portions of the manuscript. The authors take full responsibility for the content of the manuscript.

\section*{Conflicts of interest}
The authors declares no conflicts of interest.